# Extension of the Blackboard Architecture with Common Properties and Generic Rules

Jonathan Rivard & Jeremy Straub
Institute for Cyber Security Education and Research
North Dakota State University
1320 Albrecht Blvd., Room 258
Fargo, ND 58102
Phone: +1 (701) 231-8196
Fax: +1 (701) 231-8255
Email: jonathan.m.rivard@ndsu.edu, jeremy.straub@ndsu.edu

**Abstract**

The Blackboard Architecture provides a mechanism for embodying data, decision making and actuation. Its versatility has been demonstrated across a wide number of application areas. However, it lacks the capability to directly model organizational, spatial and other relationships which may be useful in decision-making, in addition to the propositional logic embodied in the rule-fact-action network. Previous work has proposed the use of container objects and links as a mechanism to simultaneously model these organizational and other relationships, while leaving the operational logic modeled in the rules, facts and actions. While containers facilitate this modeling, their utility is limited by the need to manually define them. For systems which may have multiple instances of a particular type of object and which may build their network autonomously, based on sensing, the reuse of logical structures facilitates operations and reduces storage and processing needs. This paper, thus, presents and assesses two additional concepts to add to the Blackboard Architecture: common properties and generic rules. Common properties are facts associated with containers which are defined as representing the same information across the various objects that they are associated with. Generic rules provide logical propositions that use these generic rules across links and apply to any objects matching their definition. The potential uses of these two new concepts are discussed herein and their impact on system performance is characterized.

## 1. Introduction

The Blackboard Architecture has been used, previously, for a variety of applications, ranging from math proofs to vehicular control, some of which are described in Section 2. Its capability for storing knowledge, decision-making rules and actuation capabilities within a single common system make it versatile. Because of its use of propositional logic, the decisions that it makes are inherently understandable and reliable.

While the rule-fact-action network has numerous benefits, it does not have an effective way to store non-operational relationships. An association of facts related to a single object or location for example, would need to either be created using a non-system managed naming scheme (embedding the organizational information in the facts' free text names) or a collection of rules which should never be triggered (implementing only associations). Alternately, it could be managed outside of the system. At best, any of these approaches make implementing the organizational structures complex and time consuming. The use of the free text fields is potentially error prone. More problematically, none of these approaches facilitates the use of this organizational / associational data in the system's decision-making processes, despite the fact that it could be potentially useful data for various applications.

The concepts of containers and links were proposed in prior work as a way to model associations which are non-operational (i.e., not propositional logic rules). These concepts, through the introduction of two additional object types, facilitated the implementation of these organizational relationships. Containers are namable objects that store collections of facts and links represent defined (namable) relationships between containers. However, an inherent usefulness limitation still exists, as each container and container-associated fact is considered to be unique. In many cases, containers representing particular objects or real-world concepts may have common information that they store and common logical decision-making and actuation requirements.

This paper, thus, proposes a solution to this need through the introduction of common properties and generic rules. Common properties are facts with well-defined meanings within container-objects. Thus, a given common property for any container can be taken to have the same meaning and used for purposes applicable to this meaning. Generic rules are rules that can act on any link between container-objects meeting the requisite definition, which is stated in terms of required pre- and post-condition common property facts.

Using these two object types, and the previously introduced links and containers, systems can be created which have well defined organizational and other structures with collections of data – and rules which act on this data, potentially triggering actions or modifying other data elements. This can facilitate the automatic creation of understandable networks based upon autonomous sensing while allowing well-defined activities to be conducted using this sensed data. Herein, the implementation of these two object types is described, their performance is evaluated and several possible application areas are considered.

This paper continues, in Section 2, with a review of prior work in several areas which provides a foundation for the work presented herein. In Section 3, the implementation of these new concepts within the Blackboard Architecture is discussed. Section 4 presents an example of the system's use. Next, Section 5 presents the experimental methodology utilized. Section 6, then, presents and briefly analyzes the data collected using this methodology. In Section 7, discussion and analysis are presented before the paper concludes, in Section 8, with a discussion of conclusions drawn from this work and potential areas of needed future work.

## 2. Background

This section reviews work in several areas that provide a foundation for the work that is presented in this paper. First, work on expert systems is reviewed. Then, prior work on the Blackboard Architecture is presented. Finally, prior extensions to the Blackboard Architecture are discussed.

### *2.1. Expert Systems*

The Blackboard Architecture is based on an earlier technology called expert systems. Rule-fact expert systems were introduced in the 1960s and 1970s with systems called Dendral [1] (which separated knowledge storage and the knowledge processing engine [2]), and Mycin [2] (which some consider to be the actual expert system).

Expert systems use a rule-fact network and perform inference, based on rule pre-conditions facts causing rules to be triggered and triggered rules setting post-condition facts to a specified value [3]. The most basic expert systems use facts that have binary values and use rules which assert facts as true, based on two precondition facts being true. They also utilize a rules processing engine which scans the collection of facts and rules seeking those with satisfied preconditions to run (and, by doing this, setting the rules' designated postcondition facts).

This basic concept has been built on with a variety of augmentations. One example is systems which use fuzzy set concepts [4], instead of binary values. Others have worked on optimizing expert systems [5] and creating hybrid learning systems, based on the concept [6,7]. Mitra and Pal [8] suggested a number of variations of these hybrid systems, with fuzzy principles applied in various ways.

Expert systems have been used for numerous applications. Examples include medicine [9,10] (including uses such as for diabetes therapy [11], heart disease diagnosis [12] and hypertension diagnosis [13]), power systems [14], agriculture [15] and education [16].

### *2.2. Blackboard Architecture*

The Blackboard Architecture builds on expert systems. While expert systems are defined by having an ability to state their knowledge and its basis [17], the Blackboard Architecture adds an actuation capability [18]. The system was introduced by Hayes-Roth in 1985 [19] and was based on the prior Hearsay-II system [20,21]. To provide the actuation capability, it adds a third object type, the action, to expert systems' rules and facts [18].

Like with expert systems, the Blackboard Architecture scans its rule-fact-action network for rules with satisfied fact pre-conditions. Rules which trigger can set the values of post-condition facts and/or trigger actions to affect the system's operating environment.

The Blackboard Architecture has been demonstrated for use in numerous application areas, including for design tasks (such as the creation of math proofs [22] and nuclear reactor design [23,24]) and command decision making (such as uses in robotics [25], vehicle control [26], and production [27], project [28] and elevator [29] scheduling). It has also been used for data processing (such as handwriting recognition [30] and data fusion [31]) and for scientific applications (such as modeling proteins [32] and planetary exploration [33]).

### *2.3. Prior Blackboard Architecture Extensions*

The Blackboard Architecture has been expanded and extended in a number of ways. Key capabilities, such as the ability for generalization [34,35] and solving [36–38] have been previously developed. It has also been enhanced through the creation of a problem solving language [39,40], capabilities to enhance its flexibility [41] and the addition of capabilities to support handling and filtering messages [42]. An model based on events [43], and capabilities to support hypothetical and time-based reasoning [44] have also been developed.

A number of projects have sought to expand the Blackboard Architecture's utility for distributed use through the incorporation of distributed agent [45] and multi-agent systems [23,43,46–48] concepts. Distributed processing [49–53], control [54–56] and cooperative operation [57] capabilities have also been developed. Maintenance capabilities, such as pruning to reduce processing time [58,59] and maintenance automation [60], have also been proposed.

## 3. Implementation

This section discusses the implementation of the proposed system. In Section 3.1., the basic rule-fact network is discussed. Section 3.2 presents an overview of the container and link objects, which were introduced in prior work. Next, in Section 3.3, common properties are introduced and described. Finally, in Section 3.4, generic rules are presented and their implementation is described.

### *3.1. Basic Rule-Fact-Action Networks*

The Blackboard Architecture, at its core, is a network of facts and actions which are connected by rules. The basic fact object in the Blackboard Architecture is a node that stores a Boolean value. Facts represent if some real-world observable datapoint that can be true or false. Rules connect these facts to each other. Conceptually, each rule can have as little as one to an infinite amount of input and output facts; however, not all implementations will require this amount of flexibility. Rules are selected for execution when all of their prerequisite conditions are met. When they run, they set their post-condition facts to their specified values. Running rules can also launch actions to actuate in the operating environment.

The implementation used herein allows rules to be developed with up to four input and output facts. When a rule is evaluated for execution, all of the values for its input facts are assessed and, if they are true, then the rule is triggered. In this implementation, triggering a rule sets all of the indicated output facts of the rule to true. Actions were not utilized in this study; however, these could also be triggered from conventional or generic rules, as dictated by operational needs.

This model works well for many application areas; however, it has a limited capability to support complexity. As the complexity of a project grows, the number of facts needed can increase quickly – even exponentially, in some cases.

### *3.2. Containers and Links*

To help organize larger, more complex networks, containers were introduced, previously. Containers are collections of facts which are assigned a description. They can be used for numerous applications. For example, containers could be used to symbolize computers on a network, people in a human resources system, or packages in a warehouse, among many other things. To allow the system to model relationships between containers in the network, directional links were also created. Links hold the ID of two containers: an origin (first) container and a destination (second) container. A basic two container link network is shown in Figure 1.

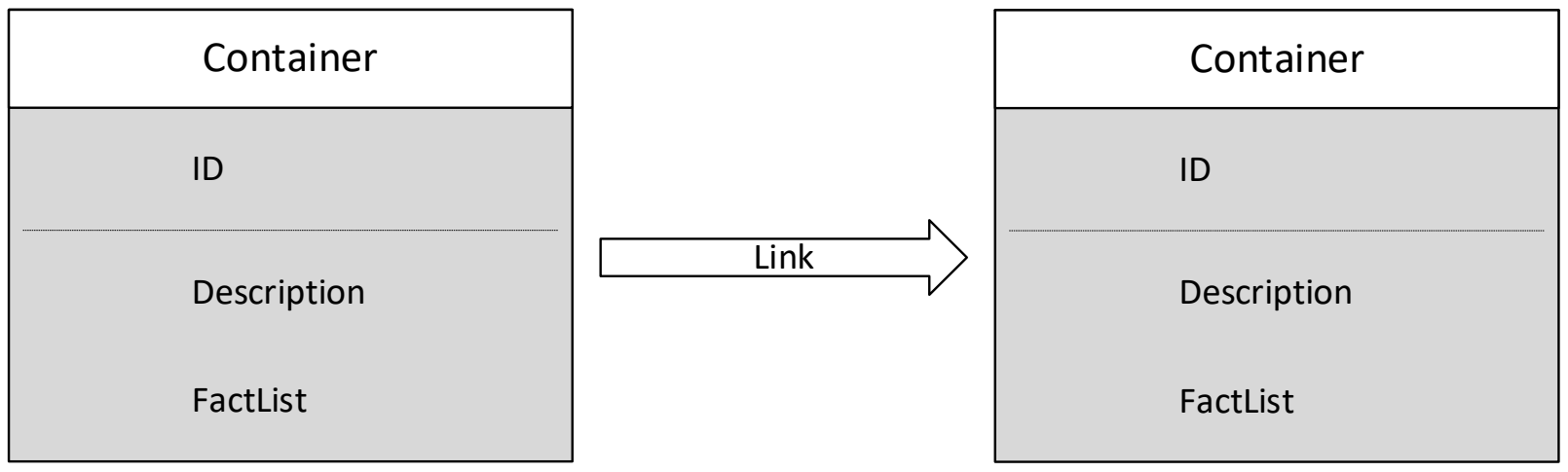

Figure 1. Network containing two containers connected by a link.

In the example of computers on a network, each container could represent a computer. Facts would then be used to indicate properties of these computers. Thus, to represent whether a computer has a Wi-Fi capability, a fact could be included in the container with the fact description of “has Wi-Fi”. Other facts could be used to store information about other hardware and software capabilities. Links can then be used to model how computers connect to each other via ethernet or Wi-Fi. For example, in a computer network, multiple computers could be connected to a network switch. In the Blackboard Architecture representation, this would be modeled as each computer having a directional link to the switch and the switch having a directional connection to each computer. This would model the multi-directional flow of data under normal circumstances and also facilitate modeling more complex environments, such as firewall restrictions and equipment damage.

This creates a challenge, though, when another container (computer) also needs to store the same information. If another computer also needs to indicate that it “has Wi-Fi” a fact would also be needed for this purpose. This cannot be the same fact as used for the first computer, as the values may be different, so a separate fact with the same description could be used. However, there is no way for the system to know that these two facts – despite having the same or a similar description, refer to the same type of data. For some applications, a way to recognize that both facts store the same data, yet are unique to their containers, would be useful. This is the purpose of common properties.

### *3.3. Common Properties*

Common properties are a robust type system for facts. Each fact still stores its own value, but facts which are instances of a common property have a description (and therefore a meaning) which is part of a common property data collection.

Common property definitions are comprised of a unique ID and a description field. Because the description is defined at the common property level, facts which are instances of common properties do not have their own description value (instead using that of the common property). They are, thus, comprised of fact-specific unique IDs, a stored reference to the common property’s unique ID, and their own value. Figure 2 shows the relationship between instance facts and common properties.

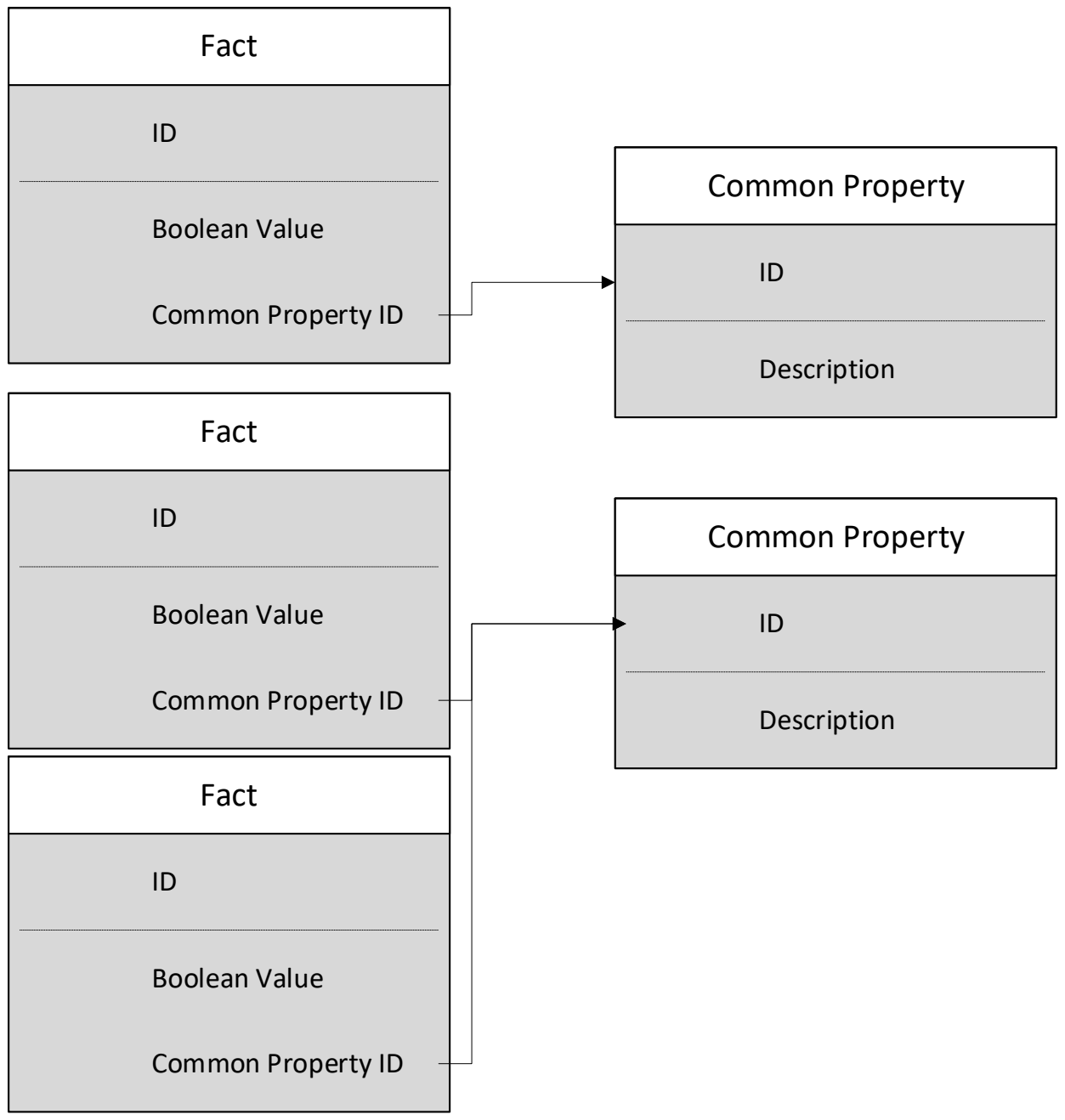

Figure 2. Multiple instance facts referencing common properties.

Using common properties, facts in separate containers can be compared or utilized with the knowledge that they represent the same data regarding the container. In the example of computers in a network, it is now easy to identify computers (or only execute rules for computers) that have a Wi-Fi capability.

### *3.4. Generic Rules*

Generic rules build upon and take advantage of common properties to facilitate logical decision making that is instance-agnostic. While a basic rule is defined in terms of and interacts with facts using their individual unique IDs, generic rules operate on containers and their associated facts, identifying pre- and post-condition facts based upon their common property IDs. Each generic rule contains the details shown in Table 1 and is implemented as shown in Figure 3. A generic rule can be run on any containers matching its definition.

Table 1. Data stored for generic rules.

| | Container One | Container Two | Booleans |
|---|---|---|---|
| Before Run | Common Properties, Desired Values | Common Properties, Desired Values | IgnoreIfNotPresent |
| After Run | Common Properties, Desired Values | Common Properties, Desired Values | CreateIfNotPresent |

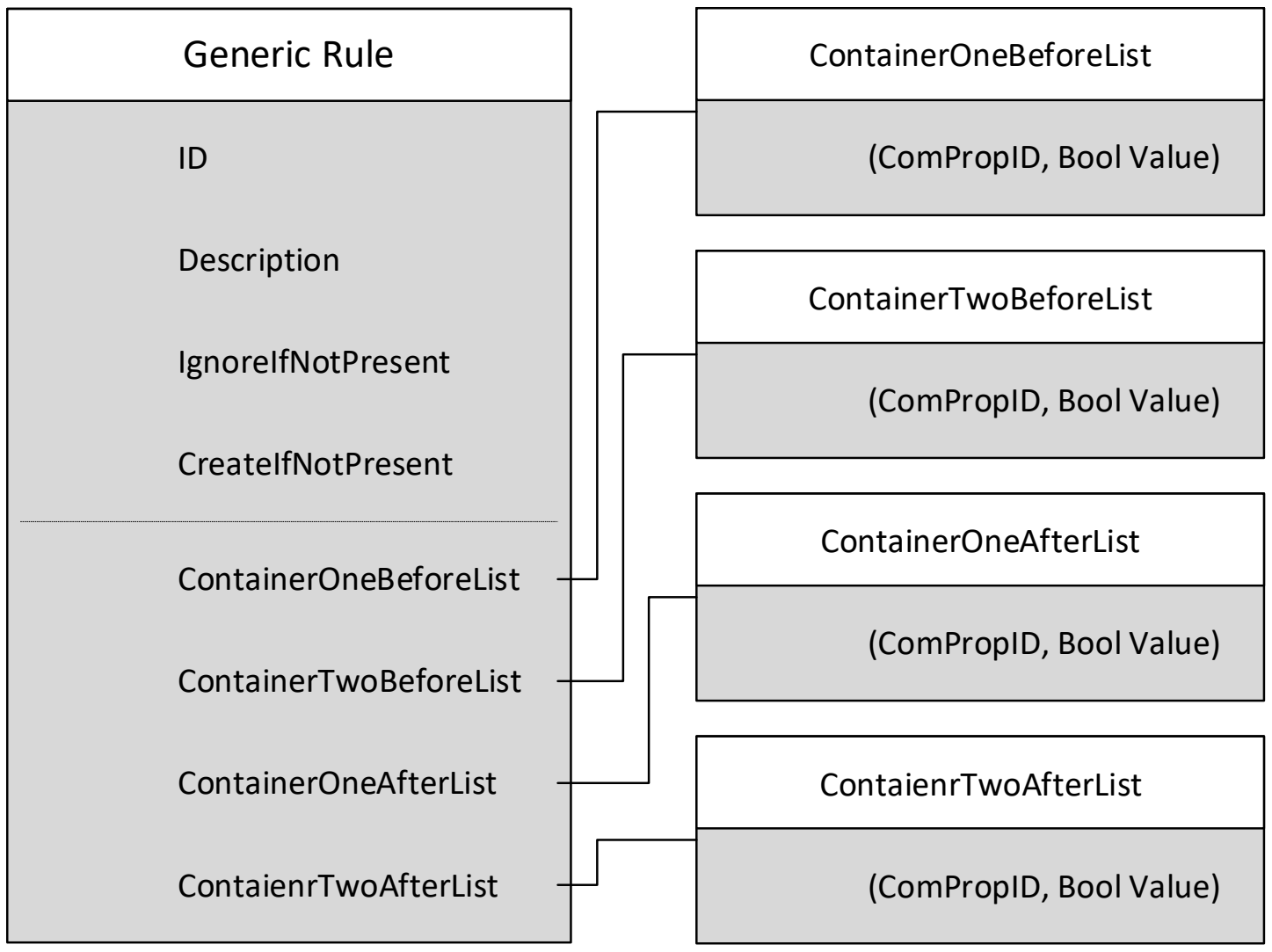

Figure 3. Generic rule implementation.

The container one and two 'before' and 'after' conditions are implemented as four lists of tuples. Each tuple is made up of a common property ID and its desired value. To determine whether a generic rule can be run on two containers, these containers' facts are validated. The 'before run' lists are used, initially, and the containers' facts are checked to see if they have the required common properties present and the required values stored in the tuples. The IgnoreIfNotPresent Boolean value indicates whether the absence of a 'before run' common property causes a mismatch. The CreateIfNotPresent Boolean indicates whether the absence of an 'after run' common property will cause it to be added to the container.

The operations of generic rules are comprised of two phases: checking and execution. Checking only involves the 'before run' row of Table 1. The facts of the both containers are checked for the 'before run' common properties and if their actual values match the required ones. If any common properties. in the container fact list are not present or their values do not match the required ones, the checking step fails and the generic rule doesn't run for these containers. The only exception to this is if the IgnoreIfNotPresent value of the generic rule is set to true. If the IgnoreIfNotPresent Boolean is set to true, any common properties that are not present in the container are ignored instead of causing the checking step to fail.

If the checking step succeeds, the system moves onto execution. During execution, the facts indicated in the 'after run' row of Table 1 are set to their target values by searching though the facts of both containers and updating the fact values of any facts where the common property IDs match the generic rule's definition. Any facts without common property matches are ignored. If the CreateIfNotPresent value of the generic rule is set to true, then a fact will be added with the common property ID and target value for any common properties listed in the 'after run' row which are not present in the container.

**4. Example Uses**

The proposed capabilities, described in Section 3, have numerous uses. To demonstrate how these capabilities can be utilized, an example is presented in this section. This example use is a human resources management system and the task, in particular, of promoting an employee to a manager position is shown. Figure 4 shows the model that was implemented for this example.

This system implemented for this demonstration is comprised of two containers. One is for an employee and one is for the front desk team. Every employee on the front desk team is linked to the front desk container. Notably, employees could instead be placed as sub-containers in the front desk container. At present, though, container nesting is not supported. It is planned as a potential future feature.

The employee, John Doe, has two facts with the common properties isEmployee and isManager set to true and false respectively. The front desk container has two facts with the common properties isTeam and hasManager set to true and false respectively. There is a link connecting John Doe to the front desk team. To promote John Doe to the manager of the front desk team, the promote employee generic rule must be run.

This rule is simplified somewhat, as a real-world implementation would need to evaluate or be told which employee should be promoted so that multiple employees are not promoted (or the first employee found is not just promoted). The operations are comparatively simple. First, the rule checks that John Doe is an employee and isn't a manager. Because the IgnoreIfNotPresent property is true, the rule skips the objection that John Doe doesn't have isManager fact. Then the rule checks that the front desk team is a team and doesn't have a manager. Because both of these checks pass, the generic rule executes. Because the CreateIfNotPresent property is true, John Doe has an isManager fact added with the value true. Then the front desk team has its hasManager fact updated to true.

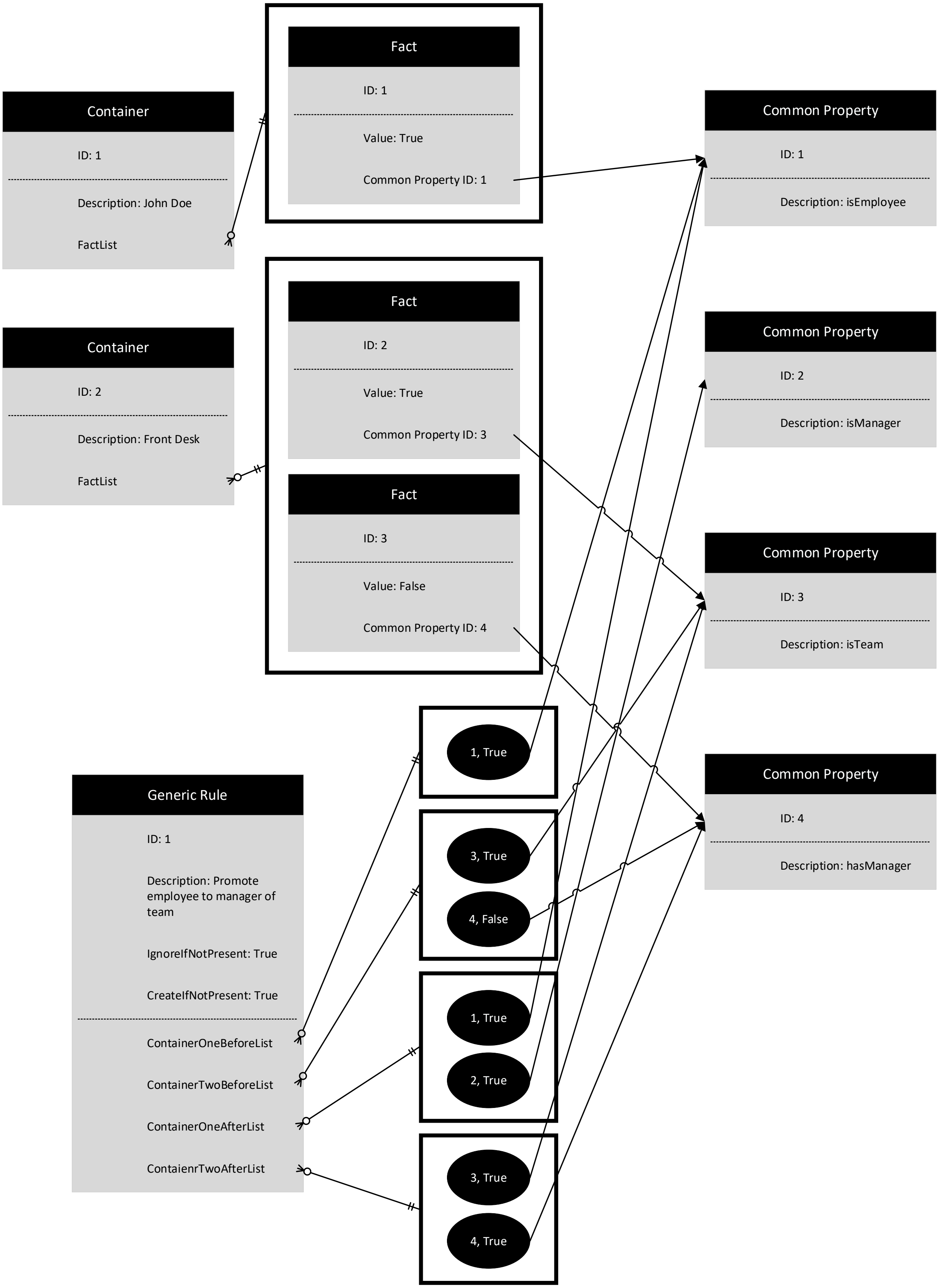

Figure 4. Model for a human resource management system.

## 5. Experimental Methodology

To test the effectiveness of the system, multiple experiments were run. Simulations were run with randomly generated networks using the generation process described below. Fifty-six combinations of input parameters were tested. For each of the input combinations, 50 simulations were run and the results were averaged. The standard parameters for the simulations are shown in Table 2.

Table 2. Standard Input Parameters.

| Input Parameter | Value |
|---|---|
| Fact Count | 100 |
| Rule Count | 100 |
| Link Count | 100 |
| Container Count | 50 |
| Common Property Count | 50 |
| Number of Common Properties per Rule | 10 |
| Fact Assignment Method | Uniform (alternative of Random) |
| Hybrid Rule Chance | 50% |
| IgnoreIfNotPresent Chance | 25% |
| CreateIfNotPresent Chance | 25% |

### *5.1. Random Network Generation (Common Properties, Facts, and Generic Rules)*

A random network is generated with multiple steps. First, common properties and facts are created. Facts are randomly assigned to be associated with common properties. Second, generic rules are created. There are three different way to generate generic rules: uniform, random, and hybrid. Each of these ways had an equal chance of occurring for each rule generated. The uniform generic rule generation method randomly selected common properties for the container and the two lists. The pre- and post-execution lists have matching common properties; however, the Boolean value is randomly generated. The random generic rule generation method randomly selects common properties for all four lists with randomly generated Boolean values. The hybrid generic rule generation method uses a mix of the two. It randomly selects common properties for the container and two lists; however, in the post lists each common property has a chance (based on the hybrid rule chance input parameter) to randomly select a different common property. Values are still randomly generated. All generic rules have a chance to have the IgnoreIfNotPresent and CreateIfNotPresent settings enabled, based on the input parameters with the same names.

### *5.2. Random Network Generation (Containers, Links, and Picking Start and End Points)*

The next step in network generation is assigning facts to containers. Depending on the selected fact assignment method, facts can either be assigned uniformly or randomly. Under uniform assignment, containers are iterated through and each is assigned a random remaining fact until all facts are depleted. Under Random assignment, facts are iterated through and assigned to a random container.

The next step is randomly picking two unique containers to act as the start and end points for traversal simulation. Finally, the most difficult and last part of network generation, creating a linked network, is performed.

Links could be created randomly but this leads to the problem of containers being isolated or the start and end containers not being connected. To create an interconnected network, while there is an equal amount or more links left to create than containers, one link is created for each container to another randomly selected container. After there are not enough remaining links to assign one to each container, the start and end points of the links, assigned to two random, unique containers, are used. Then the linked network is checked for traversability. A depth first search is started from the start container to the end container. The check requires a minimum depth of the container count divided by two, to prevent networks with start and end containers linked directly together. The search will also timeout at a depth of 1000. If a path from the start to end container isn't found, it has too low of a complexity, or times out, all links are deleted and the link creation process is started over.

### *5.3. Testing Results*

Multiple results from each test are stored. The data collected is listed in Table 3. It is important to note that each result is an average of 50 tests done for each input combination. Parameters with the word average in the name are an average of 50 averages of each state in the simulated network.

Table 3. Output Parameters.

| Result Parameter | Details |
|---|---|
| Combination ID | ID given to each input combination |
| Test ID | ID 0-49 given to each individual test |
| Network Settings | Each input parameter is recorded in the results output |
| Time to Link | Time (in ticks) to run the depth first search from start to end containers |
| Average Time to State | Average time (in ticks) to traverse between two containers |
| Total Traversal Time | Total time (in ticks) to traverse from start to end containers |
| Initial Network Size | Initial network storage size (in bytes) |
| Average State Size | Average storage size (in bytes) of each state |
| Total Storage Size | Total storage size (in bytes) for a test |
| Path Length | Length of path between start and end containers |

### *5.4. Simulating Traversal*

After network generation, the network is searched using a depth first search algorithm to find the shortest link path between the start and end containers. The network then simulates a traversal across this path. For each link in the path, each generic rule is checked if it is compatible. If the generic rule is compatible with the current link, the rule is run, and the changes are output to a changes file explained in Section 5.5. The most notable drawback of this method is that some rules may become compatible because other rules have run; however, they are not checked again. While some possibilities may be missed, this approach is used to prevent loops of rules changing the state of the network between two states across numerous iterations (or forever) and, thus, interfering with data collection.

### *5.5. Storing Network Data*

During the testing process, data is saved in four stages. The first stage is the "saves" folder. In this folder, the initial network generation is output to a text file with the name "idNumber.txt." The ID Number matches the test number given in the input dataset. Each line follows the comma separated format shown below.

| Type | Action | Value1 | Value2 | …. |
| --- | --- | --- | --- | --- |

The type field has a value of either a setting, start container, end container, shortest path, or an item in the network such as a fact or link. From the data stored, the entire network can be reinitialized to the same state. The action field states setting names for the setting type or addition. Then, the value after the type and action depends on what is being defined. An added fact line has three values: fact ID, common property ID, and an initial Boolean value. A generic rule, on the other hand, is much longer. It has an ID, title, Boolean values indicating the create and ignore properties are enabled, and then four period-separated lists of common property ID – Boolean pairs.

After an initial network is initialized and saved, the second stage of data is stored in the "changes" folder. Each link the network traverses during the traversal simulation has one text file assorted with it named "testId-traversalIterator.txt." The test ID is the ID of the given test as well as its initial save file name. The traversal iterator starts at zero and, for each link, increases by one. Each file, then, only saves which link the chain is at, which generic rules are run, and what changes they implement. The changes that generic rules cause are stored in the same format as the initial save file with a type, action, and values. Changes are facts being added or changed.

The third stage is collecting all test results into a comma separated value list. First, each combination has all of its tests results exported to a comma separated list in a folded named "comboResults." Then, all these csv files are combined into one final dataset file that includes the average data for each combination.

**6. Data and Analysis**

This section considers the performance of the simulated traversal system. Each of the input parameters outlined in Section 4 was tested as an independent variable and the results outlined in Table 2 were recorded. All of the results are separated into two categories: uniform fact assignment and random fact assignment. Because the categories often have results that roughly mirror each other, they are compared in every figure, to look for differences.

***6.1. Fact Count***

The first area of analysis is how the number of facts in the network affects its results. Figure 1 shows how the number of facts affects the amount of time needed to find the shortest path. The time required (tick count) stays between 150 and 250 for all fact counts with a notable result of uniform fact assignment taking longer than random fact assignment, except for at the highest fact count level tested. This makes sense as the number of facts themselves does not greatly affect the efficiency of the program, rather the number of paths would cause additional processing requirements.

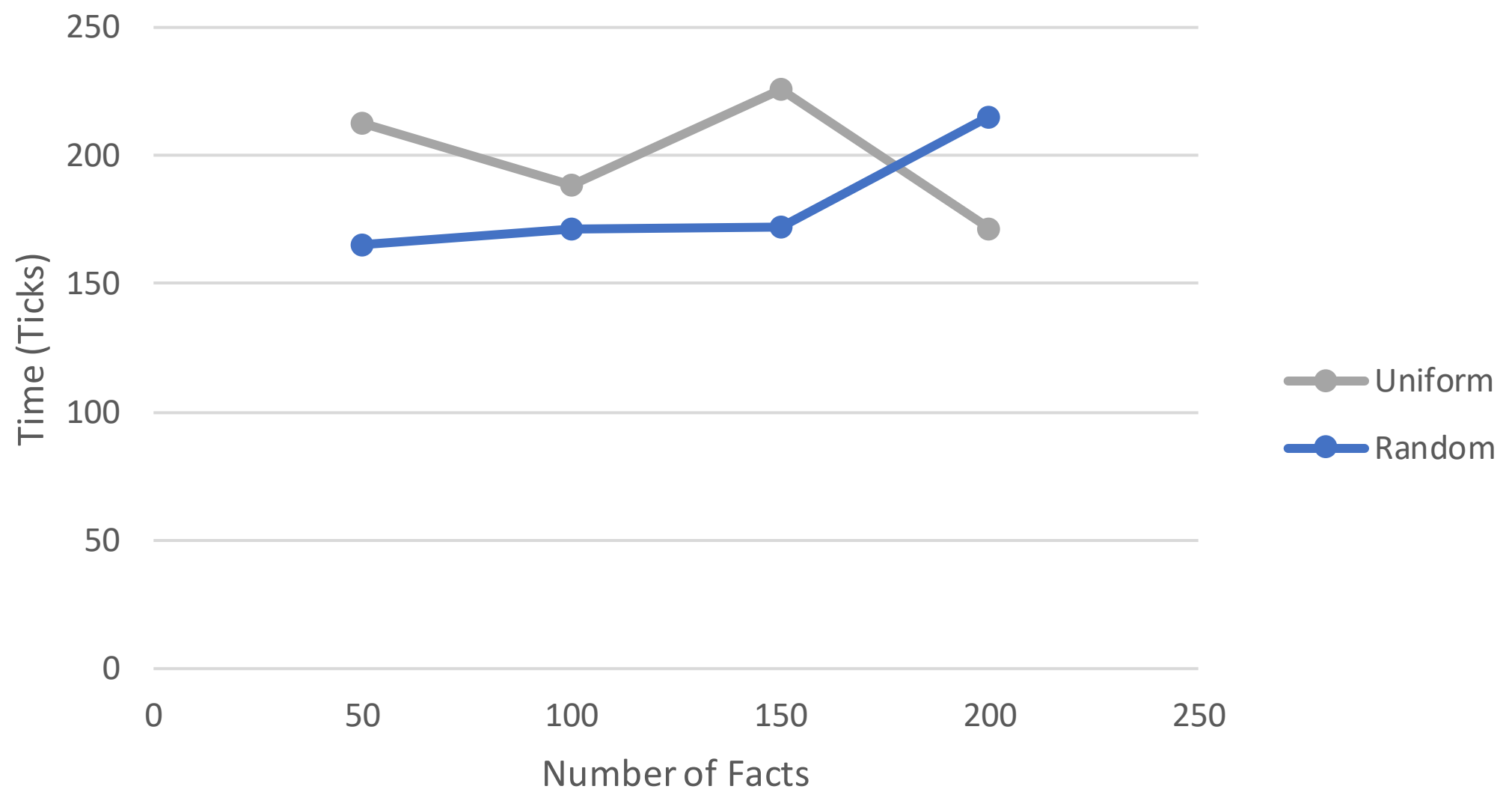

Figure 1. Fact Count by Time to Find Shortest Path

Figure 2 shows the average amount of time (ticks) to reach the next state in the network. While the data shows that the uniform fact setup initially has a slightly higher time requirement than the random one, that switches between the 100 and 150 facts levels. Also, as all of the data points are within 200 ticks of each other, the difference is minimal.

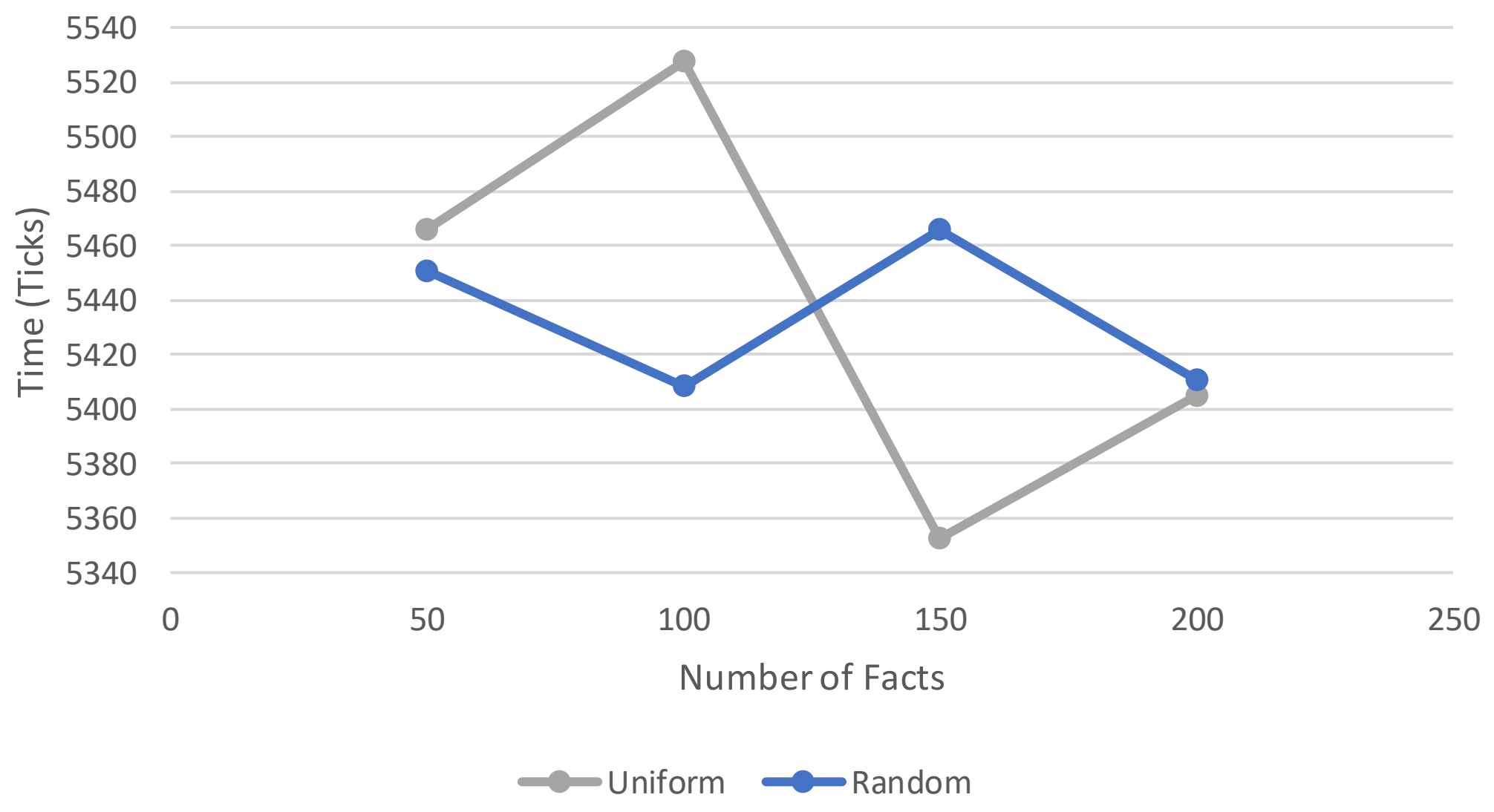

Figure 2. Fact Count by Average Time to Next State.

Figure 3 shows the total time (in ticks) needed to traverse the network. The uniform and random networks mirror each other while the number of facts increases. Uniform networks start high, at 50 facts, and then decrease until 150 facts. There is then a slight increase at 200 facts. Random networks do the inverse, starting low, increasing until 150 facts, and then showing a slight decrease at the 200 facts level.

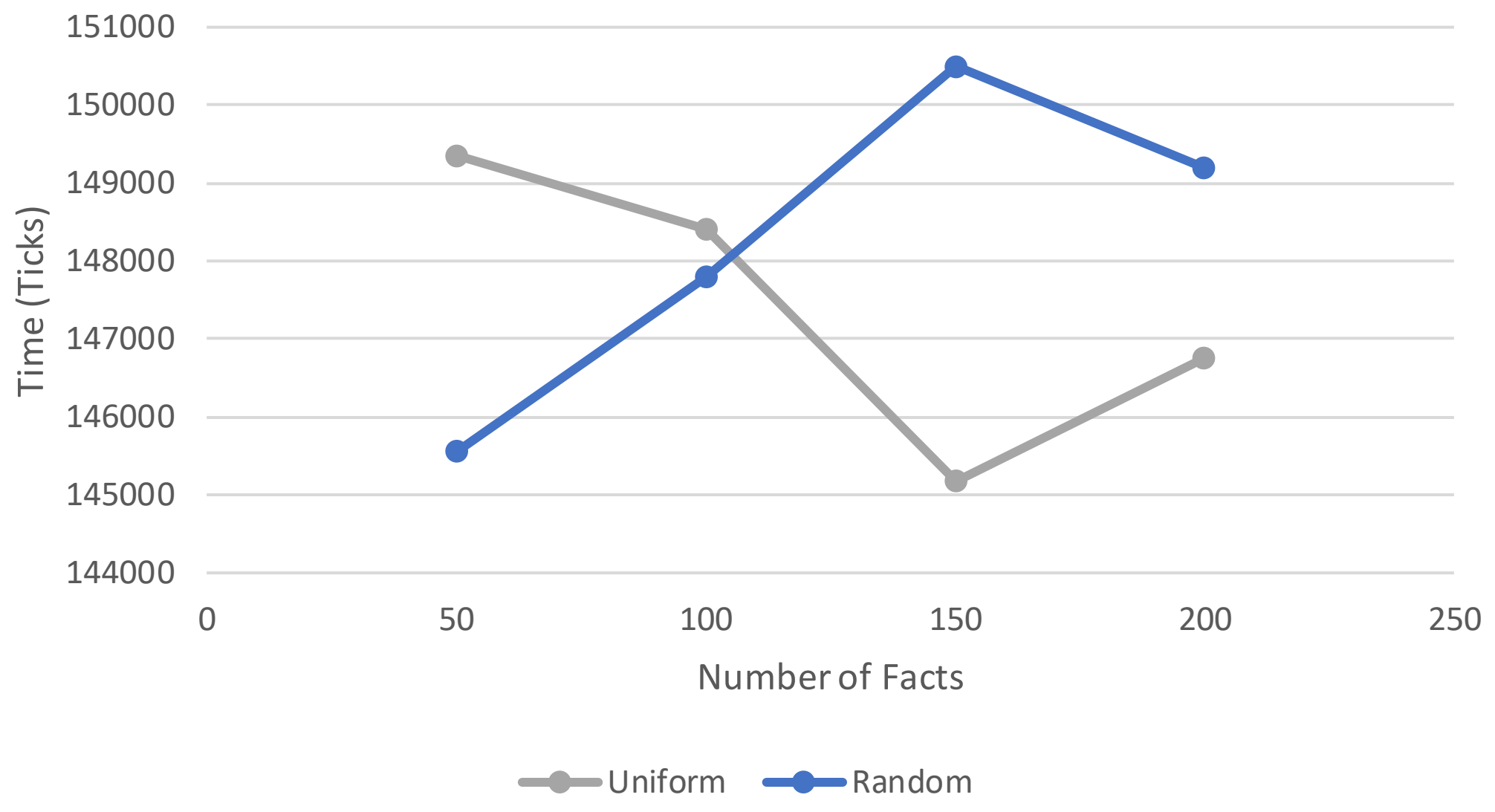

Figure 3. Fact Count by Total Traversal Time.

Figure 4 shows the initial network storage size required by fact count. When the paths and states are added to the network, they are saved as what changed on the initial network state. The initial network size is the size of that full initial state. The size is shown to be positively correlated with the number of facts, which is expected as more facts will take up more storage space.

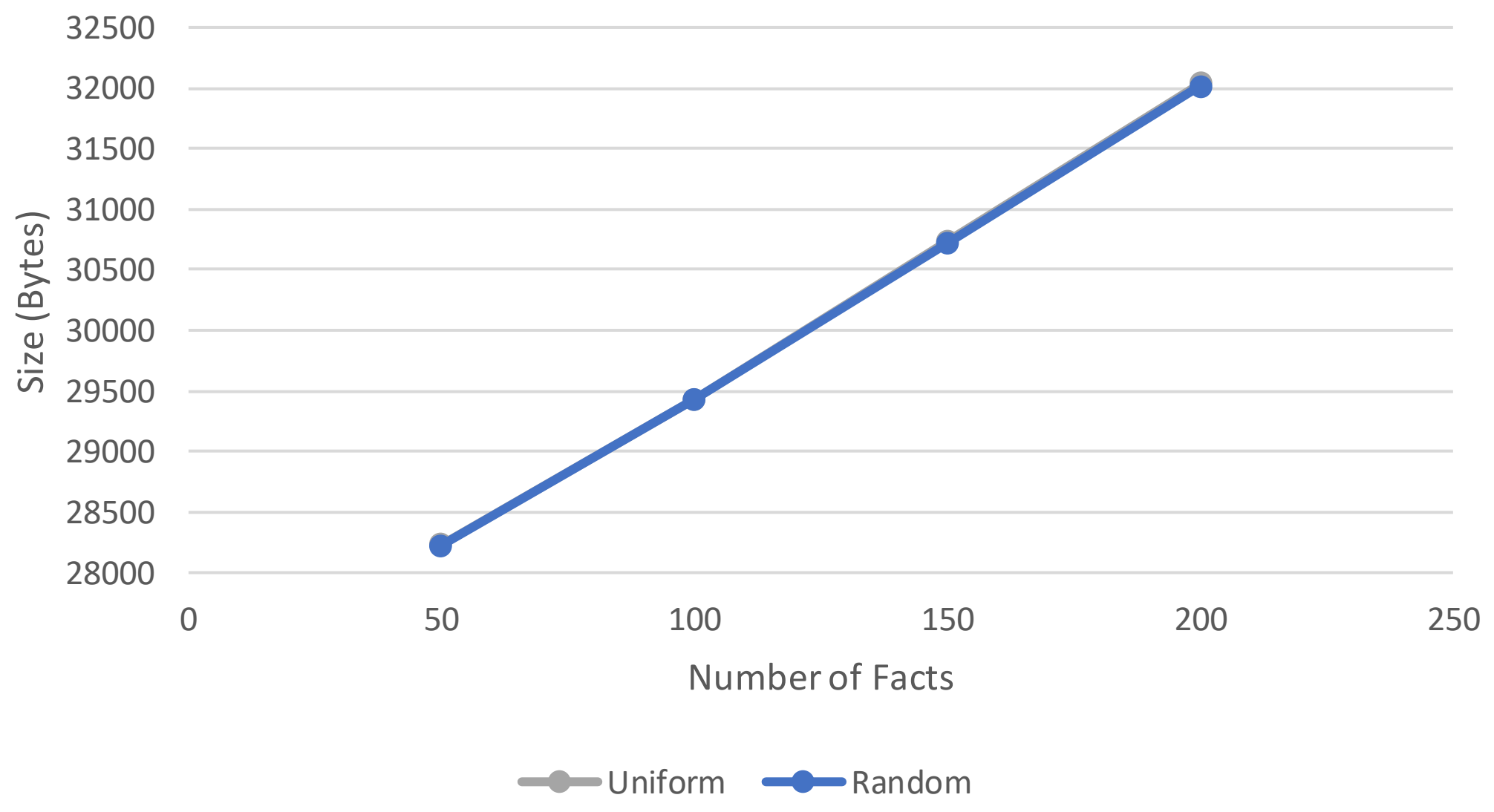

Figure 4. Fact Count by Initial Network Storage Size.

Figure 5 shows the average storage size of just one state change by fact count. New states are stored as changes from the previous one, with the initial state having the entire network saved. The average size of one change state seems to be much higher for the 50 and 100 fact count but then drops quickly for the 150 and 200 fact count. This could be because less facts means it is more likely for rules to have compatible facts and therefore more changes that are then stored.

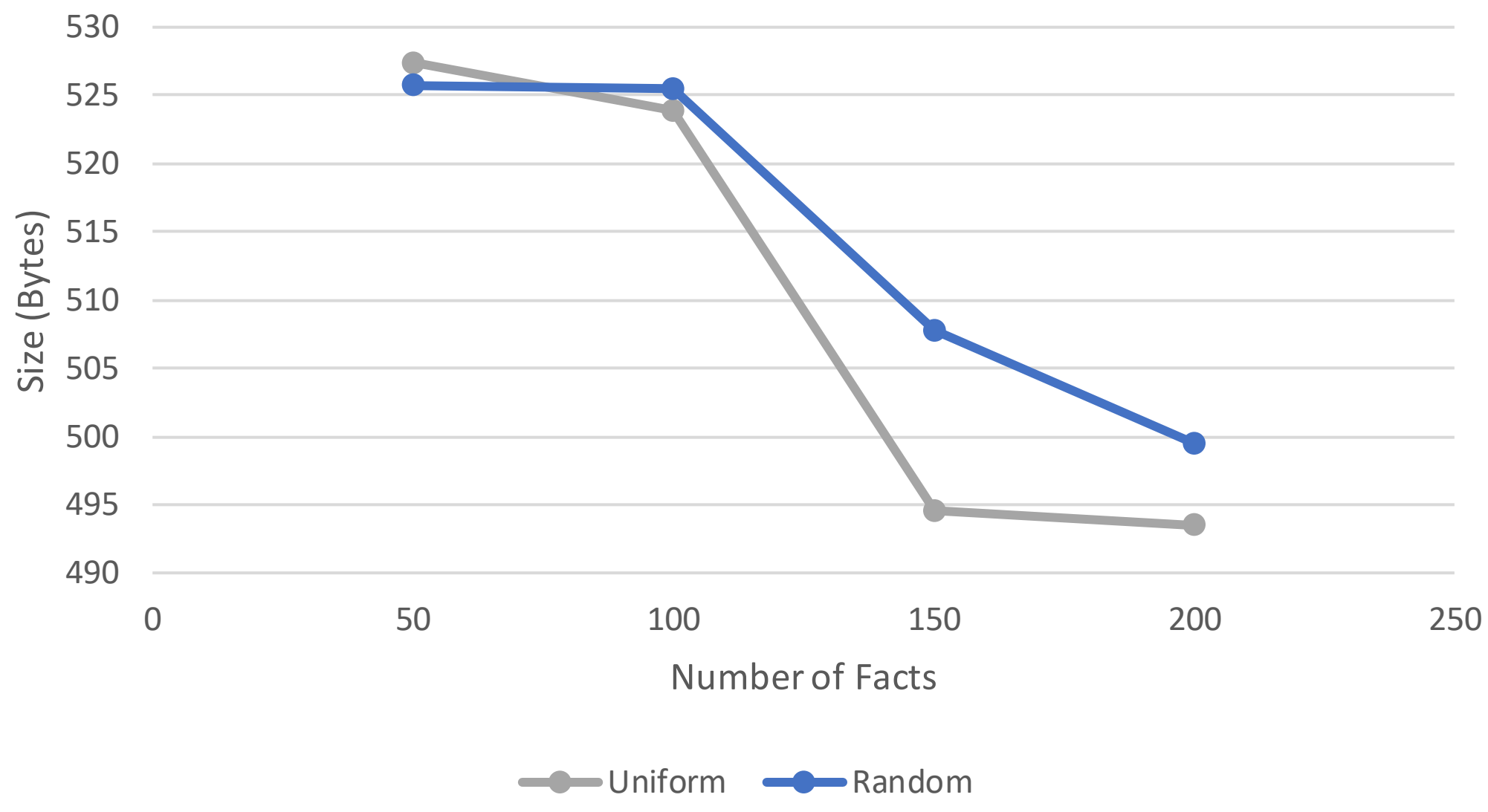

Figure 5. Fact Count by Average Storage Size of One State.

Figure 6 shows the total storage size (in bytes) of the network and all its changes. The total size and fact count are positively correlated with random networks having a slightly larger size than uniform networks, except at the 50 fact count level. This could be because random networks are less likely to have as many changes, until they have a larger amount of facts. This would lead to a smaller storage size at the lower fact count levels for random networks, but cause them to overtake uniform networks at higher fact count levels.

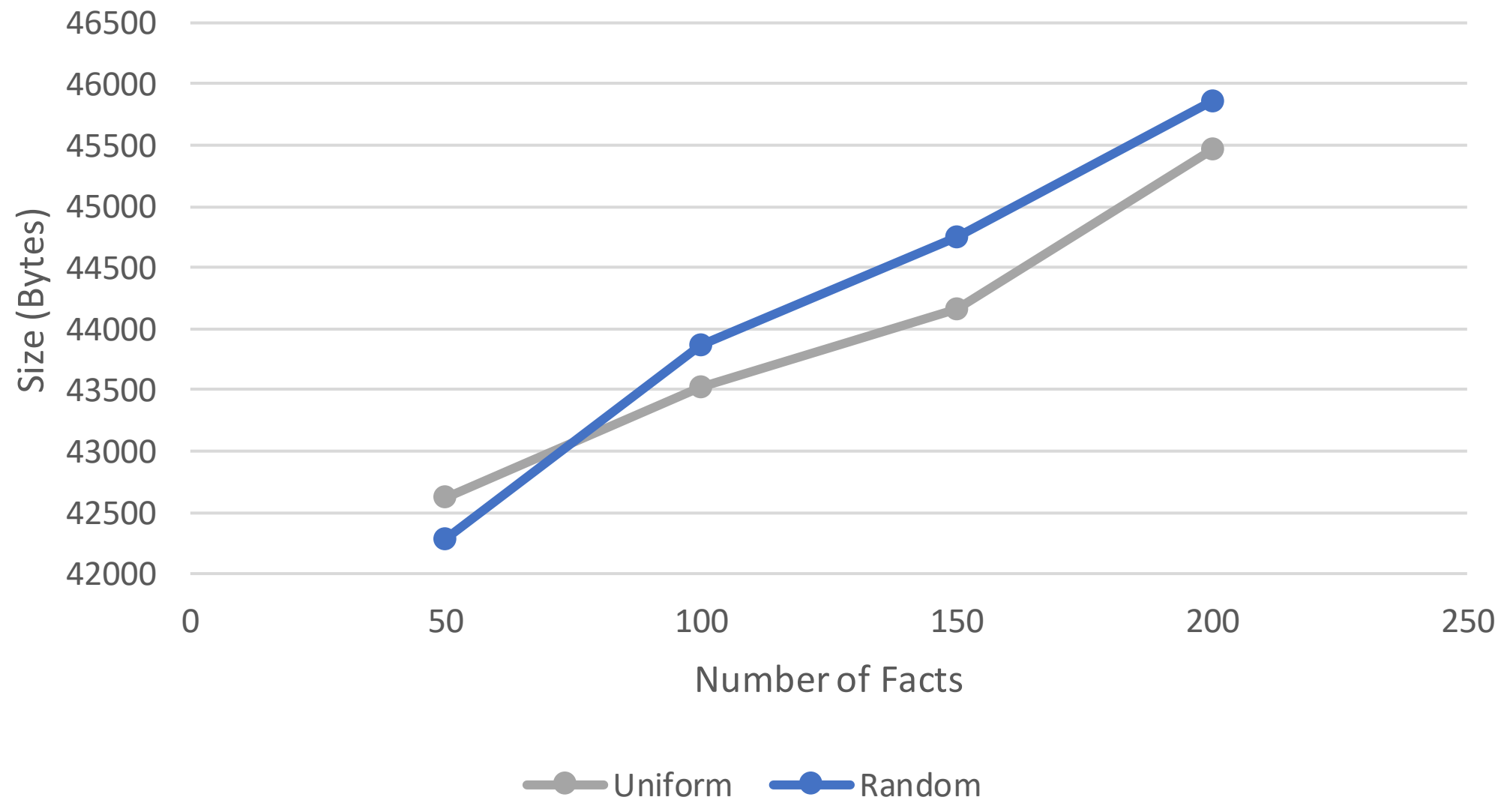

Figure 6. Fact Count by Total Storage Size.

The average length of a valid network path compared with fact count is shown in Figure 7. The average path length stays between 26 and 28 and does not change significantly between fact counts, with one exception. Random networks have a slightly higher path length than uniform networks, in all cases except at the fact count level of 50. This mirrors a common occurrence for other tests, such as shown in Figure 5. One theory for this happening is because random networks have a higher complexity than uniform networks, only if there are 100 or more facts. This would cause there to be a higher path length, on average, for uniform networks only at the fact count level of 50.

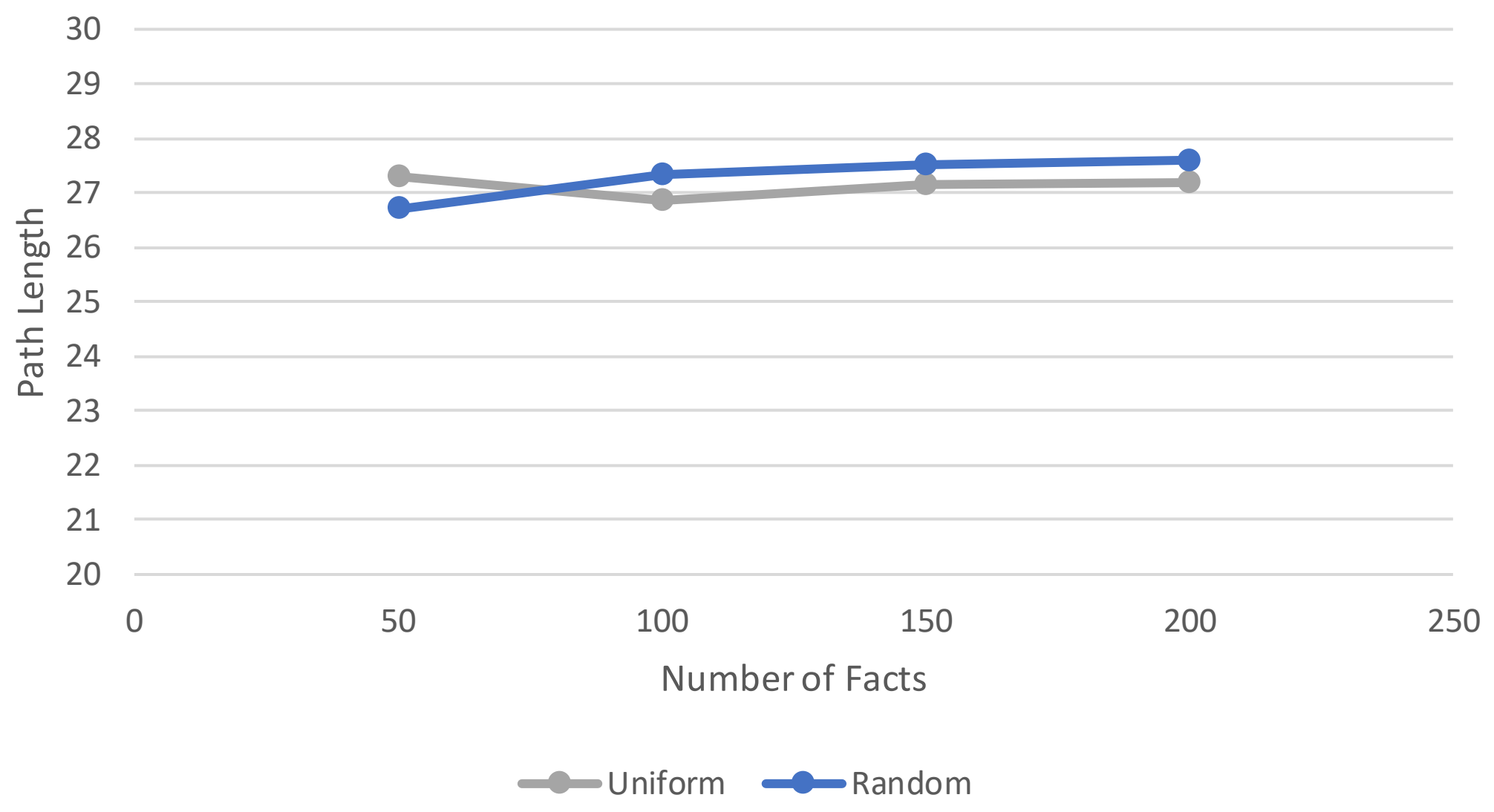

Figure 7. Fact Count by Average Path Length.

***6.2 Rule Count***

The second area of inquiry is regarding how the number of rules in the network affects its performance. Figure 8 shows how the number of rules in a network affects the time needed to find the shortest path. The tick count stays between 150 and 250 for all counts with the only notable difference being that uniform fact assignment takes longer than random fact assignment for a rule count of 100; more testing is required to figure out why this limited deviation exists.

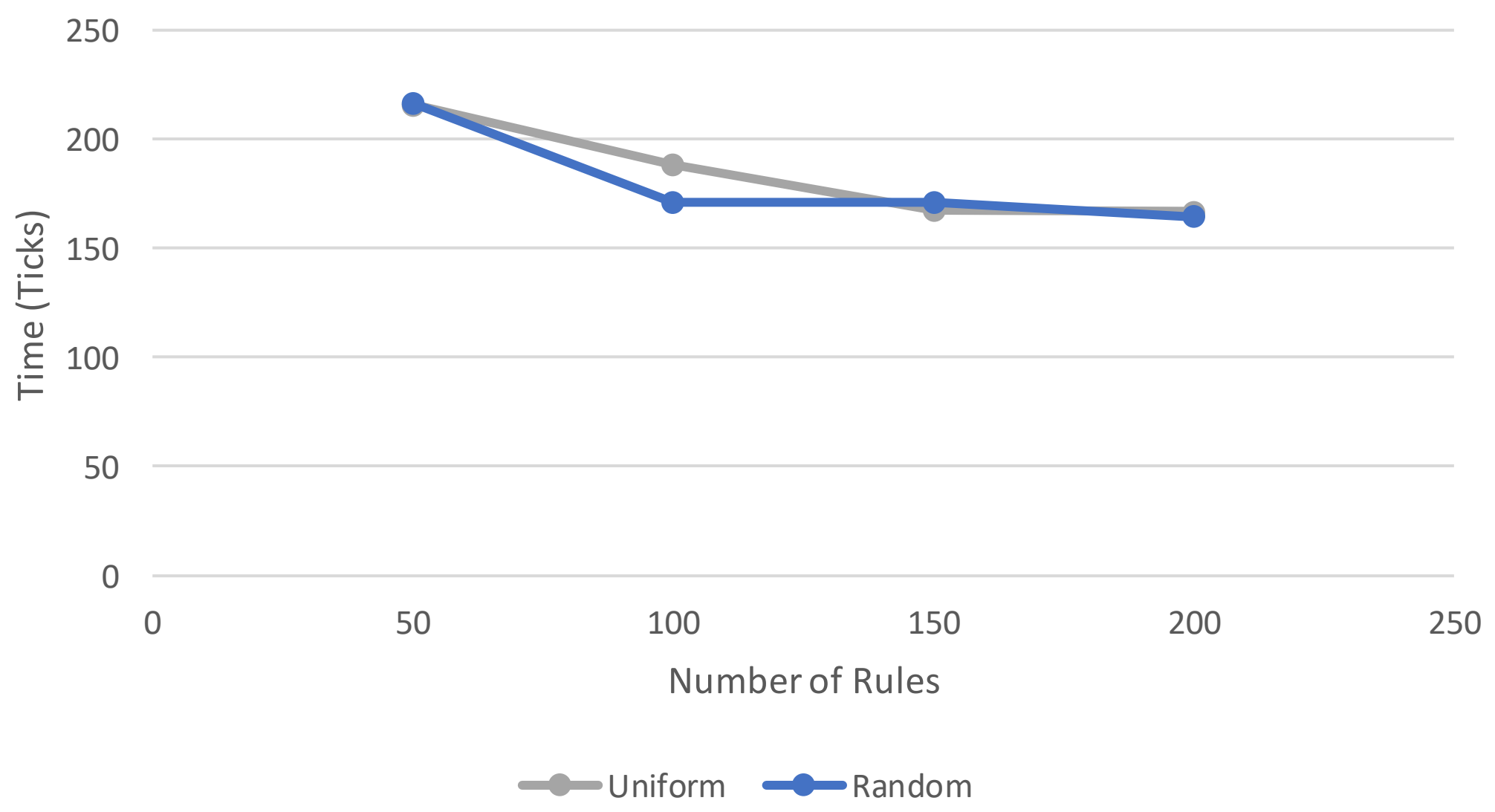

Figure 8. Rule Count by Time to Find Shortest Path.

Figure 9 shows the average time (amount of ticks) needed to reach the next state in the network. Both uniform and random networks are practically the same in terms of the amount of time needed increasing with rule count. This makes sense as more rules will results in more computation time being required.

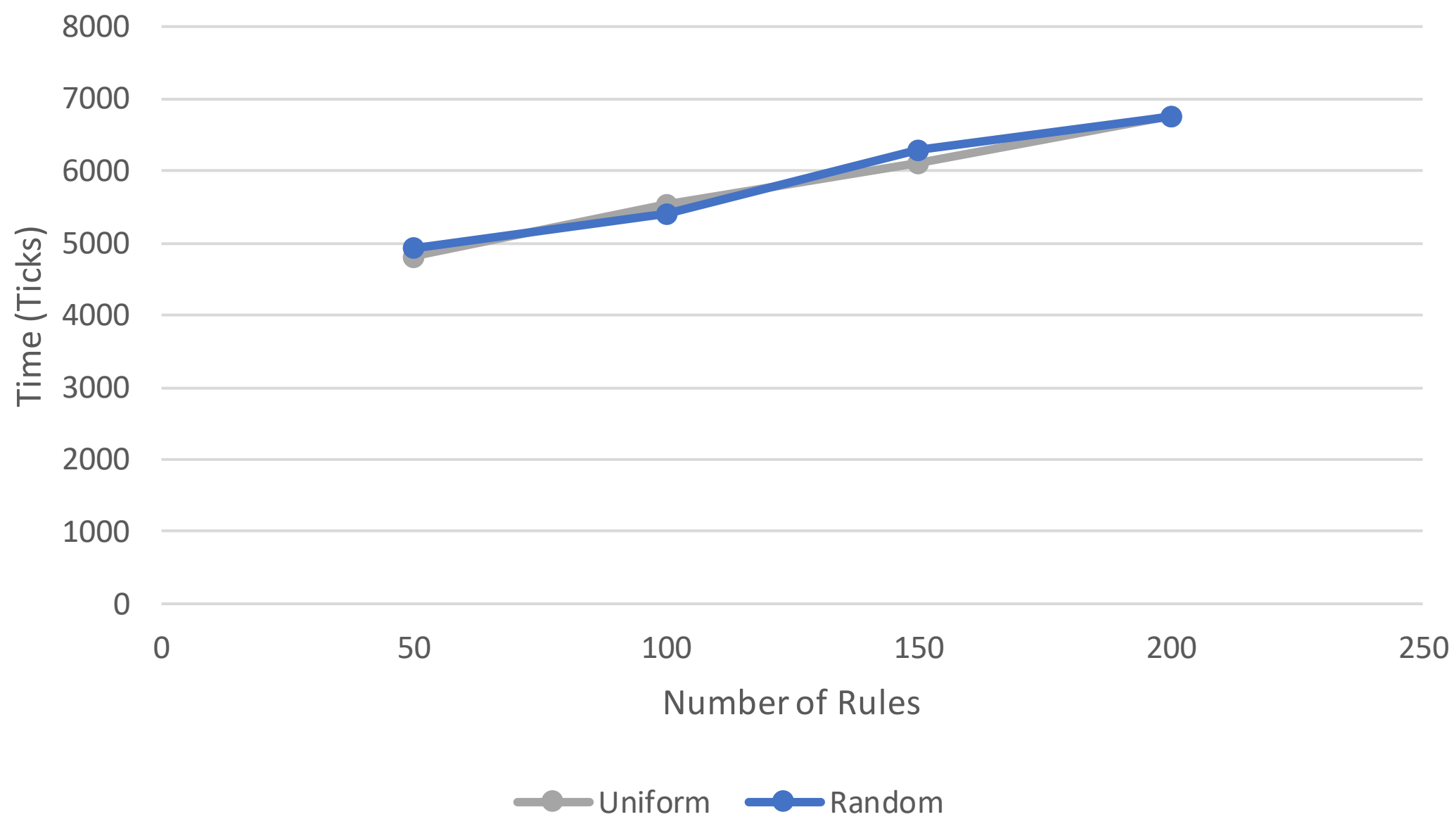

Figure 9. Rule Count by Average Time to Next State.

Figure 10 shows the total traversal time of the network at different rule count levels. The total traversal time required is shown to be positively correlated with the number of rules. This makes sense, as the more rules there are, the more rules the system needs to process during a traversal. This increase results in a longer total traversal time.

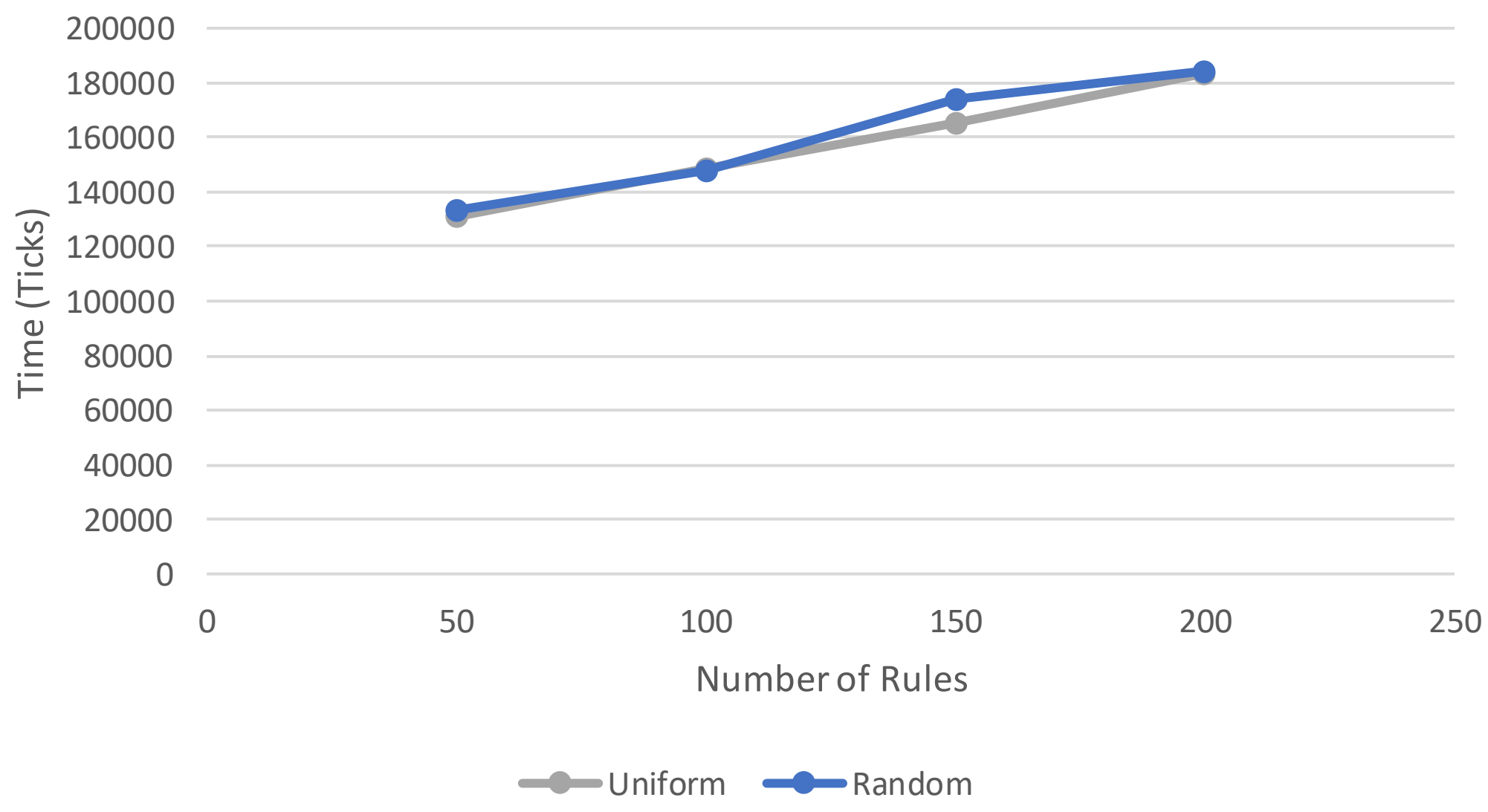

Figure 10. Rule Count by Total Traversal Time.

Figure 11 shows the initial network size compared with the number of rules. Uniform and random networks do not have a discernable difference and the size is positively correlated with the rule count. This makes sense as the more rules there are, the more space is needed to store them.

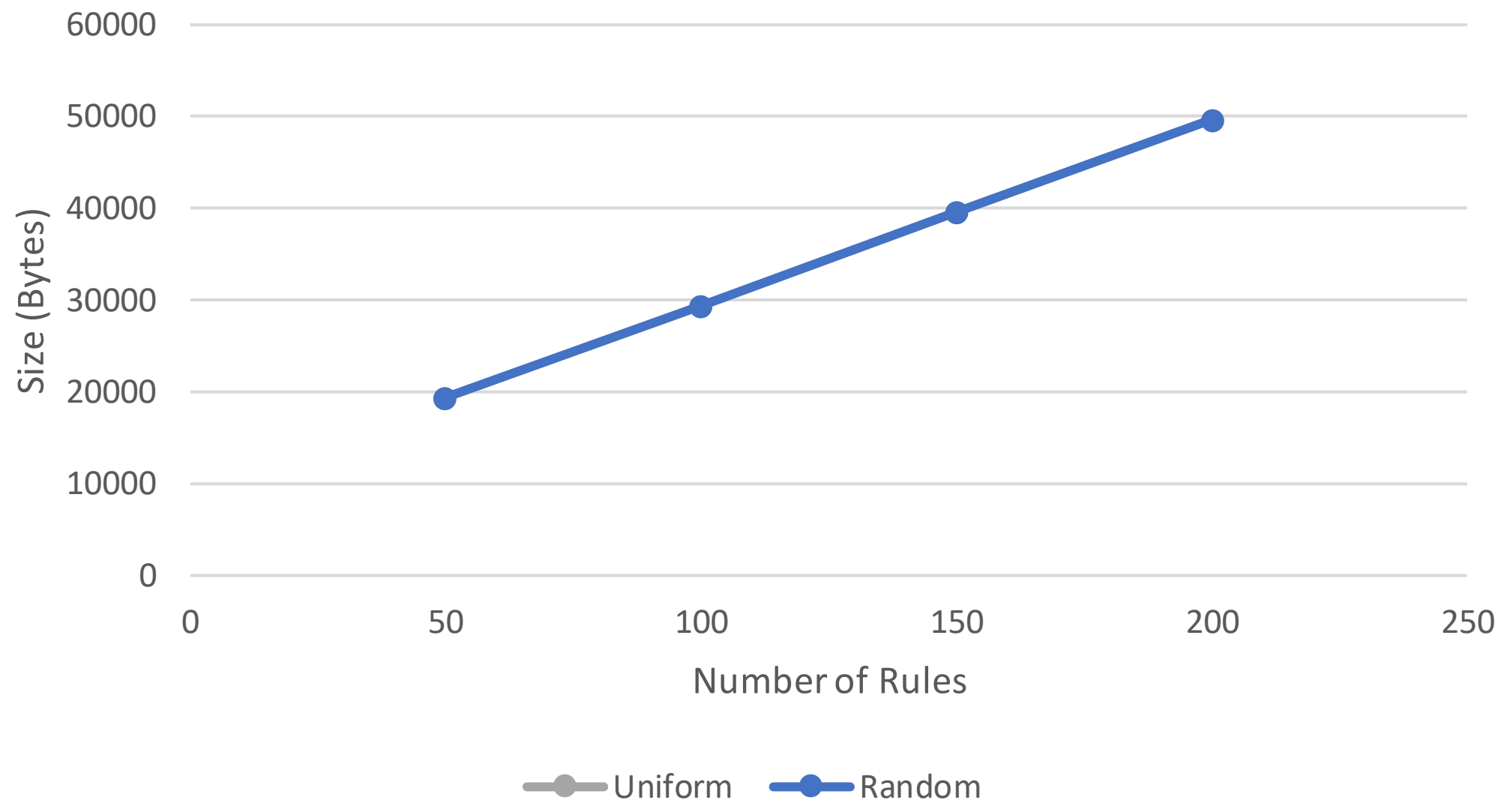

Figure 11. Rule Count by Initial Network Size.

Figure 12 shows the average storage size of one network state by rule count level. There is no discernable difference between the uniform and random networks for this. The graph trends upwards, which makes sense because the size of a state depends of what changes. The more rules there are, the higher the possibility that facts will be changed and then need to be recorded. Therefore, more rules can result in more changes and a larger storage size.

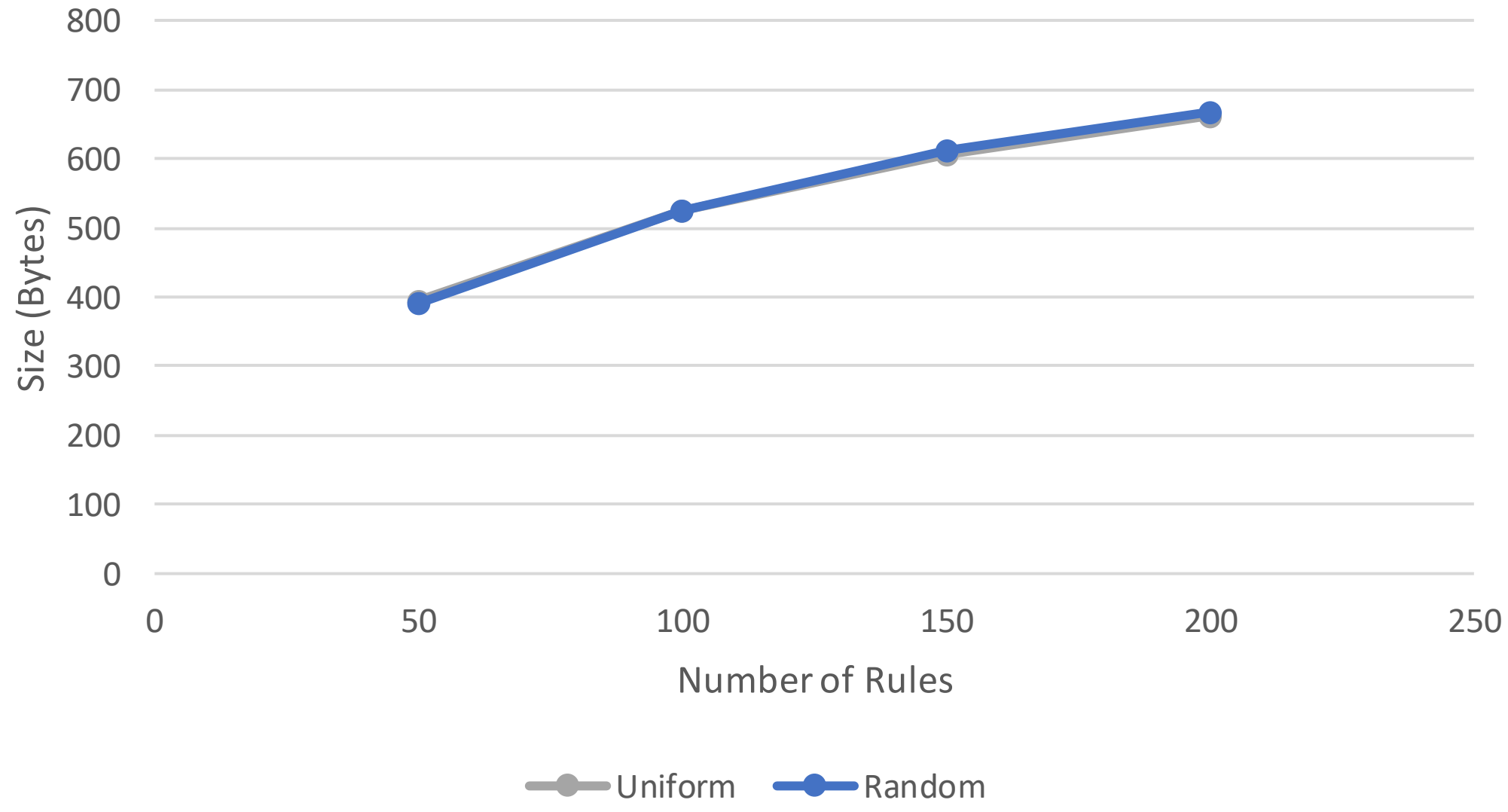

Figure 12. Rule Count by Average Storage Size of One State.

The total storage size of a network compared to the number of rules is shown in Figure 13. There is no discernable difference, for this, between the performance of uniform and random networks. This data is effectively a summation of the data presented in Figures 11 and 12. Both of those graphs have a positive correlation between rule count and size, so, for the reasons expressed in those figures, the total storage size and rule count are also positively correlated.

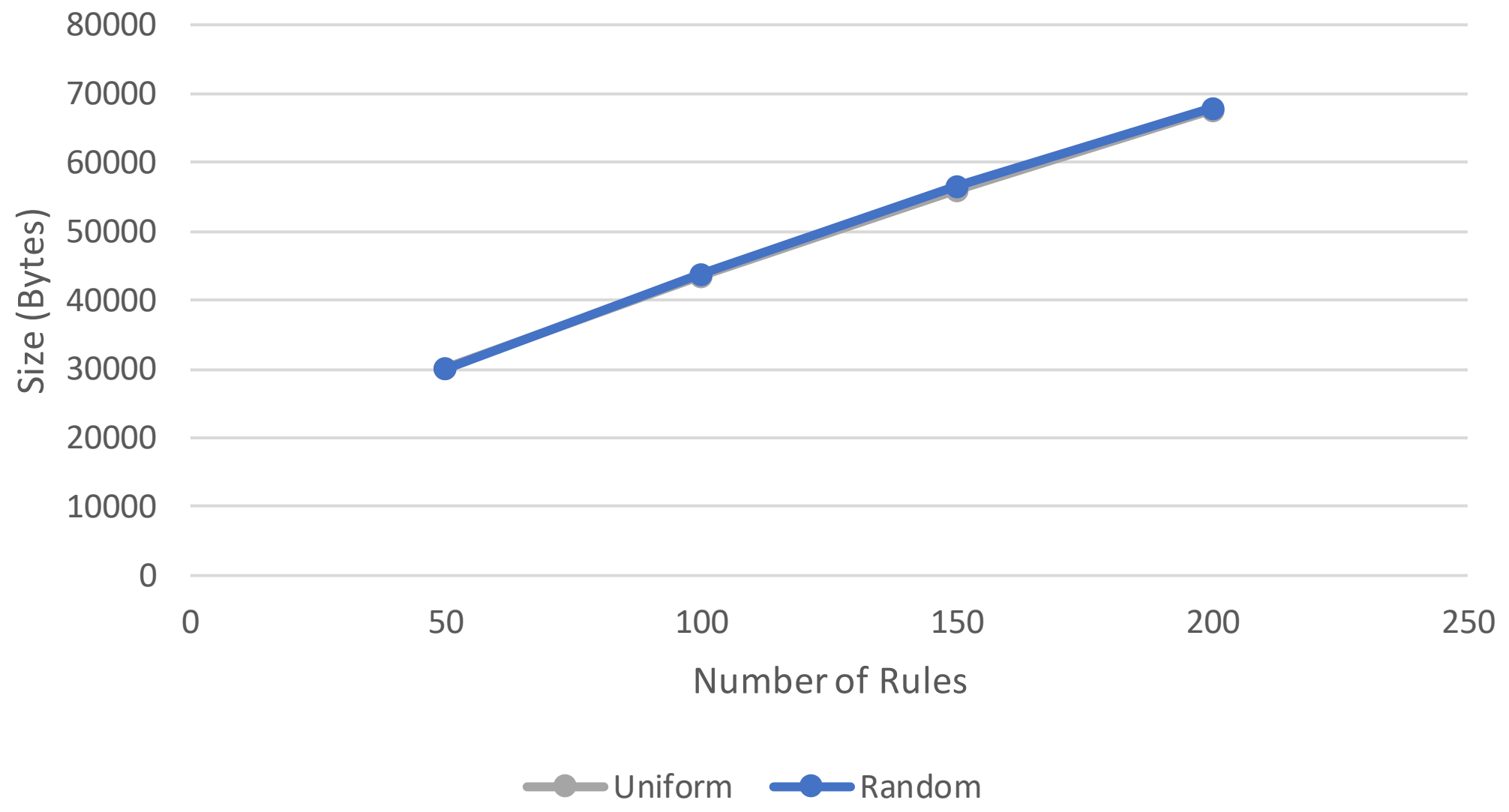

Figure 13. Rule Count by Total Storage Size.

Figure 14 shows the average path length of a network for different numbers of rules. The average path length stays between 26 and 28 facts, across all fact levels evaluated. Uniform and random networks also show nearly the same performance with random networks tending to have a slightly higher path length than uniform networks. This follows a reoccurring trend of random networks having more complexity than uniform networks, when given a larger number of variables.

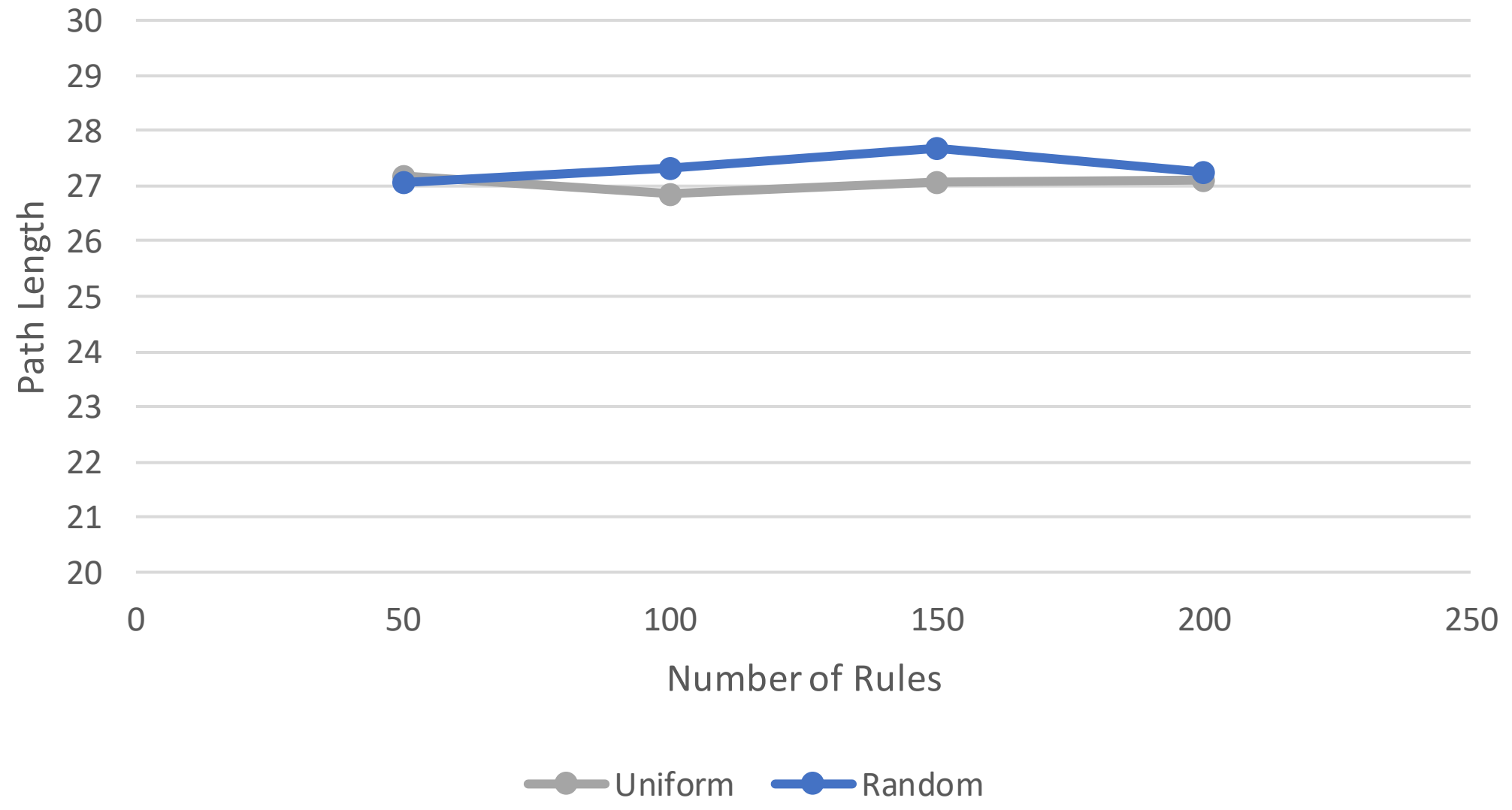

Figure 14. Rule Count by Average Path Length.

***6.3. Link Count***

The third area of analysis is how the number of links in a network affects its results. Figure 15 shows the time required to find the shortest path through a network compared with the number of links. There is a positive correlation between the link count and time required. This makes sense, as the system has to check all paths to find the shortest path. With more links, the system has to check more paths leading to an increase in processing time. Above a link count of 200, uniform networks take slightly more time than random networks. Below a link count of 200, random networks take slightly more time than uniform networks. More analysis is required to explain this.

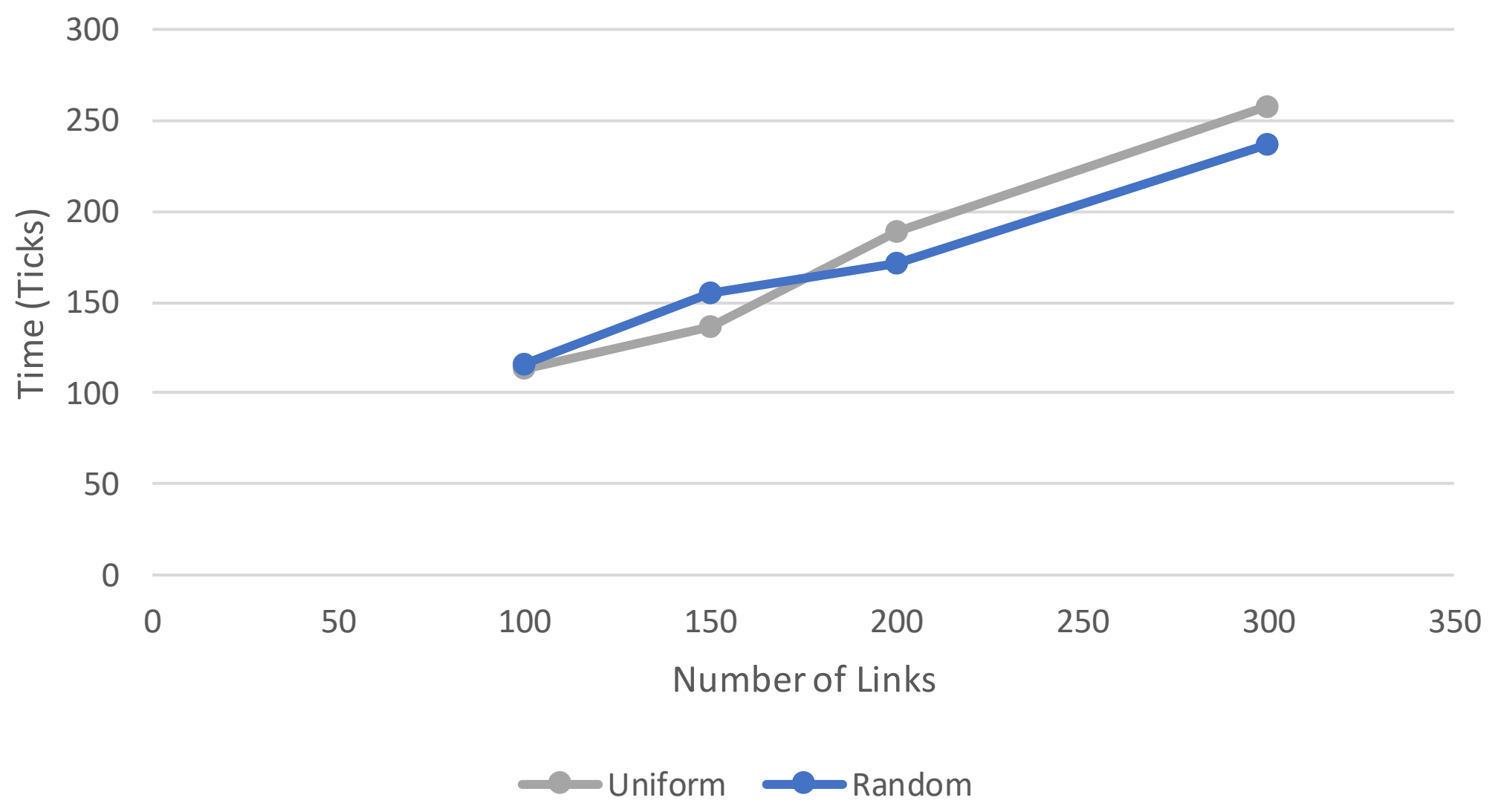

Figure 15. Link Count by Time to Find Shortest Path.

Figure 16 shows the average time required to reach the next network state, compared to the link count. Uniform and random networks vary greatly from each other in this measurement; however, in general, the time required for a network of 150 links is much lower than for 50 links. It then increases for higher link count levels. The biggest difference between uniform and random networks is in the range from 150 to 200 links. Random networks only show a slight increase in time, while uniform networks show a large increase. Much like in Figure 15, uniform and random networks switch which one takes more time at 200 Links. Uniform Networks take more time at 200 links and above and random networks take more time below 200 links. The largest difference between two datapoints at the same link count level is just under 300 ticks. More testing is required to see if this is just a relatively small difference or if this particular data is due to an unknown cause.

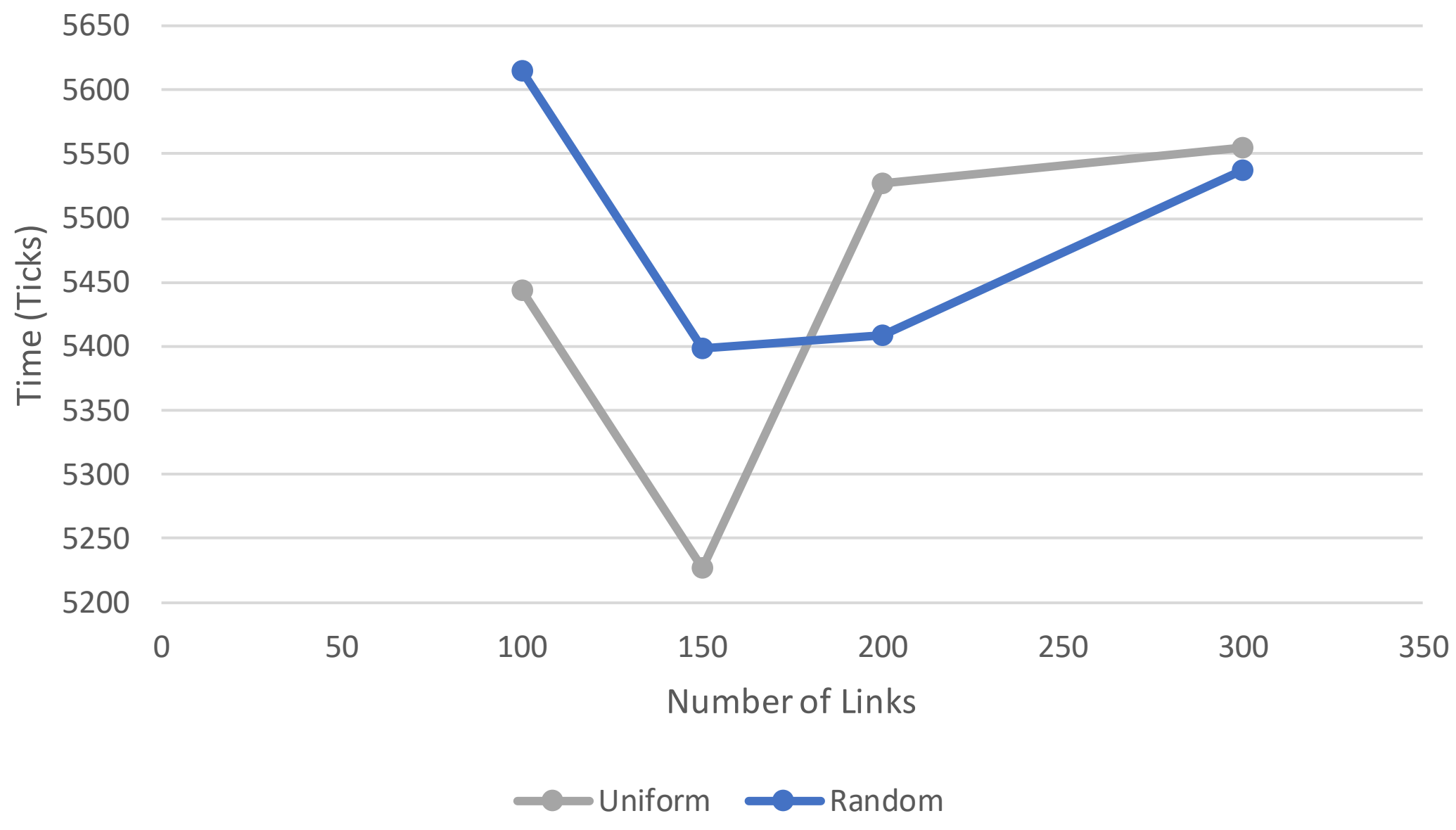

Figure 16. Link Count by Average Time to Next State.

Figure 17 shows the total traversal time compared with the number of links. This figure has many of the same features of Figure 16. For the uniform network, the time starts at a higher level and then it dips at a link count of 150, and then climbs again. The random network has a less pronounced dip in time, which may be because of the randomness of the link distribution normalizing the data. More testing is required to understand why, specifically at a link count of 150, traversal is more efficient.

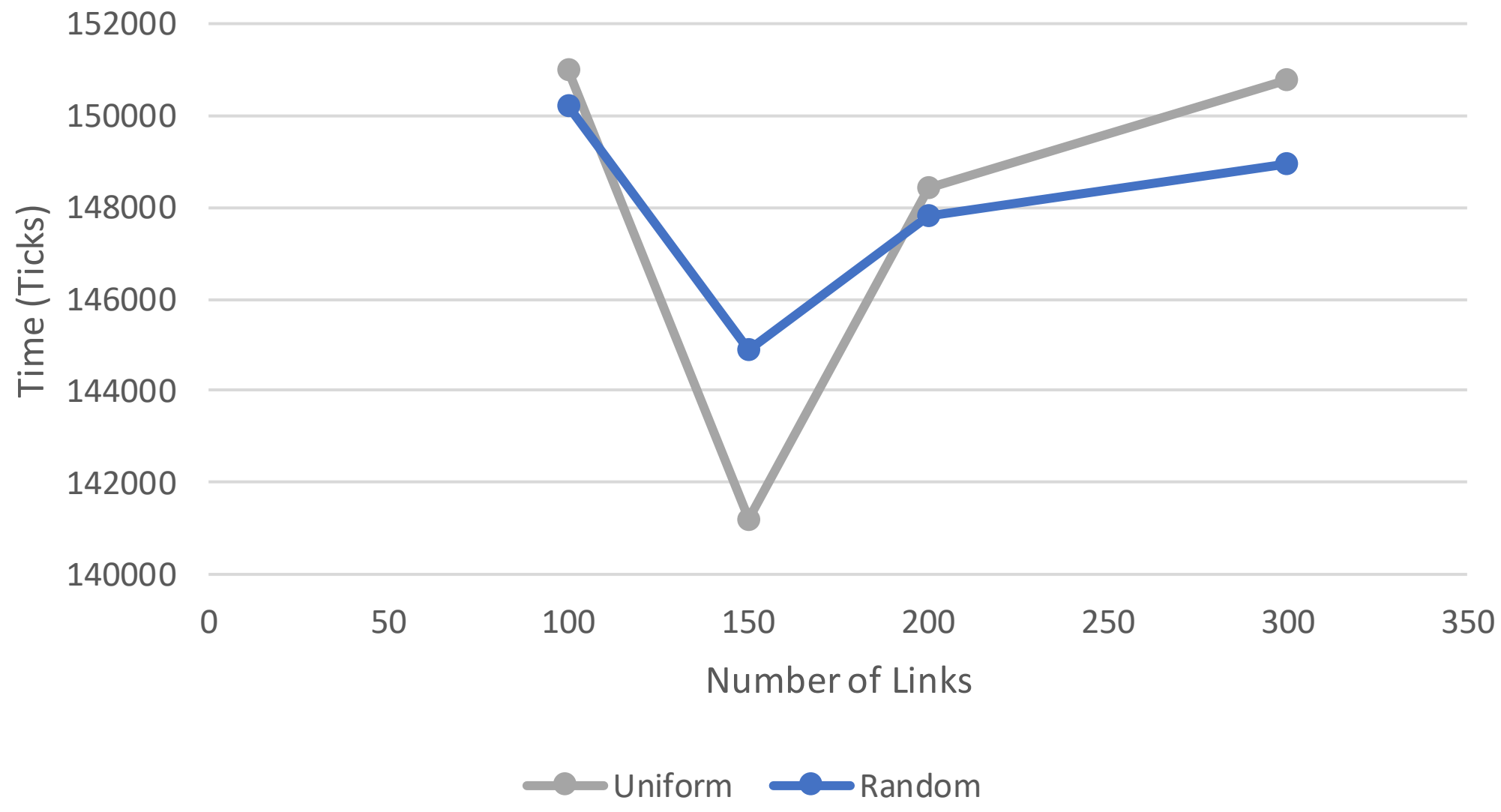

Figure 17. Link Count by Total Time to Next State.

Figure 18 shows the initial storage size required compared with the number of links. The size and link count are positively correlated, as expected. This is because more links take more space to store in the initial network. There is no discernable difference between the size requirements for uniform and random networks.

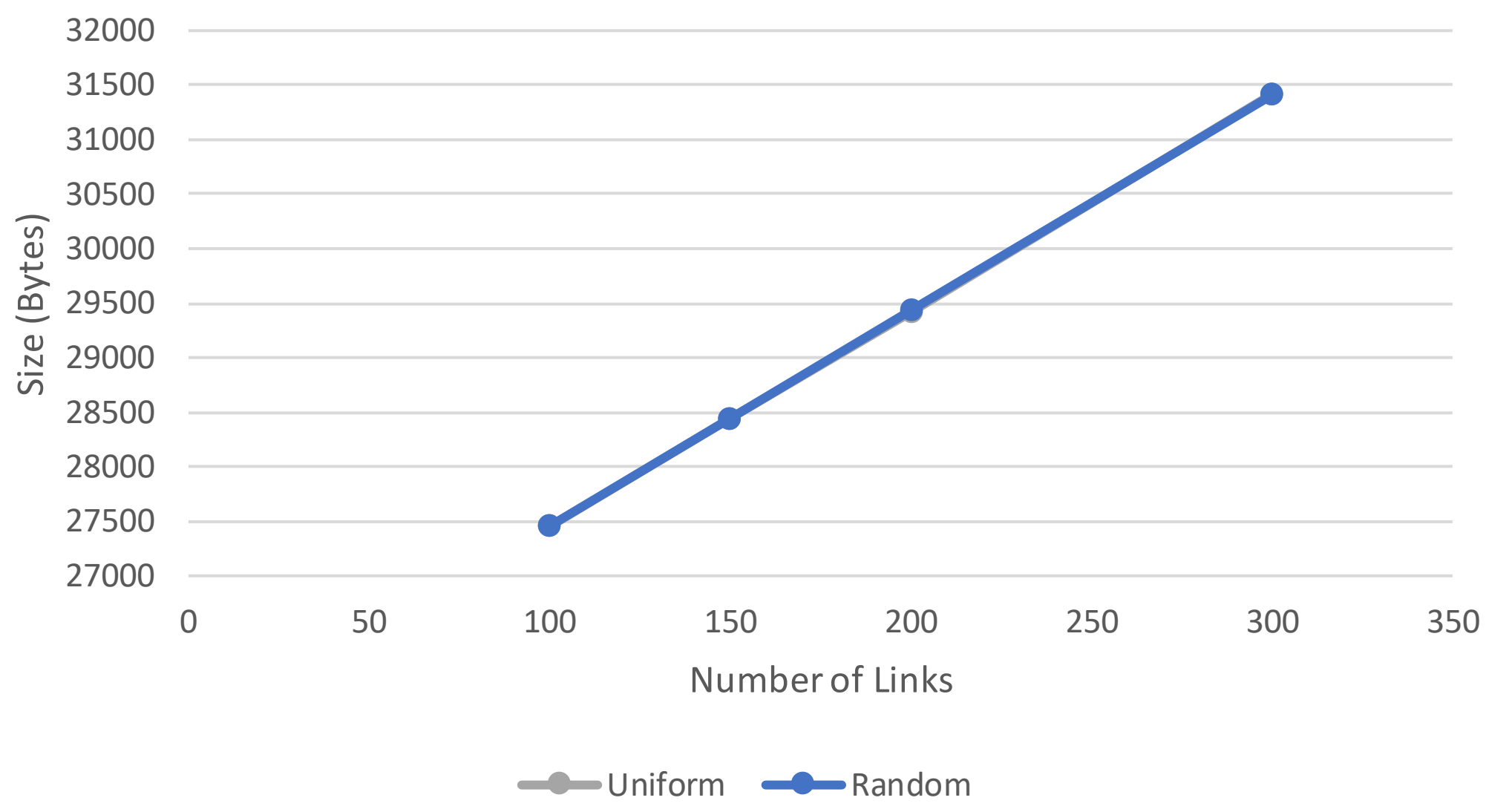

Figure 18. Link Count by Initial Storage Size.

Figure 19 shows the average storage size of one state in the network compared with the number of links. In general, the size peaks at around 520 bytes at 200 Links. At most, there is a 20-byte difference between the highest and lowest datapoint which means the shape of this data could be due to random variation. The rough similarities with Figure 17 mean that there may be a similar cause to this data pattern, as it makes sense that state storage size and state traversal time may mirror each other. More testing is required to determine this cause.

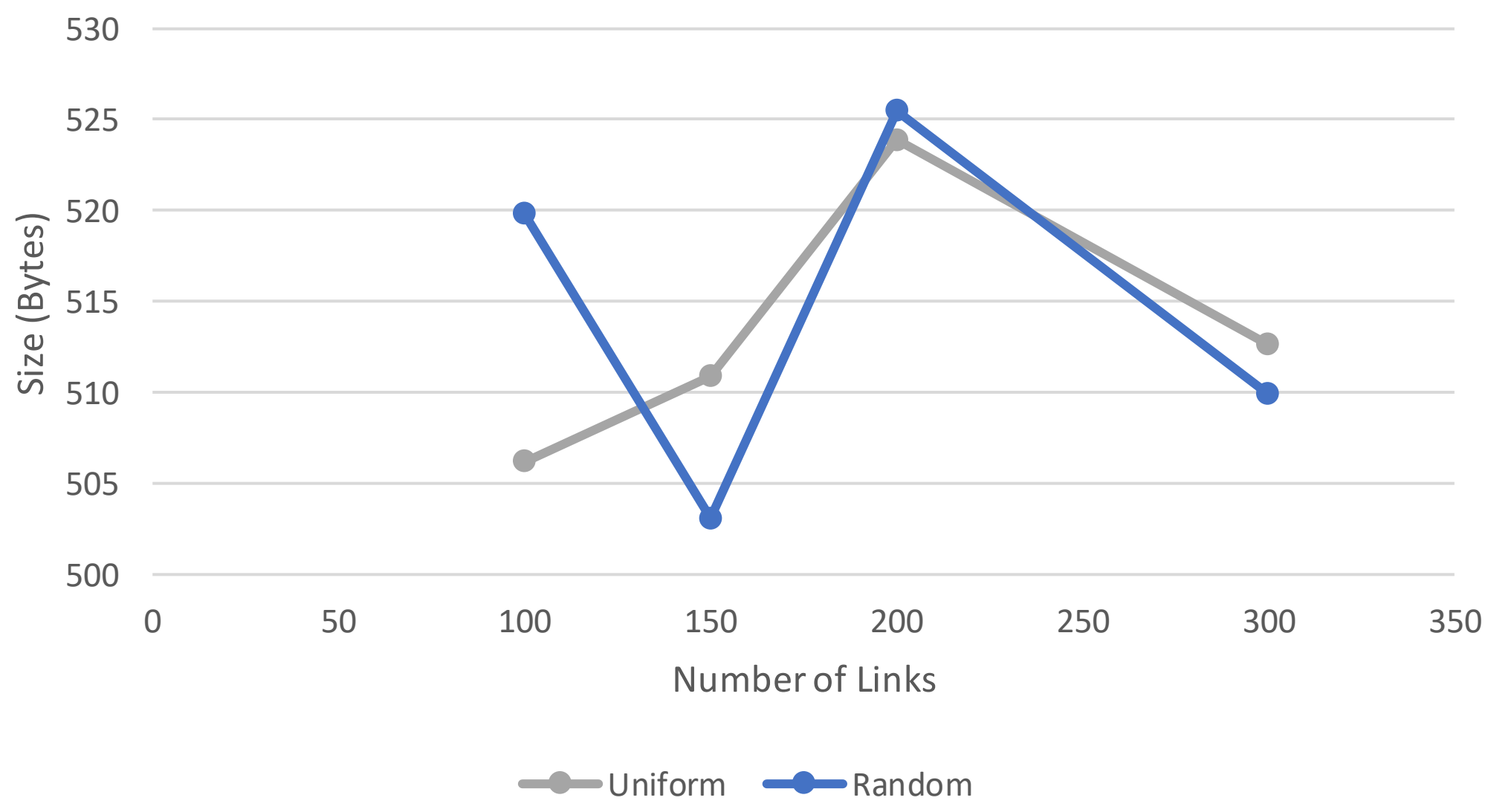

Figure 19. Link Count by Average Storage Size of One State.

Figure 20 shows the total storage size of a network compared with the number of links. There is a positive correlation between the size and link count. This makes sense as the higher the link count, the higher the initial storage size will be and the more potential changes there can be, leading to a larger state size as well.

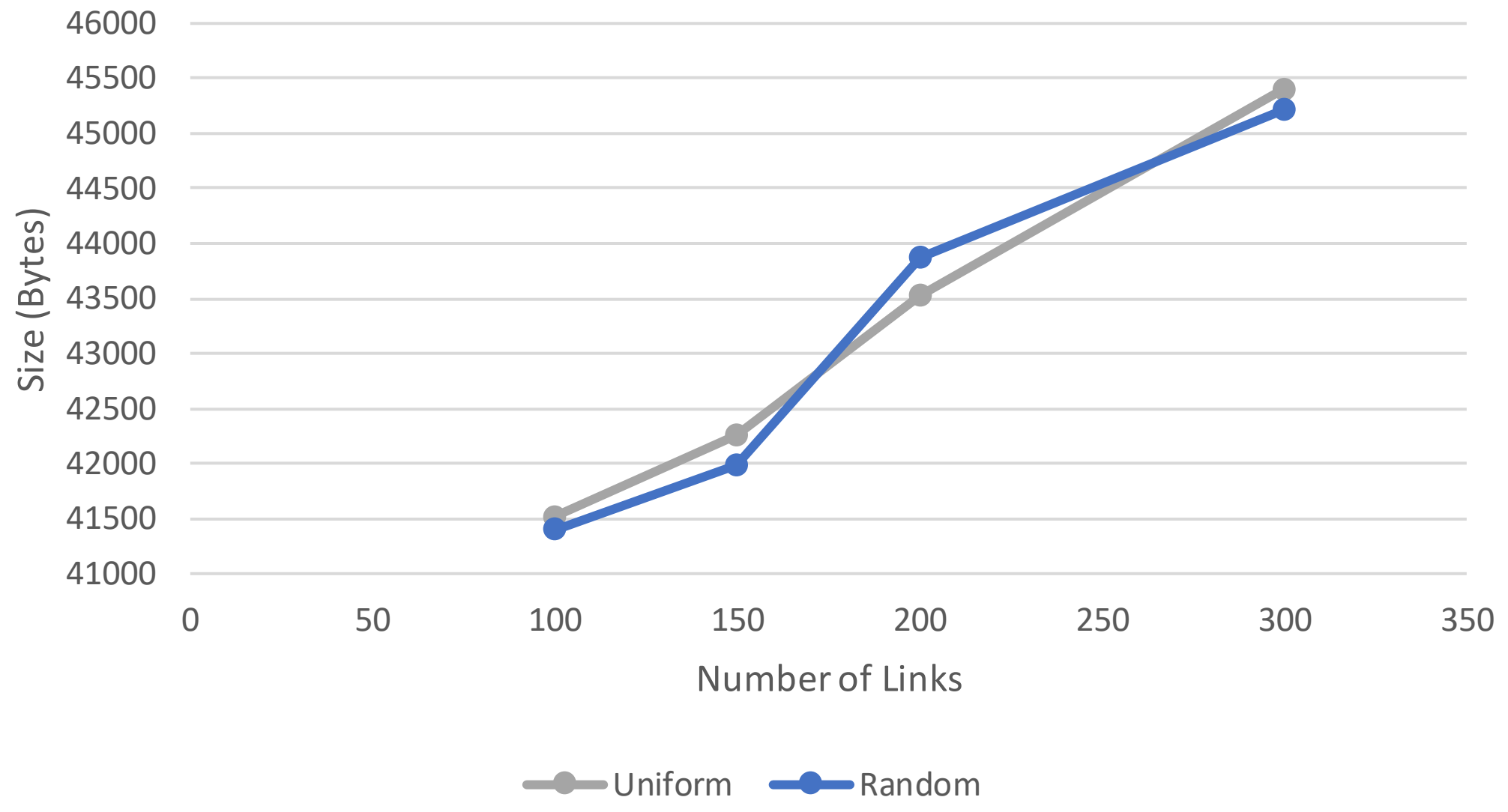

Figure 20. Link Count by Total Storage Size.

Figure 21 shows the average path length of the network compared to the number of links. The average path length stays between 26 and 28 facts. The uniform and random networks have nearly the same average path length, except at the link count level of 100. Here, the uniform networks have nearly a 1.0 difference, above random networks. This could be because uniform networks, which make sure each node gets around the same amount of links, would lead to a longer path being found, on average, as random networks could have nodes which only have one link and are, thus, less likely to be a part of a chain.

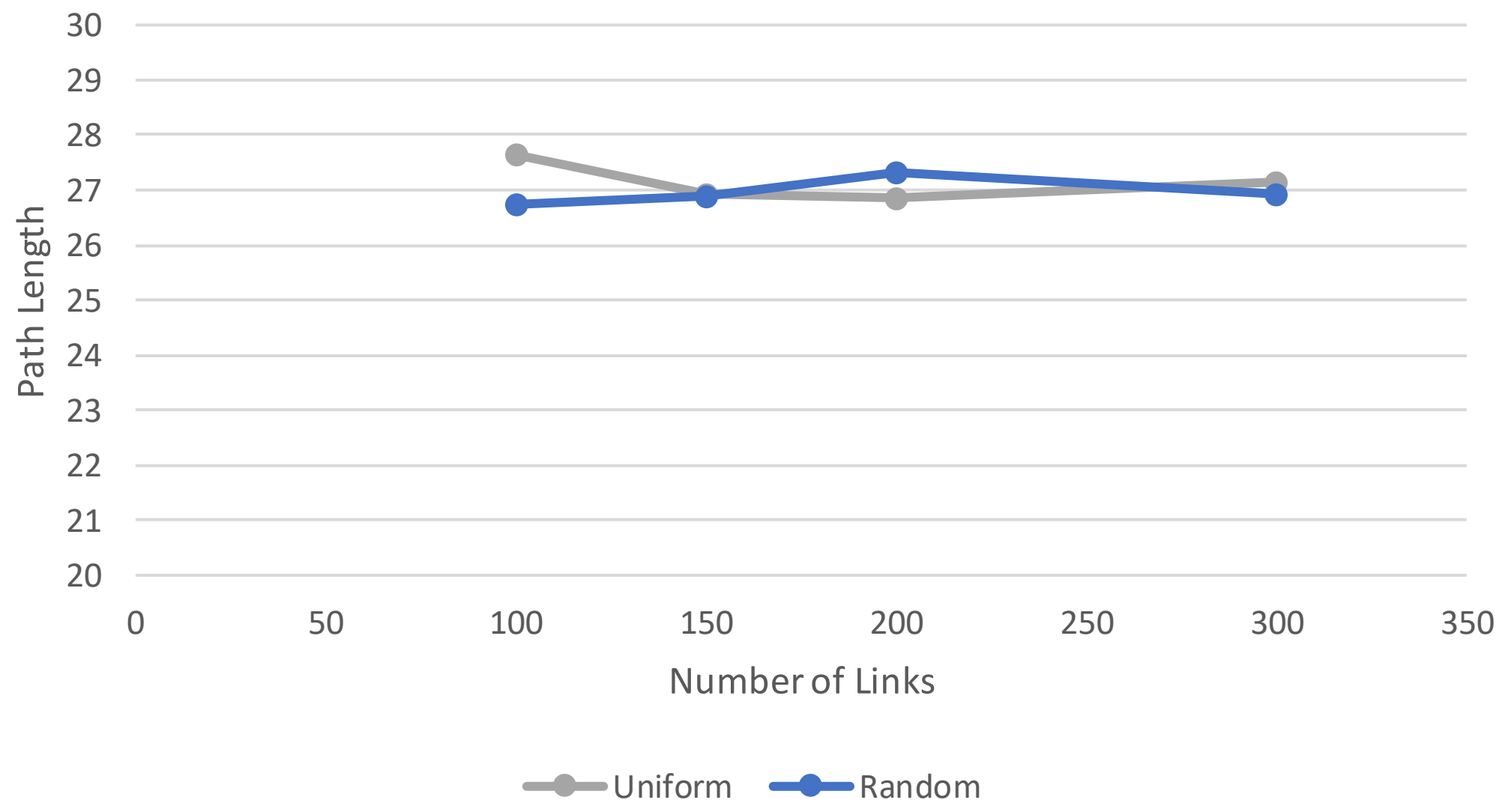

Figure 21. Link Count by Average Path Length.

### ***6.4. Container Count.***

The fourth area of analysis is how the number of containers in the network affects its performance. Figure 22 shows the time required to find the shortest path through a network, compared with its container count. These two variables are positively correlated because the more containers there are, the more paths are able to be created. The more paths there are, the longer it takes to search them all, which leads to a longer time to find the shortest path. Uniform and random networks do not significantly differ, except at the container count of 150, where random networks take about 50 ticks longer than uniform networks, on average. More testing is required to determine whether the 150 containers level for random networks represents a notable departure from the linear correlation.

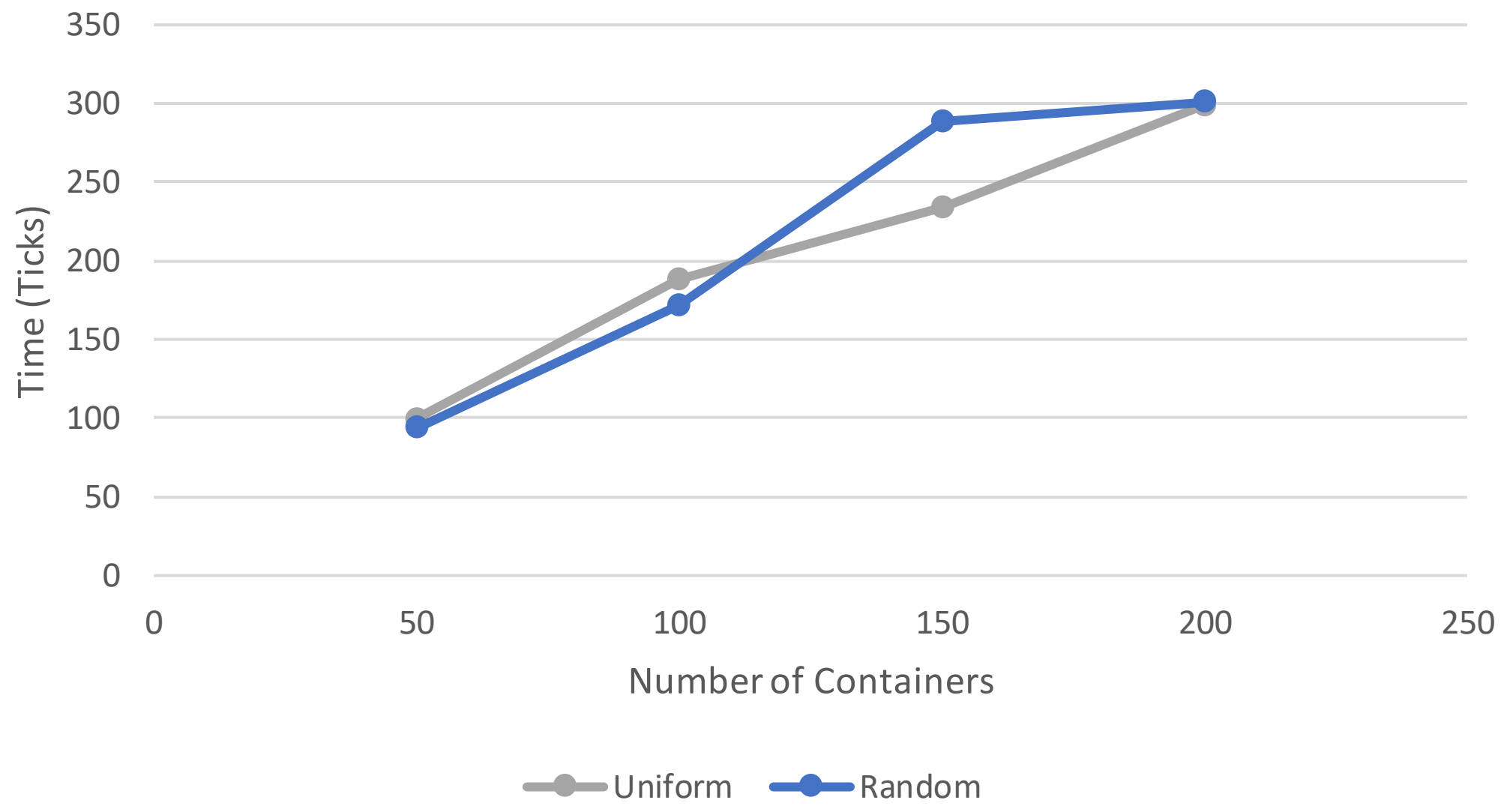

Figure 22. Container Count by Time to Find Shortest Path.

Figure 23 shows the average time required (in ticks) to reach to the next network state, as compared to the number of containers. For random networks, there is a roughly positive exponential correlation between the two variables. This makes sense, as the time to reach the next state largely depends on the complexity of the network. This is, in part, dependent on the number of containers. While uniform networks roughly follow the pattern of random networks, at a container count of 50 the average time to reach the next state of uniform networks is almost as long as at its tested max of 200 containers. It is unclear why this pattern exists.

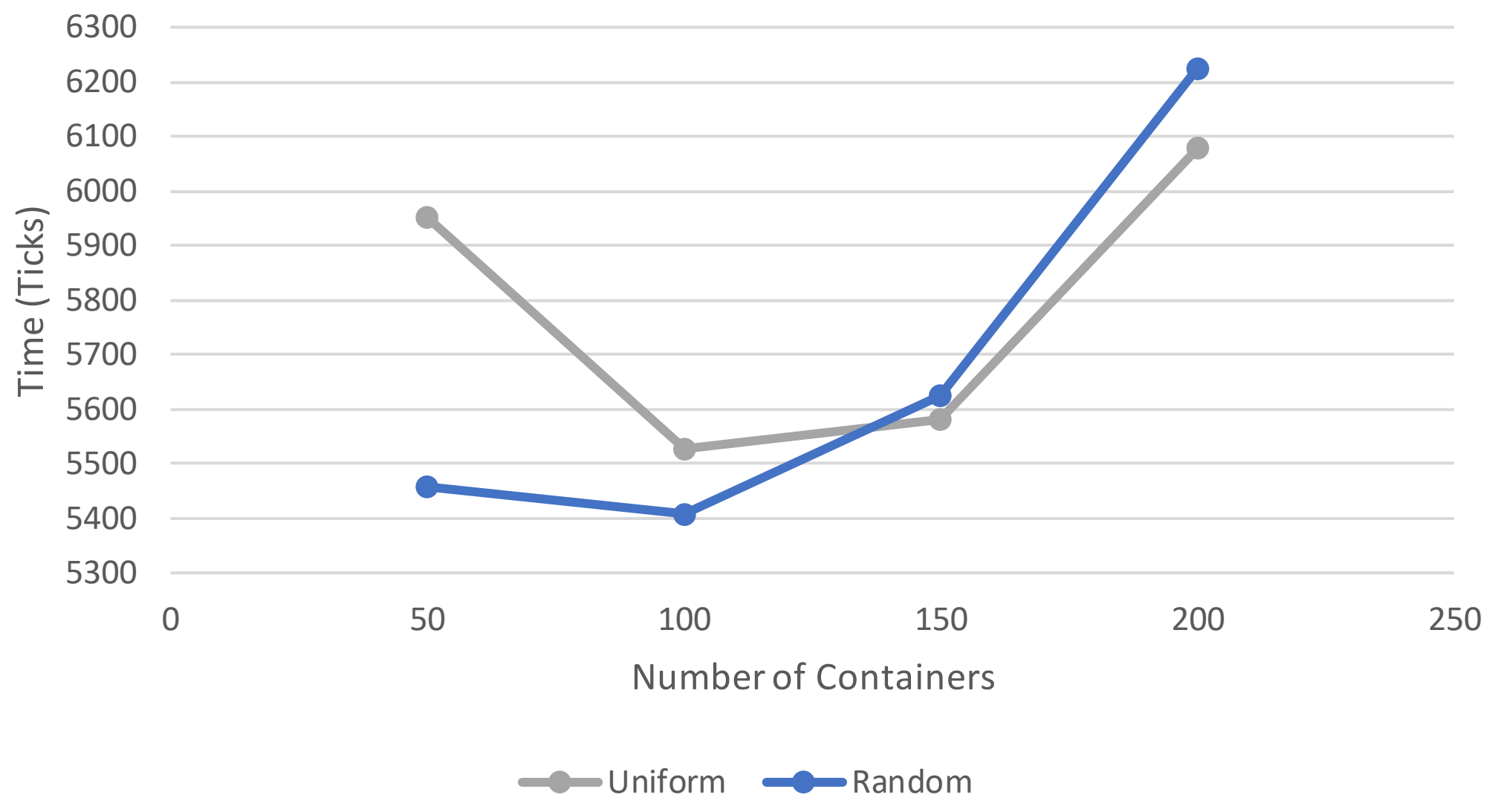

Figure 23. Container Count by Average Time to Next State.

Figure 24 shows the total traversal time of the network compared to the number of containers. There is no discernable time difference between the performance of uniform and random networks for this test. Traversal time is positively correlated with the number of containers. This makes sense, as more containers will result in a longer path (which is also shown in Figure 28). A longer path results in a longer traversal time.

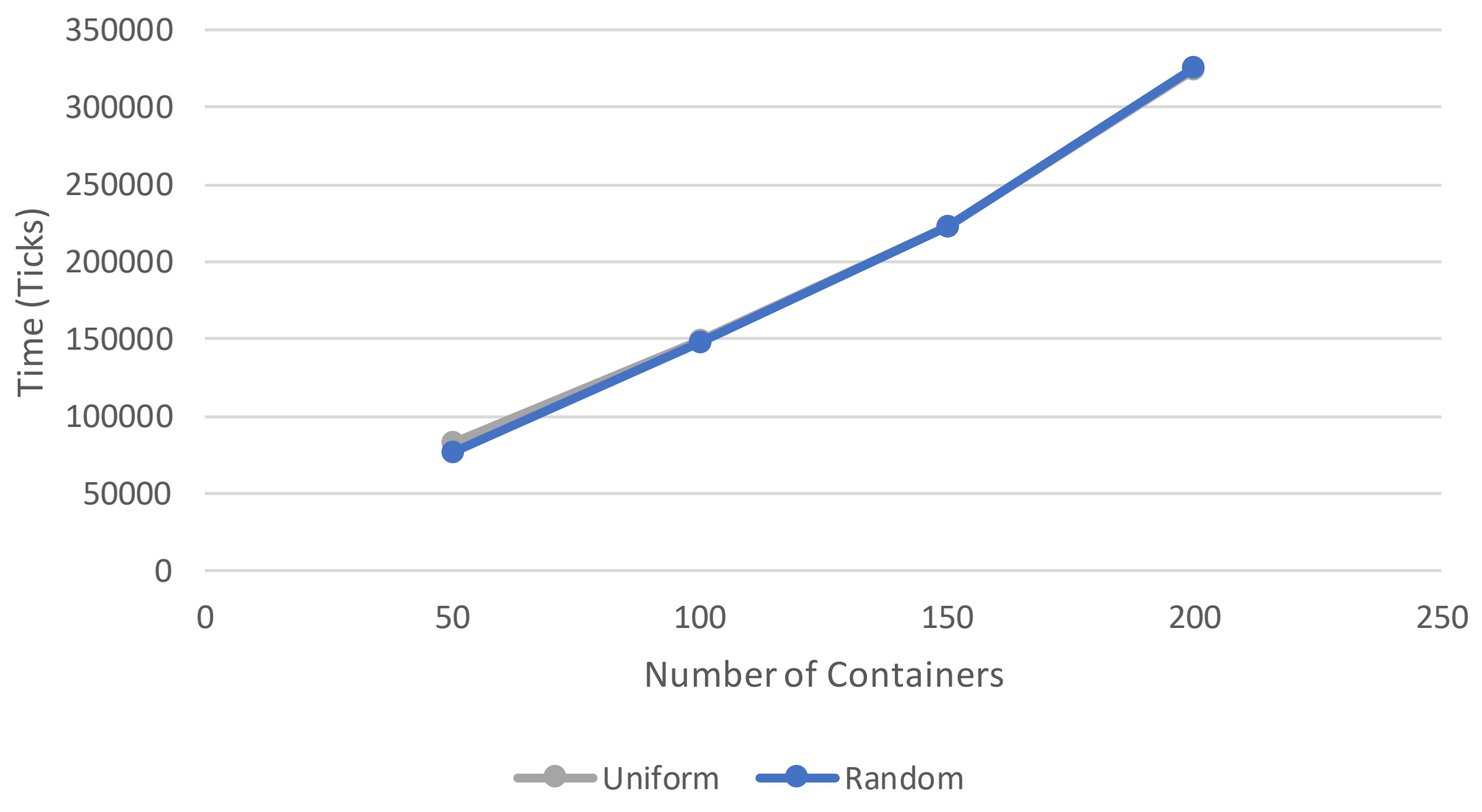

Figure 24. Container Count by Total Traversal Time.

The initial size of the network in bytes is compared with the number of containers in Figure 25. There is no discernable difference in size between the uniform and random networks. The total size is positively correlated with the container counts. This is explained by the storage needed being a relatively fixed size for the containers. More containers will result in more storage space being needed.

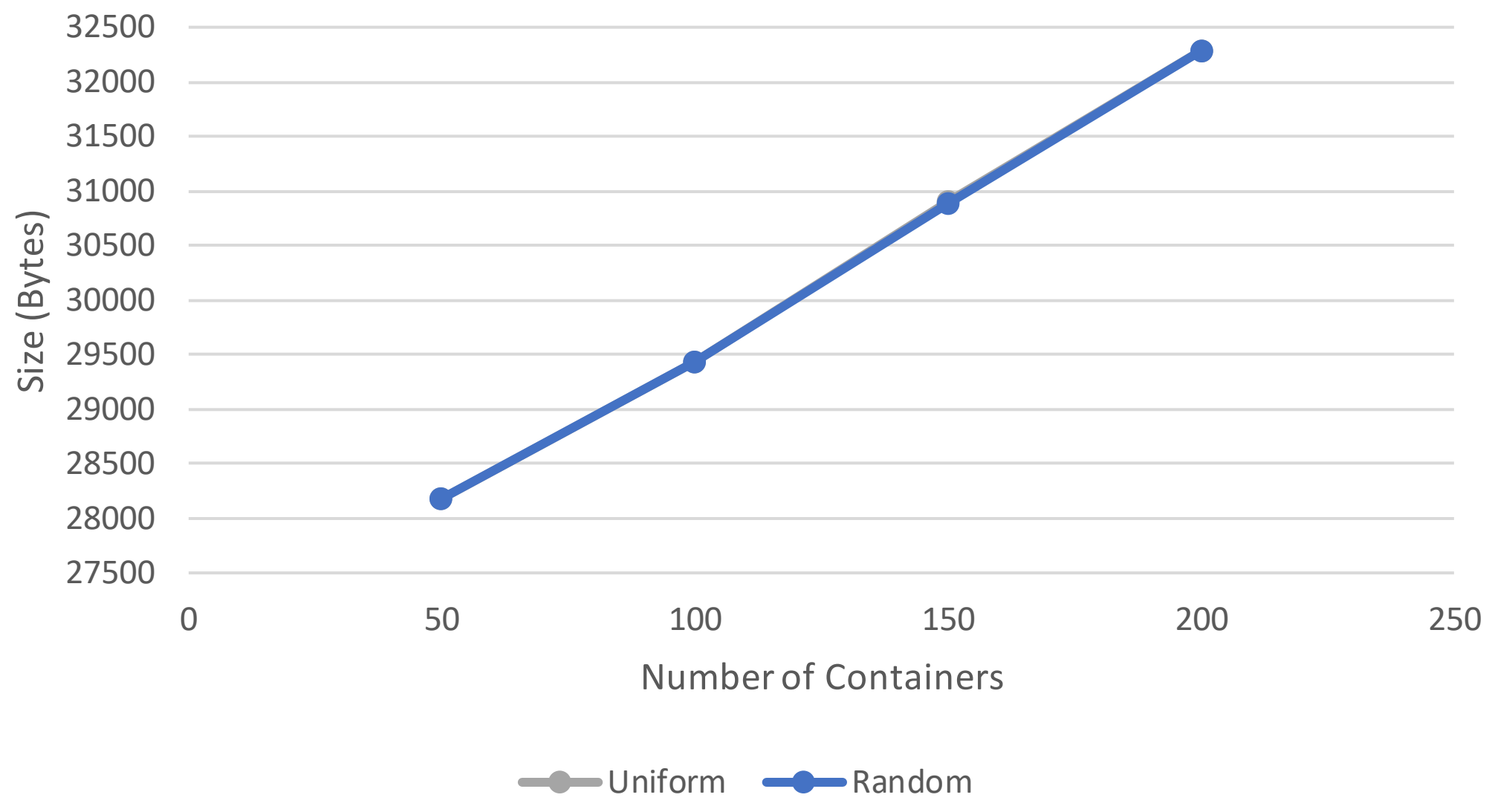

Figure 25. Container Count by Initial Network Storage Size.

Figure 26 shows the average storage size of one state in the network as compared with the number of containers. Uniform and random networks preformed roughly the same for this test. The average state size is positively correlated with the container count. This is potentially explained by more containers being declared resulting in more containers being likely to change in a state. These changes will be recorded and will result in a larger state file size.

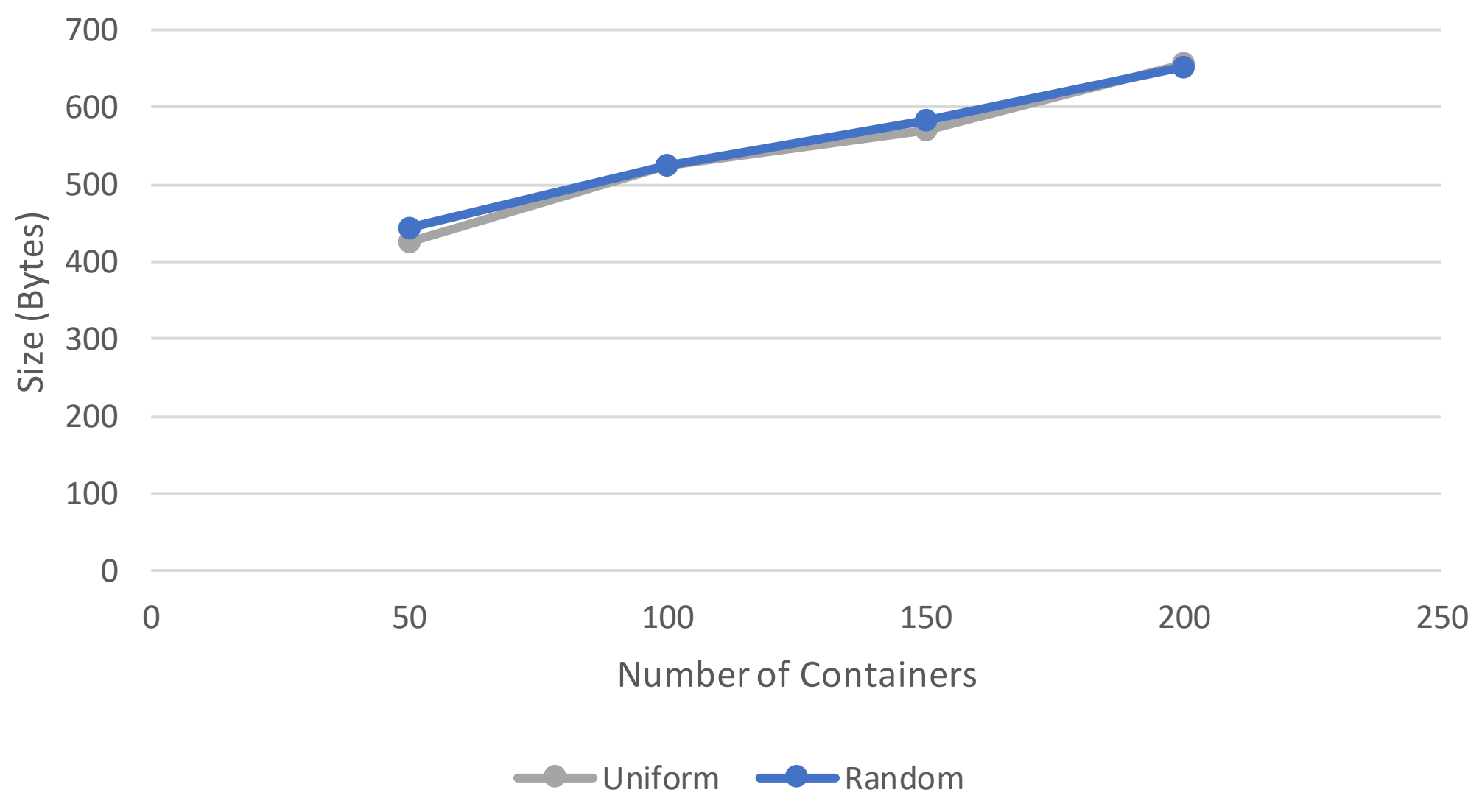

Figure 26. Container Count by Average Storage Size of One State.

Figure 27 shows the total storage size of the network (in bytes) as compared with the number of containers. Uniform and random networks preform roughly the same for this test. Storage size is positively correlated with the container count. This is explained as a summation of Figures 25 and 26. The total of the initial network size and all the states results in the total storage size. As both initial network size and average state size positively correlate with container count, the total storage size should also positively correlate with it.

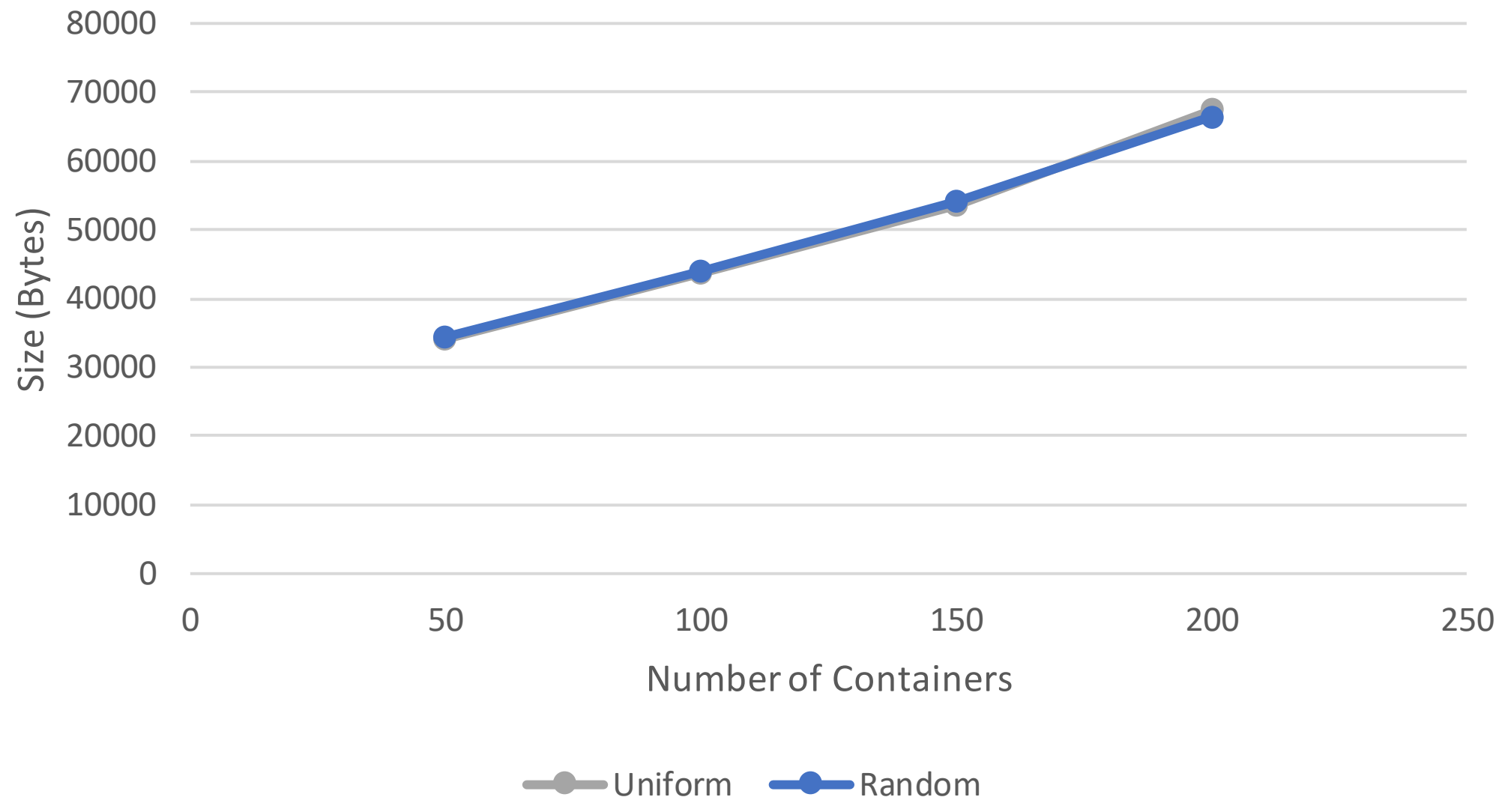

Figure 27. Container Count by Total Storage Size.

Figure 28 shows the average path length compared to the number of containers. Uniform and random networks preform roughly the same for this test. Path length and container count are nearly perfectly linearly positively correlated. This makes sense, as the shortest path length depends on the complexity of the network. The complexity depends on the number of containers in the network. In this way, when the number of containers increases, so would the average number of containers in the shortest path.

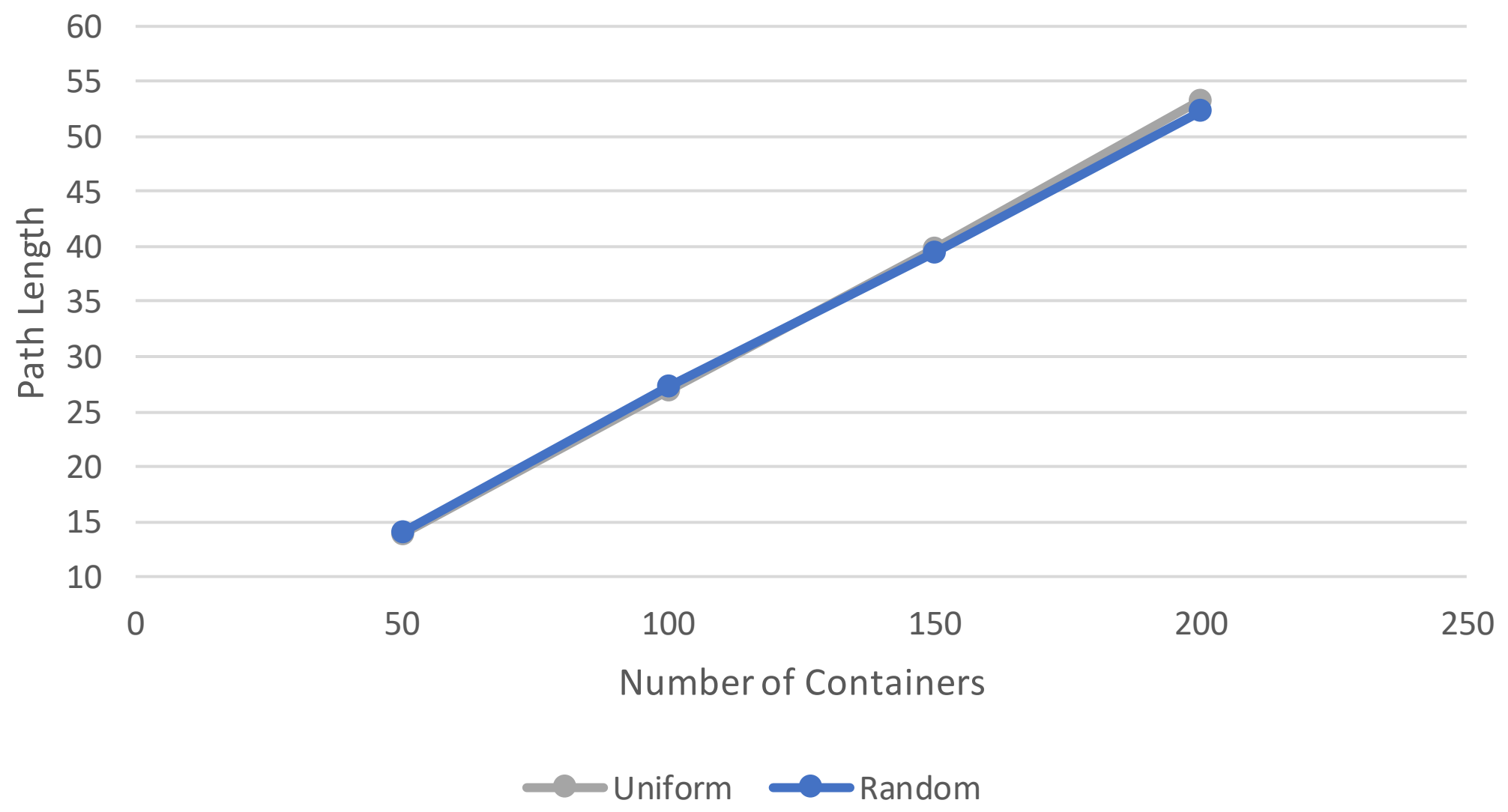

Figure 28. Container Count by Average Path Length.

### *6.5. Number of Common Properties*

The fifth area of analysis is how the number of common properties in a network affects its performance. Figure 29 shows the time required to find the shortest path of a network, as compared with the number of common properties utilized. Uniform and random networks preformed roughly the same for this test.

The shortest path time stays around 175 ticks for all common property counts, except for the count of 100. This is an outlier, where the time rises to nearly 225 ticks. This increase is mirrored to a lesser extent in Figure 35, comparing average path length. An increase in path length would explain the increase in time needed to find the shortest path.

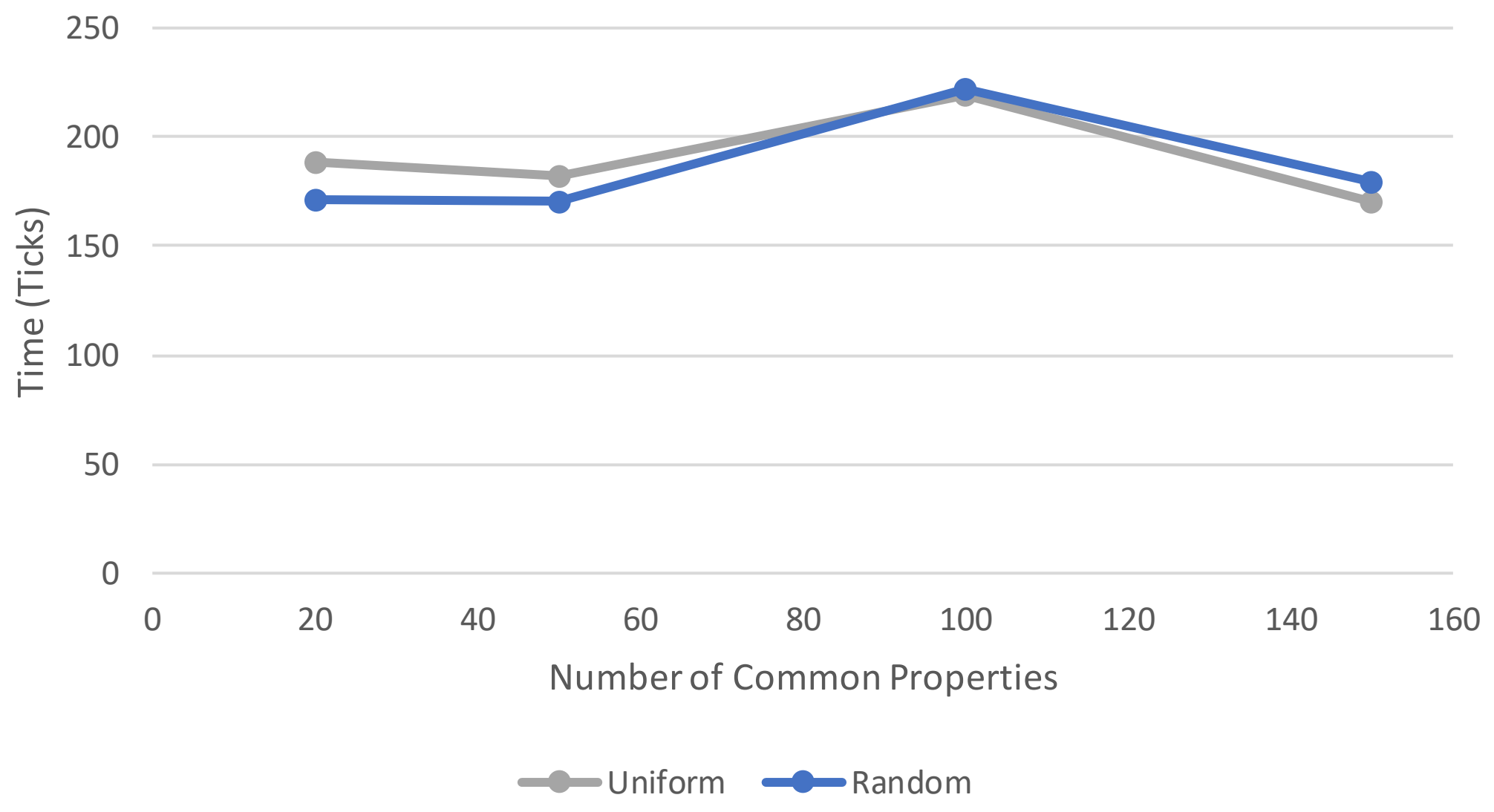

Figure 29. Number of Common Properties by Time to Find Shortest Path.

Figure 30 shows the average time to reach the next network state, as compared to the number of common properties. Uniform and random networks perform roughly the same for this test. There is a slight positive correlation between the next state time and number of common properties. This is potentially explained by the network needing to check common properties when running rules to advance to the next state. A larger number of common properties would result in a slightly longer lookup time. This could cause the small increase in run time shown here.

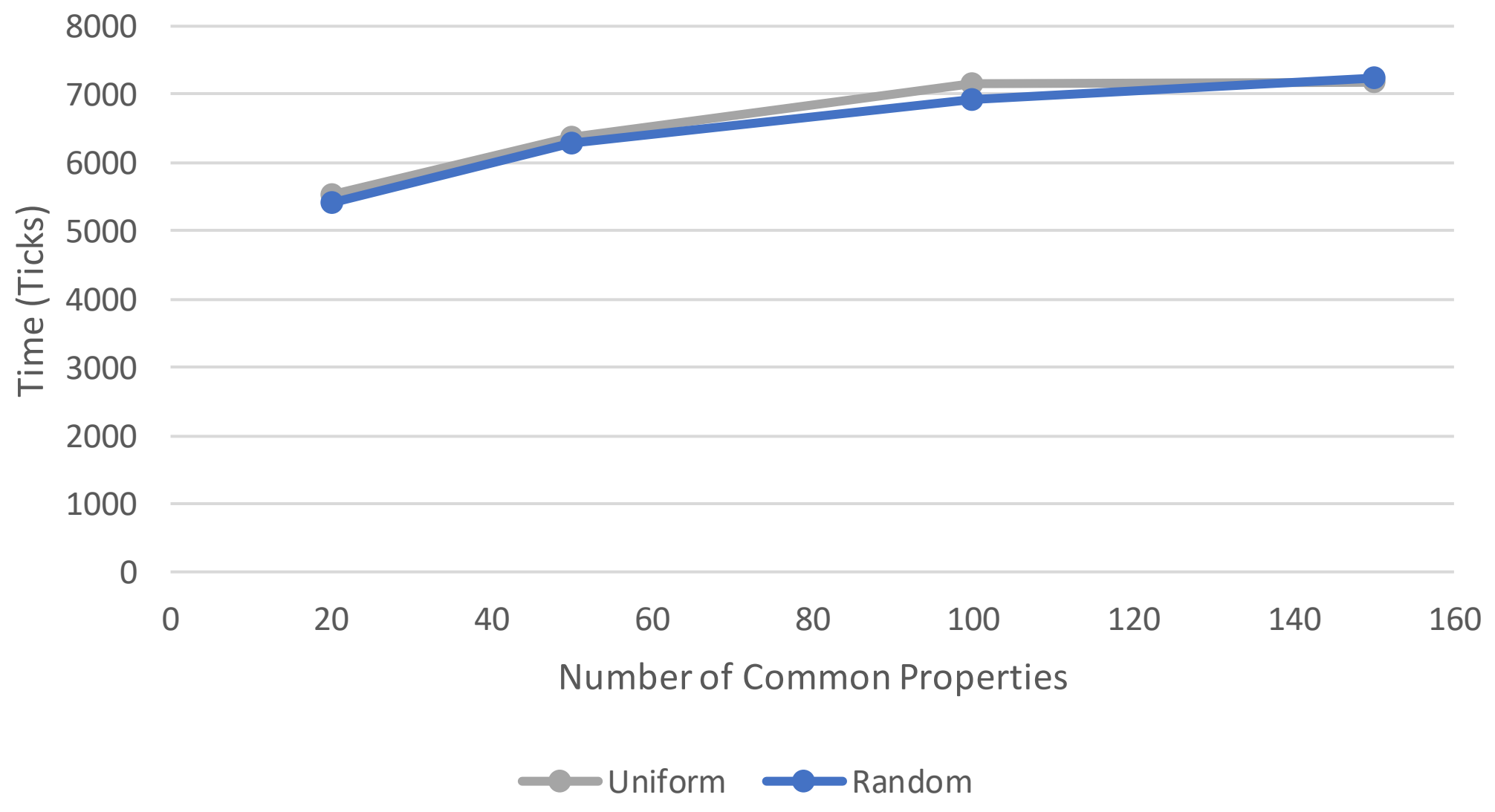

Figure 30. Number of Common Properties by Average Time to Next State.

The total traversal time of the network is compared with the number of common properties in Figure 31. Uniform and random networks perform roughly the same for this test. Much like in Figure 30, there is a slight positive correlation between the total traversal time and the number of common properties. This could be explained by the same cause of more common properties resulting in a longer lookup time and, thus, greater overall time being required.

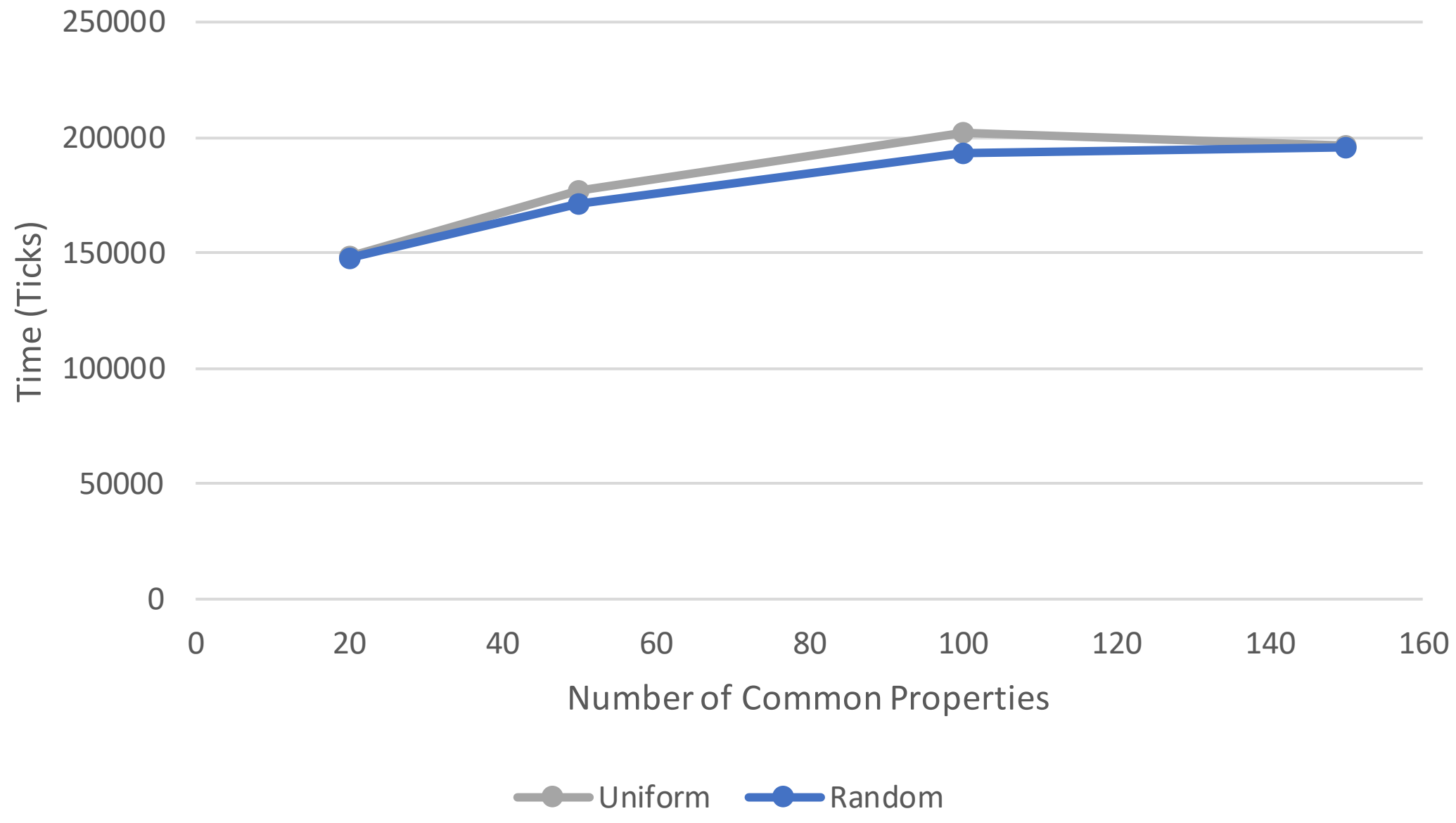

Figure 31. Number of Common Properties by Total Traversal Time.

Figure 32 shows the initial storage size of the network compared with the number of common properties. Uniform and random networks preform roughly the same for this test. There is a positive correlation between the storage size and the common property count. This is explained by the storage space needed to store common properties in the initial state. More common properties will result in more space being needed for the initial state.

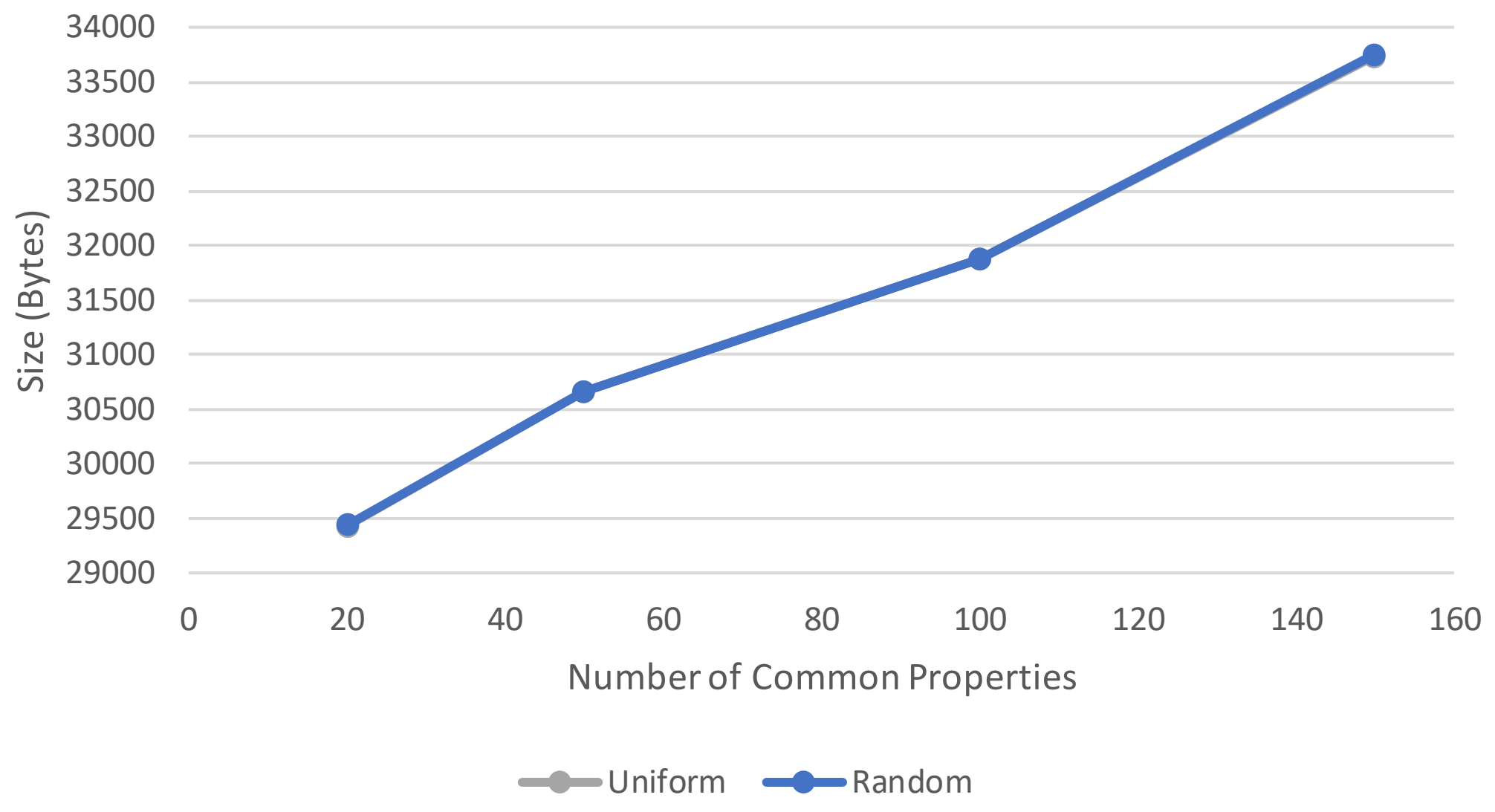

Figure 32. Number of Common Properties by Initial Network Storage Size.

Figure 33 shows the average storage size of a network state as compared with the number of common properties. Uniform and random networks perform roughly the same at common property counts of 20 and 50; however, uniform networks have a slight increase in size, over random networks, at common property counts of 100 and 150. The average network state size is positively correlated with the number of common properties. This is potentially explained by an increase in the number of common properties resulting in containers having more common properties, on average. When these common properties change, it is recorded to the state file. Therefore, an increase to the common property count results in an increase in the average storage size of a state.

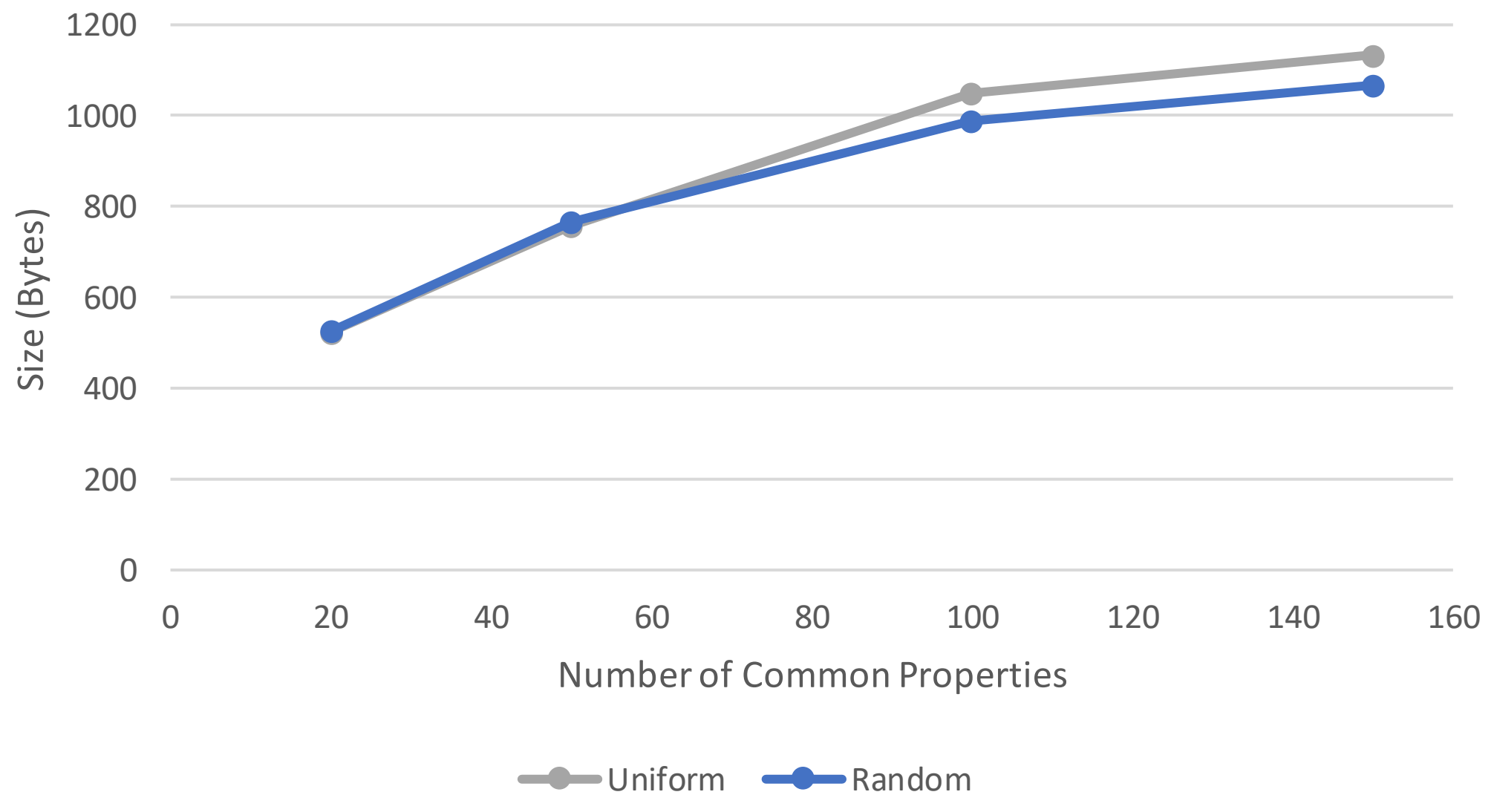

Figure 33. Number of Common Properties by Average Storage Size of One State.

The total storage size of a network (in bytes) is compared with the number of common properties in Figure 34. Much like in Figure 33, uniform and random networks preform roughly the same at common property counts of 20 and 50; however, uniform networks have a slight increase in size, over random networks, at common property counts of 100 and 150. Otherwise, the total storage size is positively correlated with the number of common properties. This makes sense, as the total storage size is a summation of the initial network size and the size of all of the state files. As both Figures 32 and 33 show a positively correlation, Figure 34 should be positively correlated as well. This is further shown in the slight increase in size that uniform networks have, over Random Networks, at the higher common property levels, just like is shown in Figure 33.

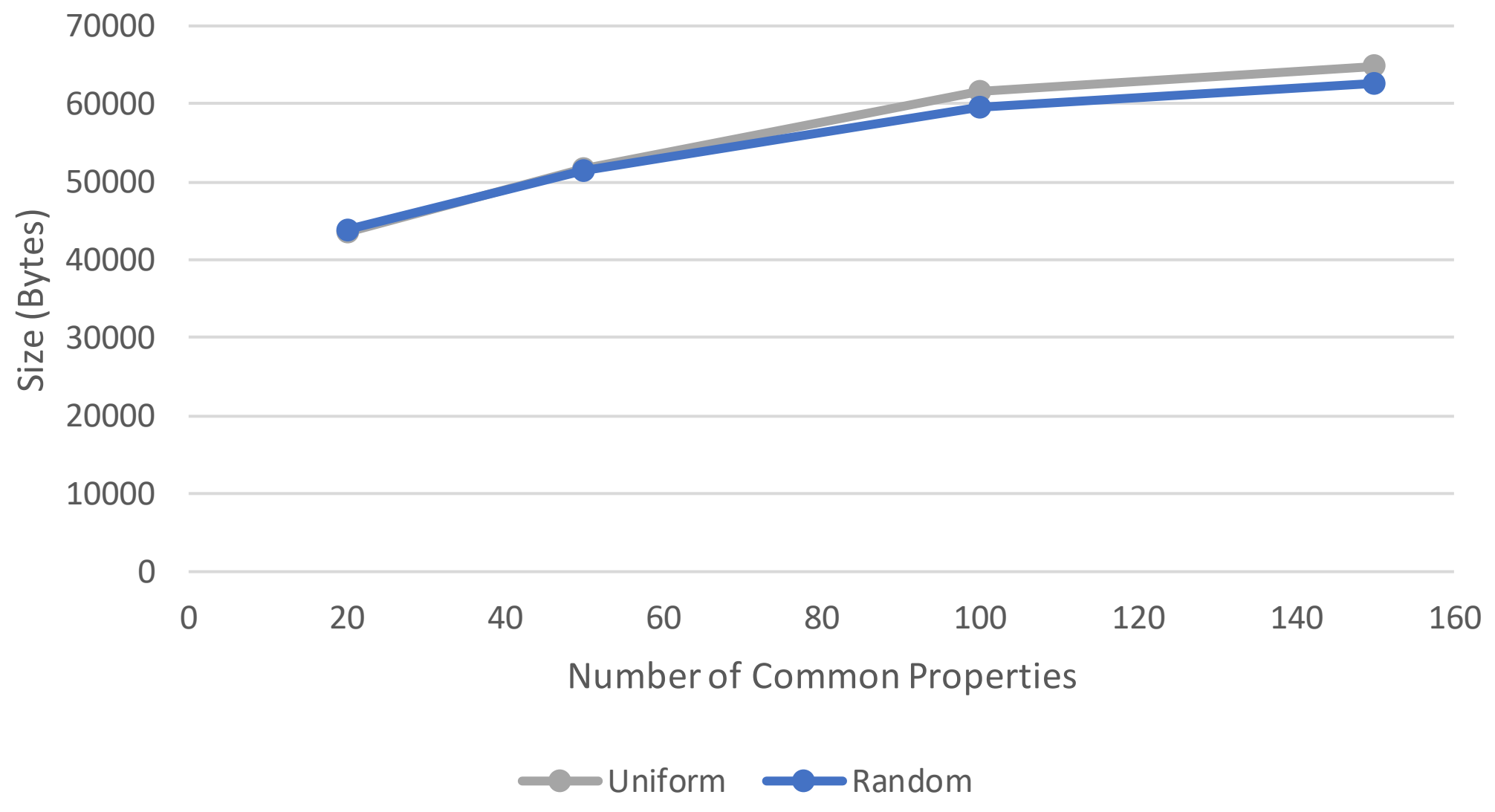

Figure 34. Number of Common Properties by Total Storage Size.

Figure 35 shows the average path length, as compared with the number of common properties. Uniform and random networks preform roughly the same, with a slight increase in path length in uniform networks over random networks. An exception to this is at the common property count of 20. This is the only point where random networks have a higher path length as compared to uniform networks. Otherwise, the average path length stays roughly the same, except at a common property count of 100, where it increases slightly for both network types.

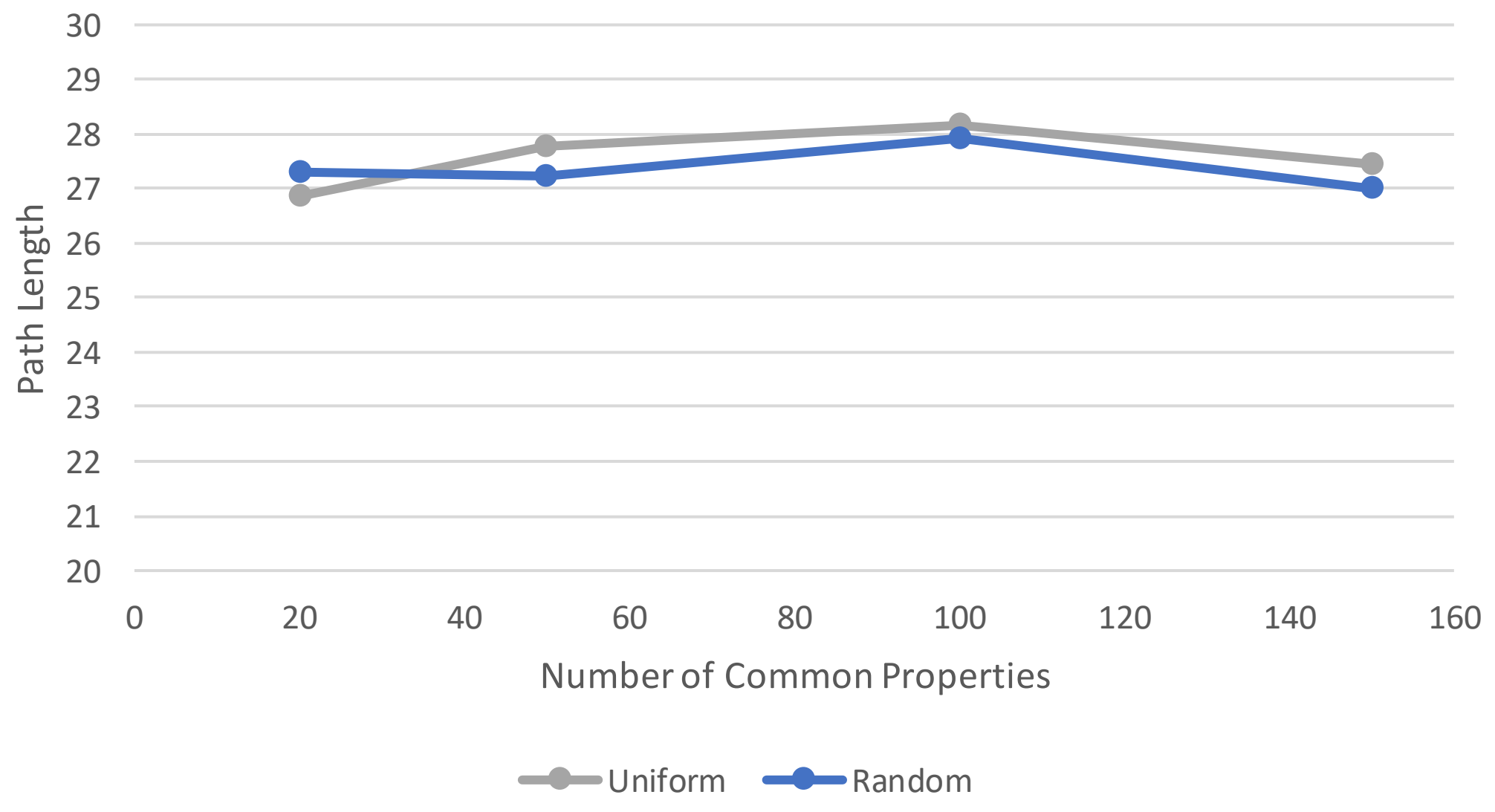

Figure 35. Number of Common Properties by Average Path Length.

***6.6. Number of Common Properties Per Rule***

The sixth area of analysis surrounds how the number of common properties per rule in the network affects its performance. Figure 36 shows the time required to find the shortest path in a network (in ticks), as compared with the number of common properties per rule. Random networks have a higher time requirement at common property counts of 1 and 2, but then fall below uniform networks at 5, and become roughly the same at 10. Random networks start higher, at around 250 ticks, and then drop down to around 175, at the common property count of 5. Uniform networks behave the same way; however, they start closer to 210 ticks. Overall, the time to find the shortest path decreases as the number of common properties per rule increases and plateaus after a common property count of 5. This seems contrary to the common logic that more common properties to check would result in a longer time rather than a shorter. More analysis is required to learn why this contradiction is observed.

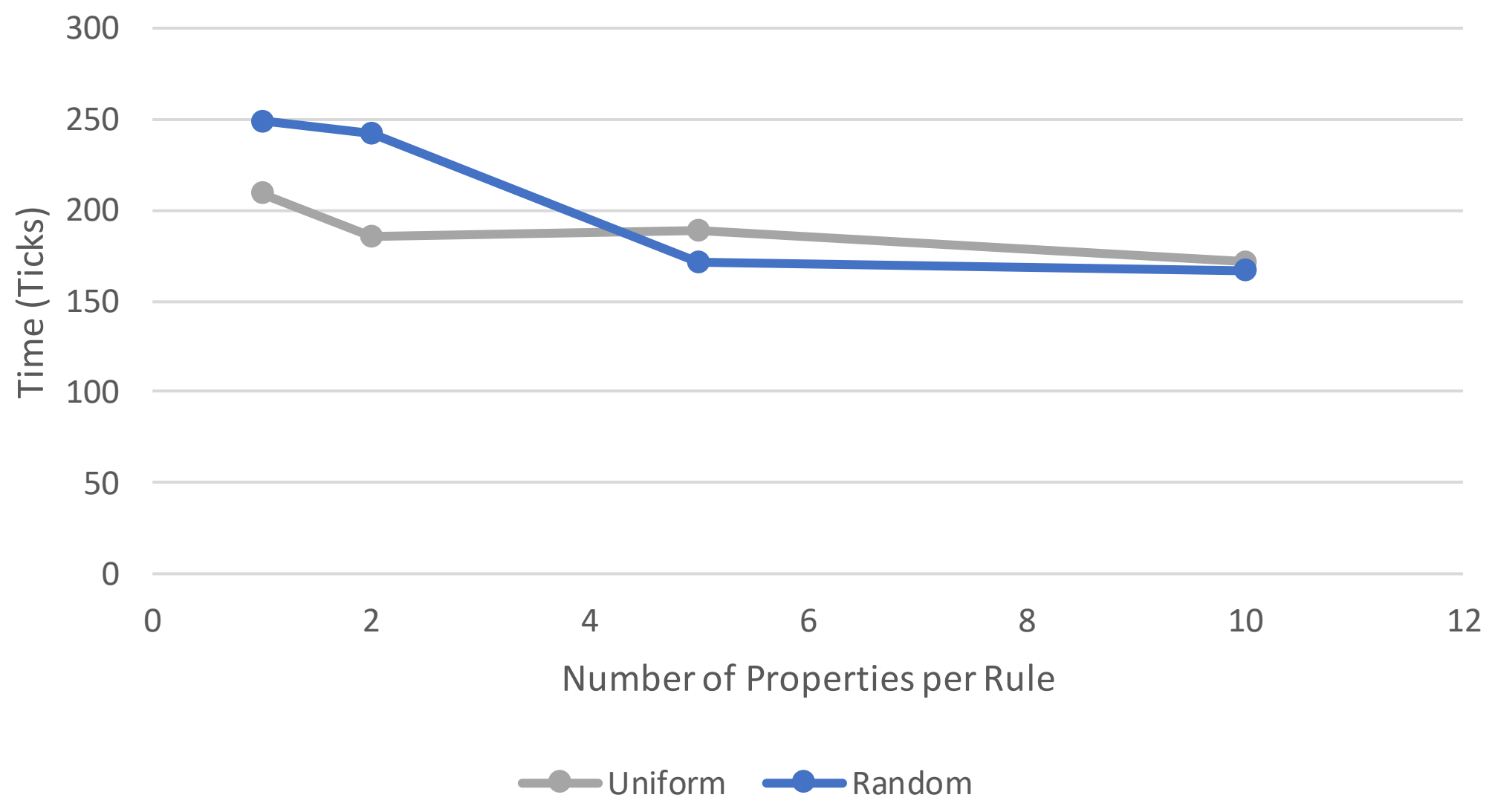

Figure 36. Number of Common Properties Per Rule by Time to Find Shortest Path.

Figure 37 shows the average time (in ticks) required to reach the next network state, as compared to the number of common properties per rule. Uniform and random networks switch at each datapoint as to which takes longer than the other. It is important to note that both have a large increase in time required from a common property count of 1 to 2 but then uniform networks' time requirements decrease as the count increases. Random networks' time requirements, in contrast, sharply decrease at a count of 5 and then climb back up at a count of 10. It would be expected that, as the number of common properties per rule increases, rules that are able to run would take a longer time. This is explanation does not match Figure 37, so more analysis is required to learn why this pattern occurred.

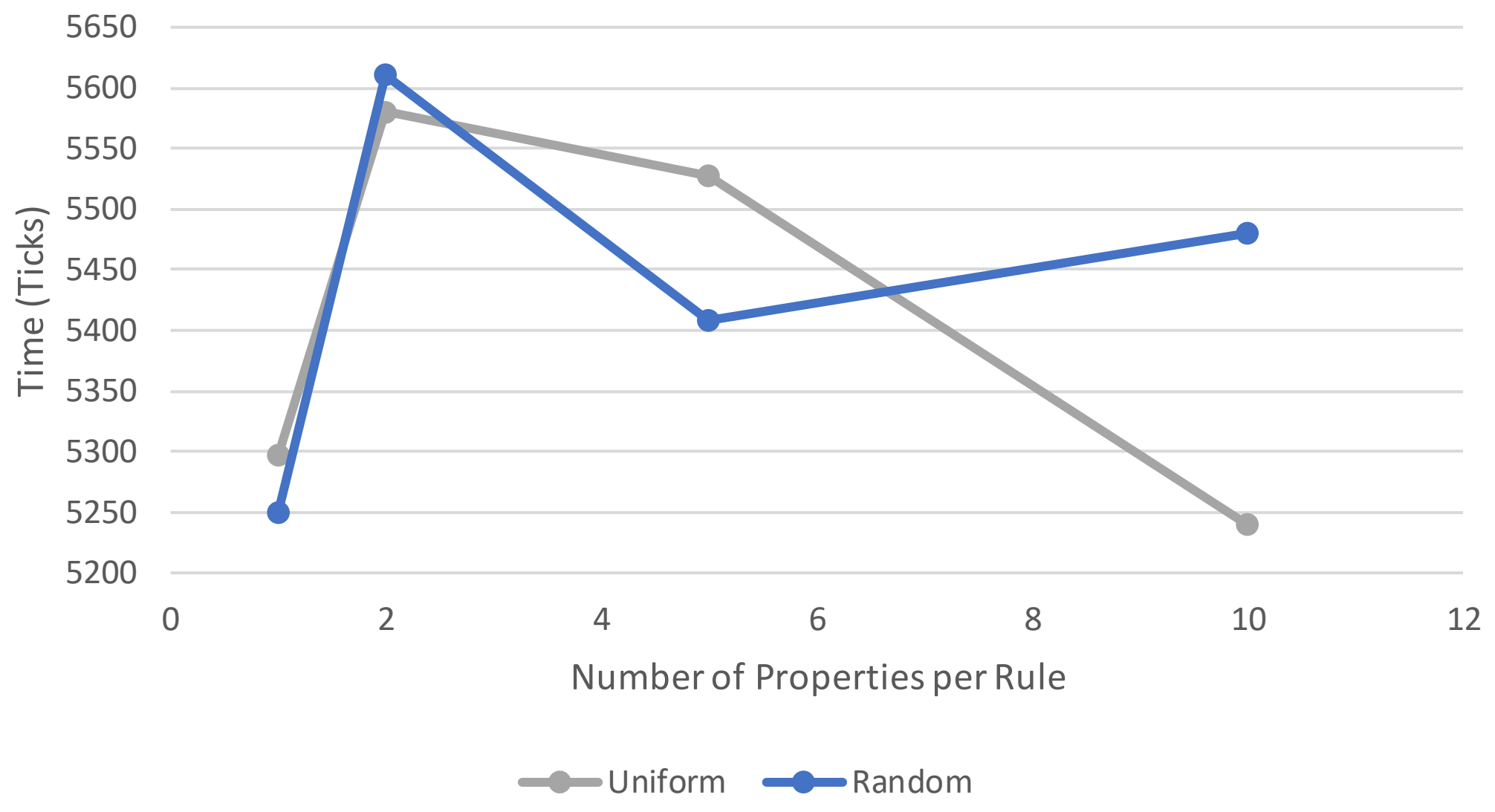

Figure 37. Number of Common Properties Per Rule by Average Time to Next State.

Figure 38 shows the total traversal time (in ticks) of networks, as compared with the number of common properties per rule. Uniform and random networks switch, at each datapoint, as to which takes longer than the other. The pattern slightly mirrors the pattern from Figure 37. Both types of networks start with a large increase and then uniform networks decrease and random networks have a sharp decrease and then final climb. It makes sense that total traversal time would have elements of the graph of average time to next state as total traversal time is a summation of all state changes.

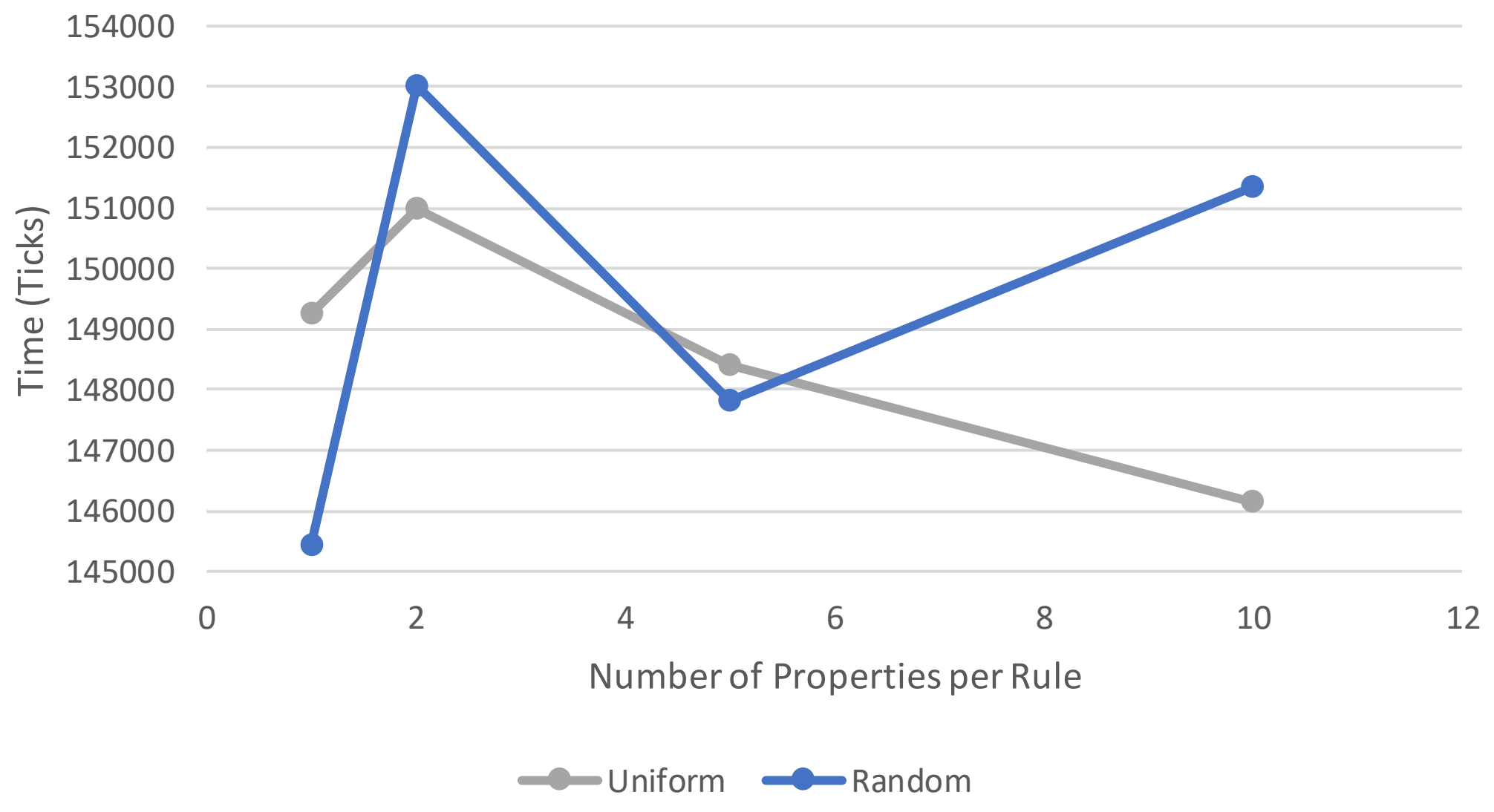

Figure 38. Number of Common Properties Per Rule by Total Traversal Time.

Figure 39 shows the initial network storage size (in bytes), as compared with the number of common properties per rule. Uniform and random networks perform roughly the same for this test. The initial storage size is positively correlated with the common property per rule count. This makes sense, as when rules are declared, their common properties are identified. More common properties per rule would make each rule size larger, resulting in a larger initial network size as the common property count increases.

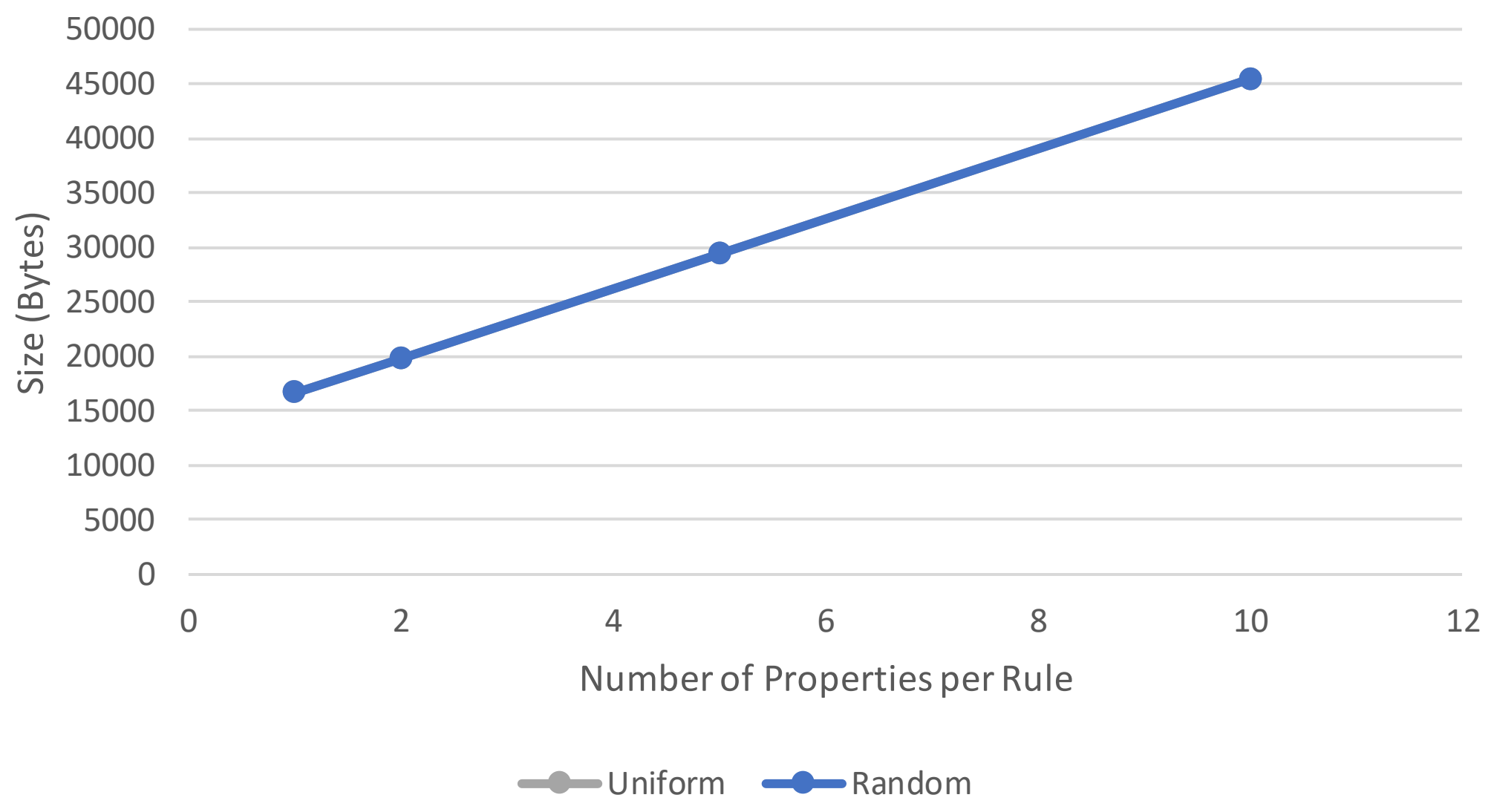

Figure 39. Number of Common Properties Per Rule by Initial Network Storage Size.

Figure 40 shows the average storage size of one network state (in bytes), as compared with the number of common properties per rule. Uniform and random networks perform roughly the same for this test. At a common property per rule count of 1 to 2, there is an increase. Then the size decreases with the property count. This could be explained by more common properties leading to more rules being selected that do not have strict requirements. Without these strict requirements, less facts are changed, leading to a smaller state size as only changes are recorded.

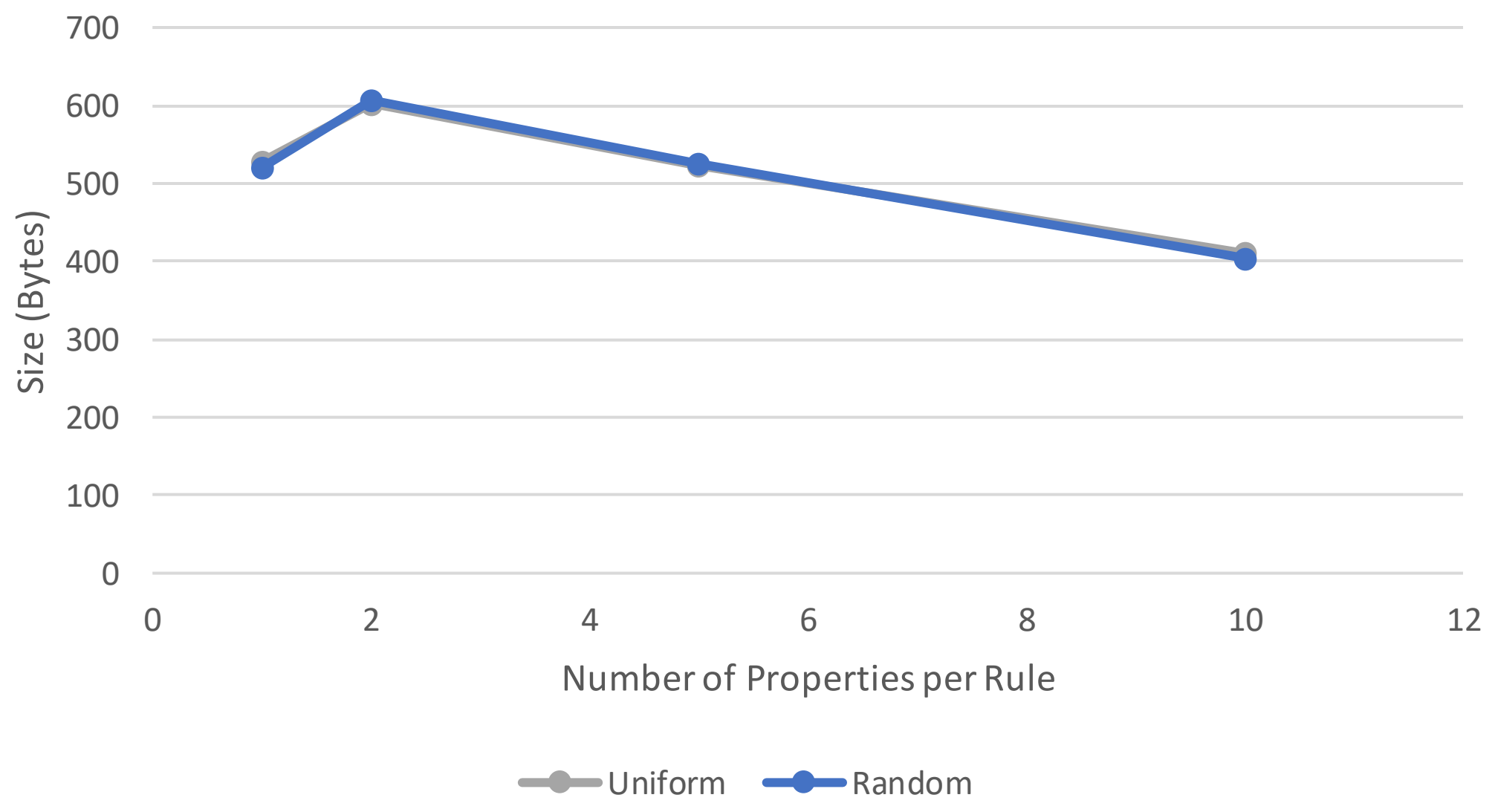

Figure 40. Number of Common Properties Per Rule by Average Storage Size of One State.

The total storage size of a network (in bytes) is compared with the number of common properties per rule in Figure 41. Uniform and random networks preform roughly the same for this test. Total network storage size and the number of common properties per rule are positively correlated. This makes sense as a large amount of the storage space is taken up by storing the initial network state, as shown in Figure 39. More common properties per rule results in larger rule declarations which in turn results in a larger required storage size.

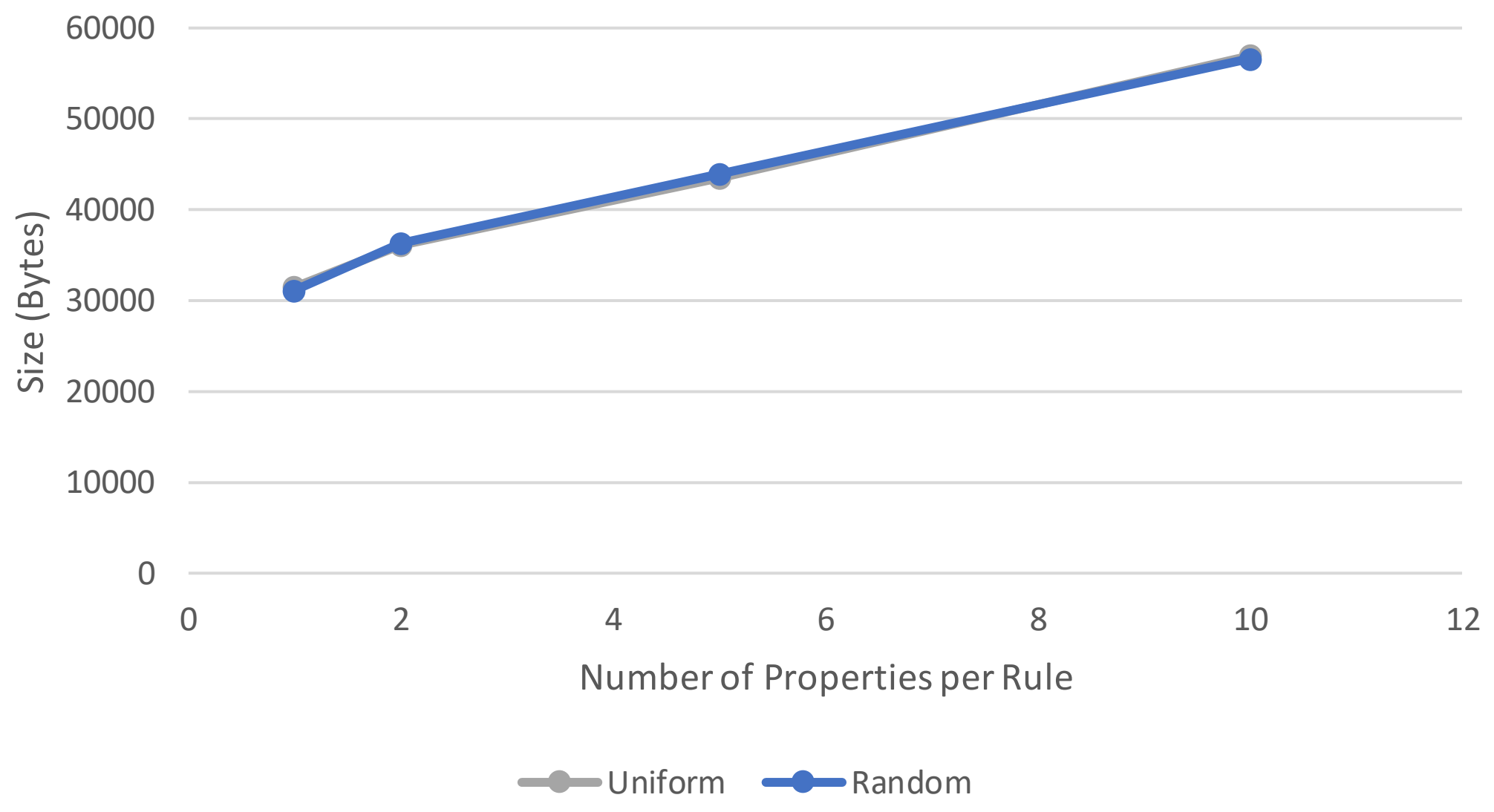

Figure 41. Number of Common Properties Per Rule by Total Storage Size.

Figure 42 shows the average path length of the network tests, as compared with the number of common properties per rule. Uniform and random networks performed roughly the same for this test, albeit with random networks changing less over time. There is a slight decrease in path length, from common property counts of 1 to 5. Then there is a slight increase, at a count level of 10.

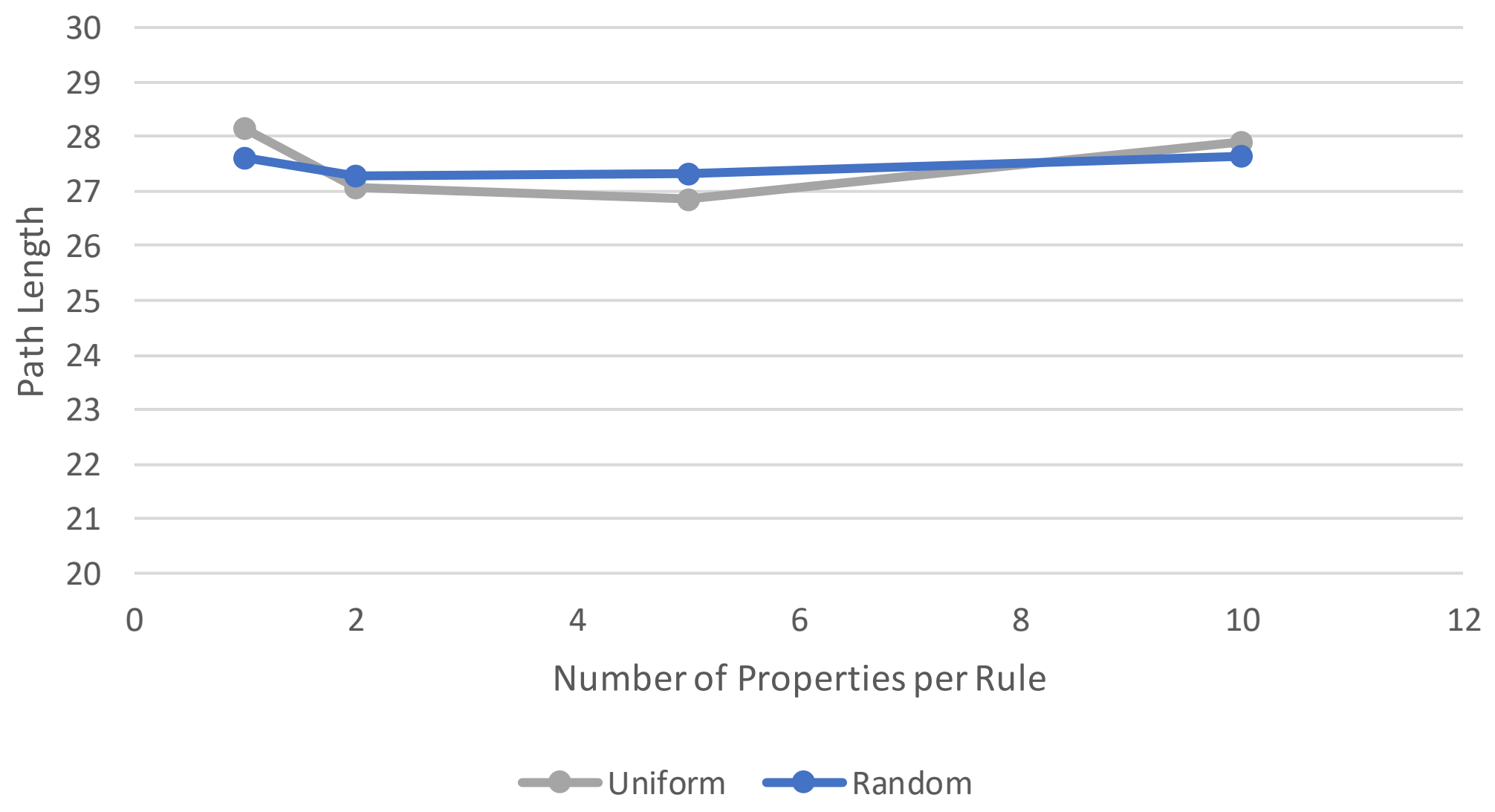

Figure 42. Number of Common Properties Per Rule by Average Path Length.

***6.7. Hybrid Rule Chance***

The seventh area of analysis is how the hybrid rule chance setting for the network effects its performance. Figure 43 shows the time to find the shortest path of a network (in ticks), as compared with the chance of a hybrid rule. Uniform and random networks perform roughly the same, except at the lowest chance tested, 25%. At a hybrid rule chance setting of 25%, random networks take about 50 ticks longer to find the shortest path.

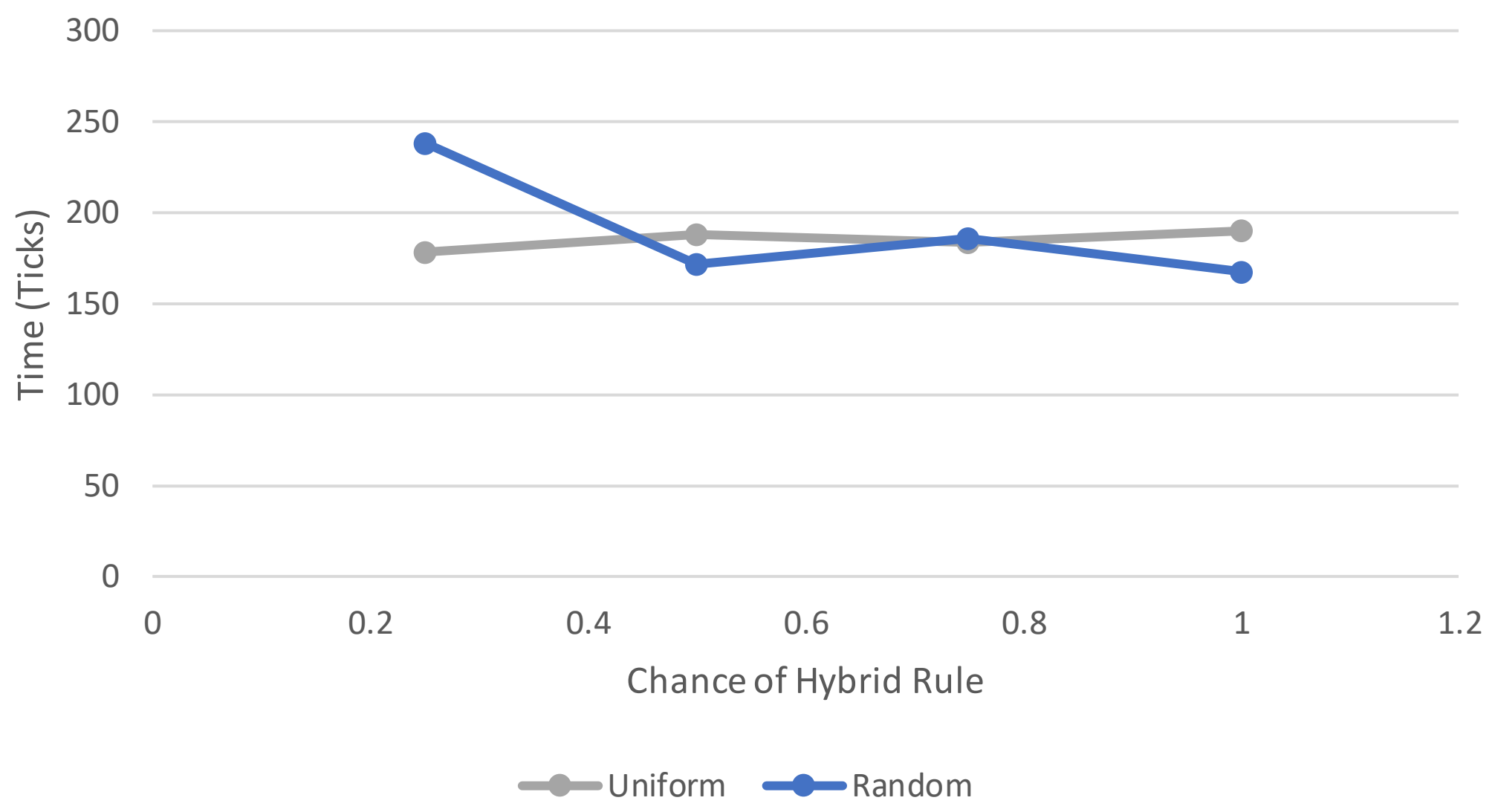

Figure 43. Hybrid Rule Chance by Time to Find Shortest Path.

Figure 44 shows the average time required to reach the next network state (in ticks), as compared with chance of a hybrid rule setting values. Random networks start at around 5550 ticks and then dip down to 5400 ticks, before returning to above 5600 ticks. Uniform networks do the opposite of this, starting around 5400 ticks, increasing to about 5525 ticks, then decreasing back down to around 5450 ticks. More testing would be required to determine why the two types of networks act in inverse patterns; however, the level of changes (as compared to the total time cost) is limited.

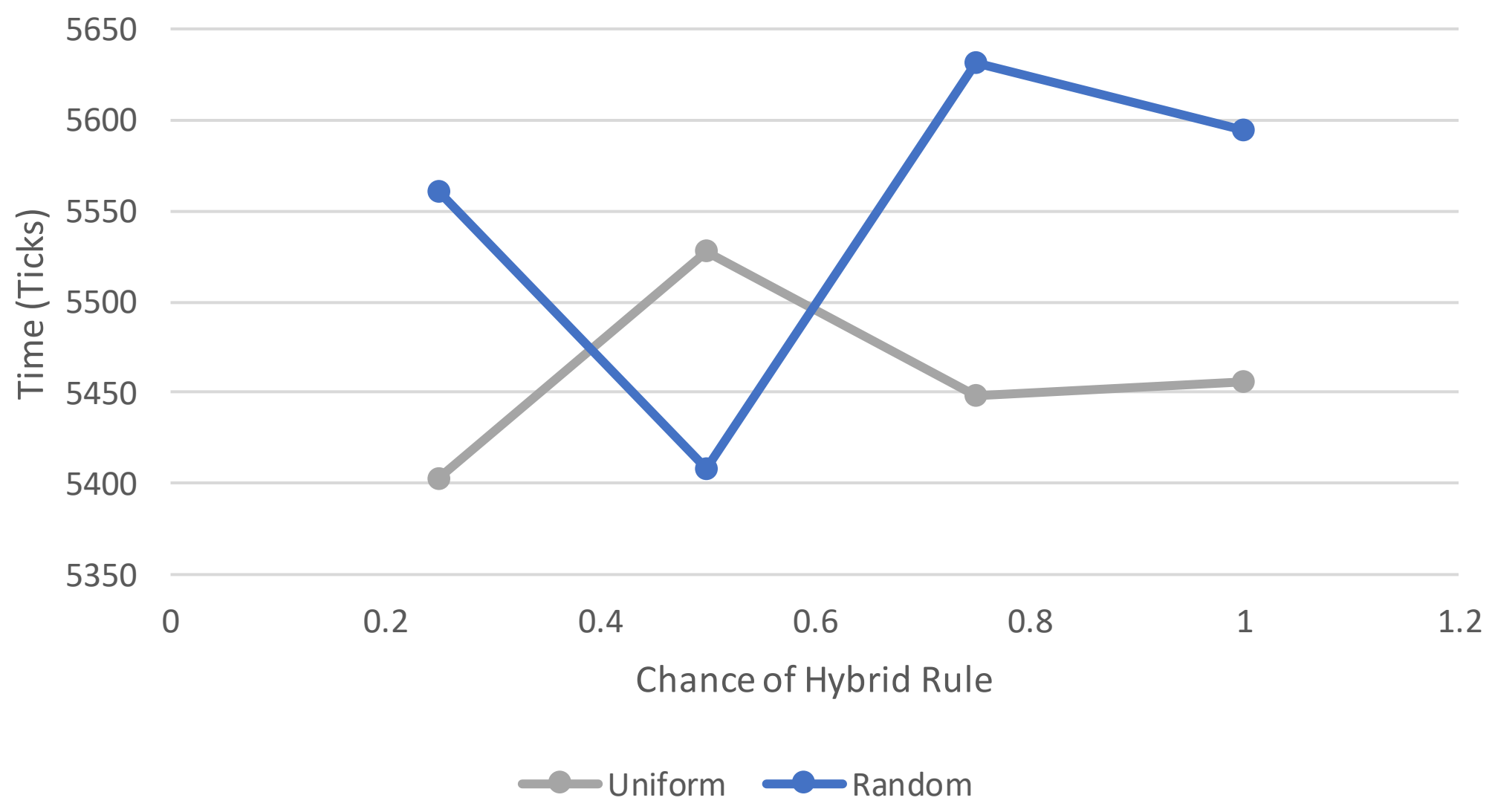

Figure 44. Hybrid Rule Chance by Average Time to Next State.

Figure 45 shows the total traversal time of a network (in ticks), as compared with the chance of a hybrid rule setting. Random networks mirror the pattern found in Figure 44. Uniform networks stay roughly the same, with an increased datapoint outlier at 75%. Mirroring the pattern shown in Figure 44 makes sense, as the total traversal time is just a summation of the time taken to reach each network state. It is unclear as to why uniform networks do not follow their respective pattern, shown in Figure 44, when random networks do.

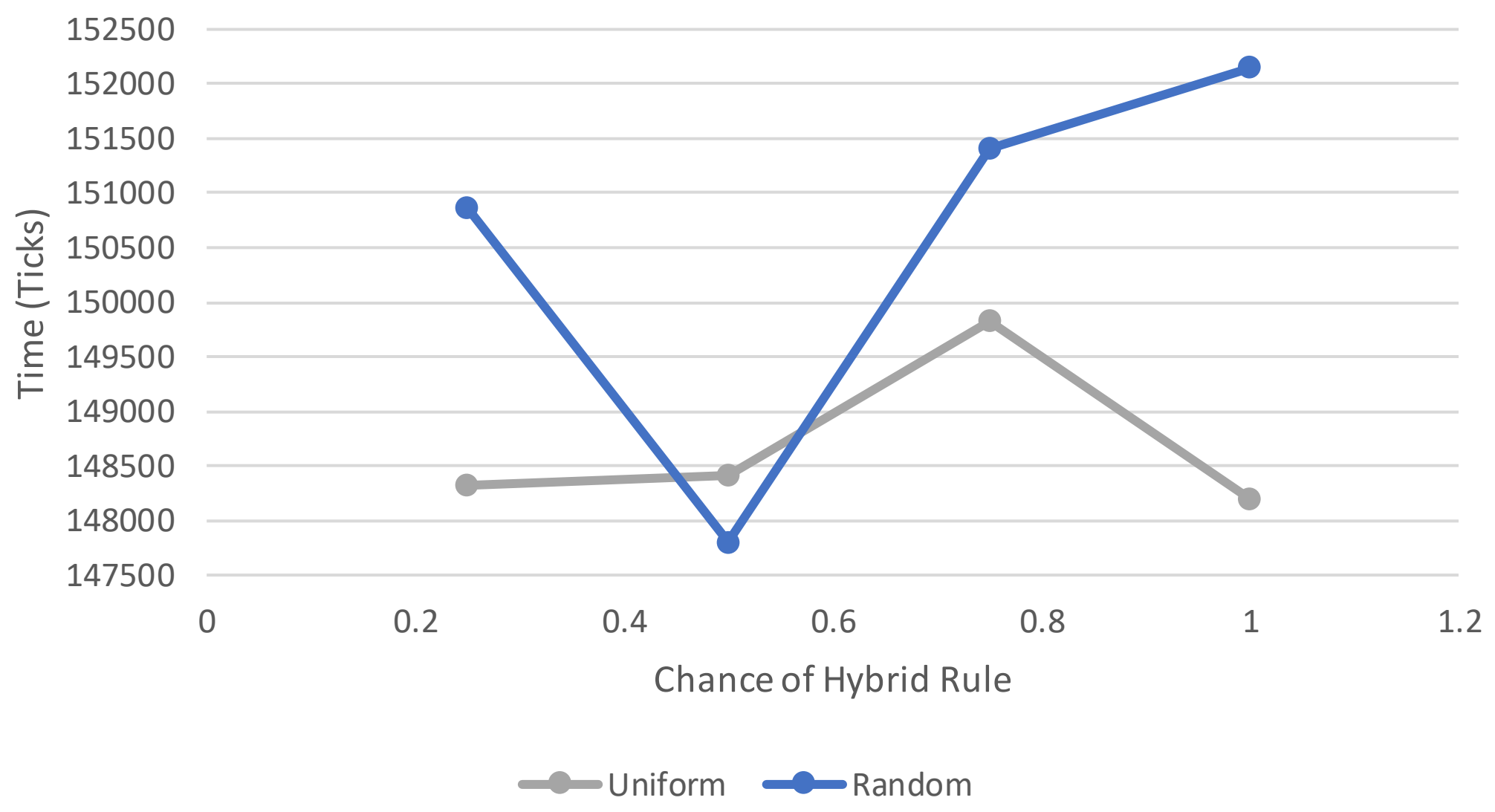

Figure 45. Hybrid Rule Chance by Total Traversal Time.

Figure 46 shows the initial network storage size (in bytes), as compared with the chance of a hybrid rule setting. The changing of one rule to be a hybrid rule should not affect the size of the rules in the initial network size. Even though this graph seems to show wild variation, the variation is only by a max of about 15 bytes or 120 bits. Considering the state data is stored in UTF-8 format, this is a change of at most of 15 characters. This could easily be accounted for by property name differences. Overall, the hybrid rule change setting has no notable impact on the initial network state size.

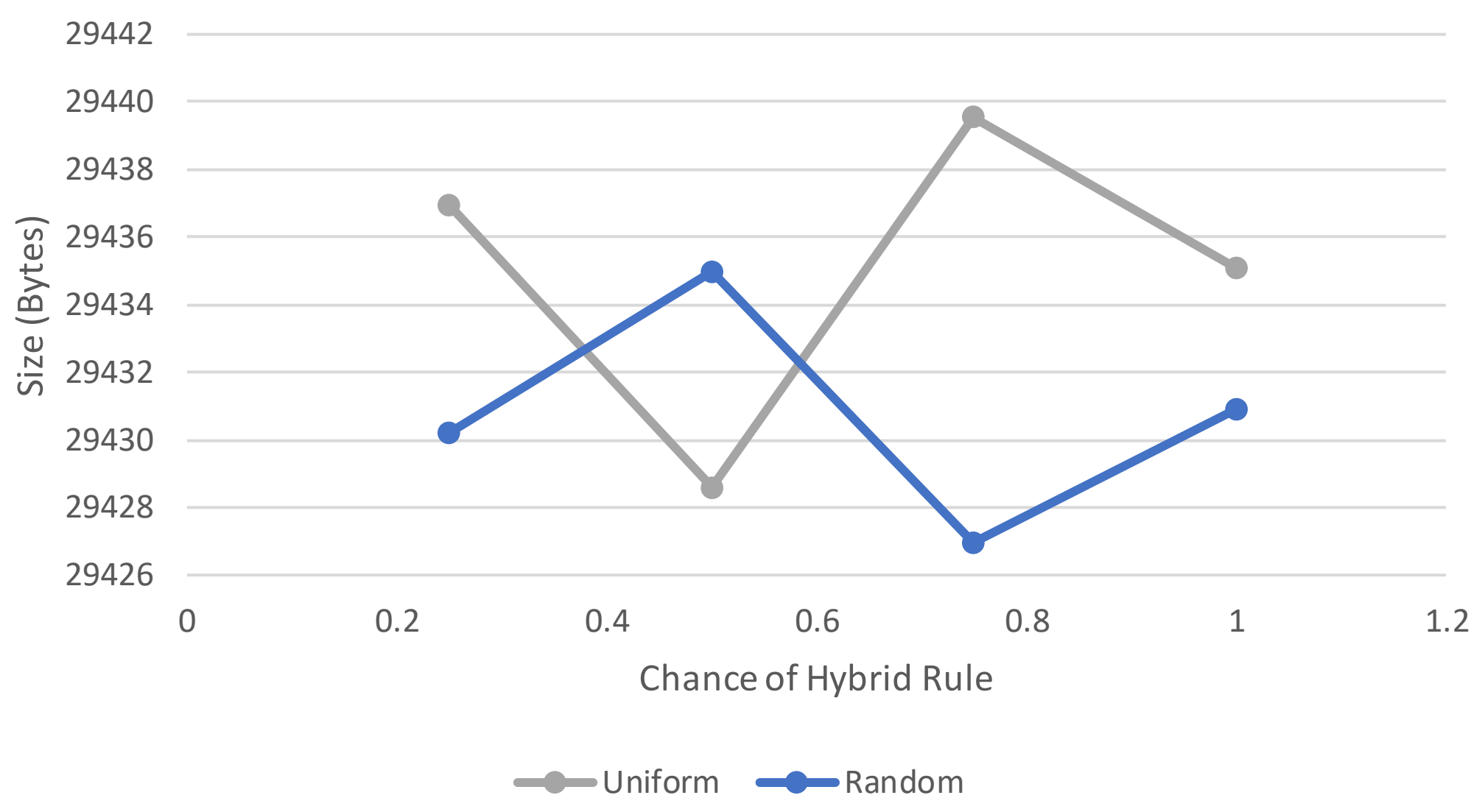

Figure 46. Hybrid Rule Chance by Initial Network Storage Size.

Figure 47 shows the average storage size of one state (in bytes), as compared with the hybrid rule chance setting value. Much like in Figure 46, there is a maximum of about 15 bytes of difference. Because the data is encoded as UTF-8, this 15 characters worth of difference can be accounted for by property name differences. Overall, again, the hybrid rule change setting has no notable effect on the network state size.

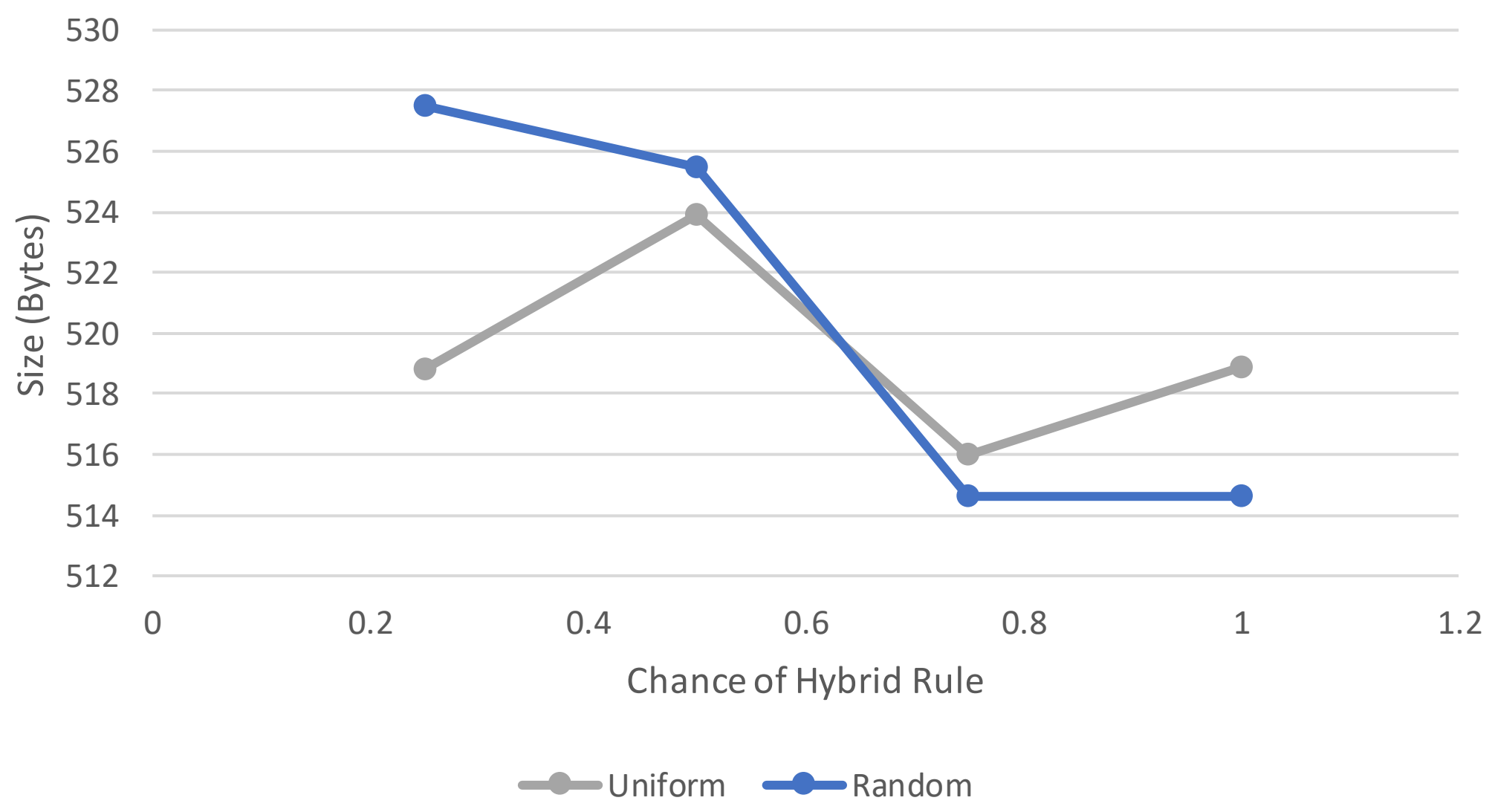

Figure 47. Hybrid Rule Chance by Average Storage Size of One State.

The total network storage size (in bytes) is compared with the hybrid rule chance setting in Figure 48. Because, in regards to Figures 46 and 47, it is explained that, overall, there is no notable size difference and the total storage size is a summation of these two figures, total storage size should also not be affected by the hybrid rule chance setting value. Similarly, minimal variation is also shown in Figure 48.

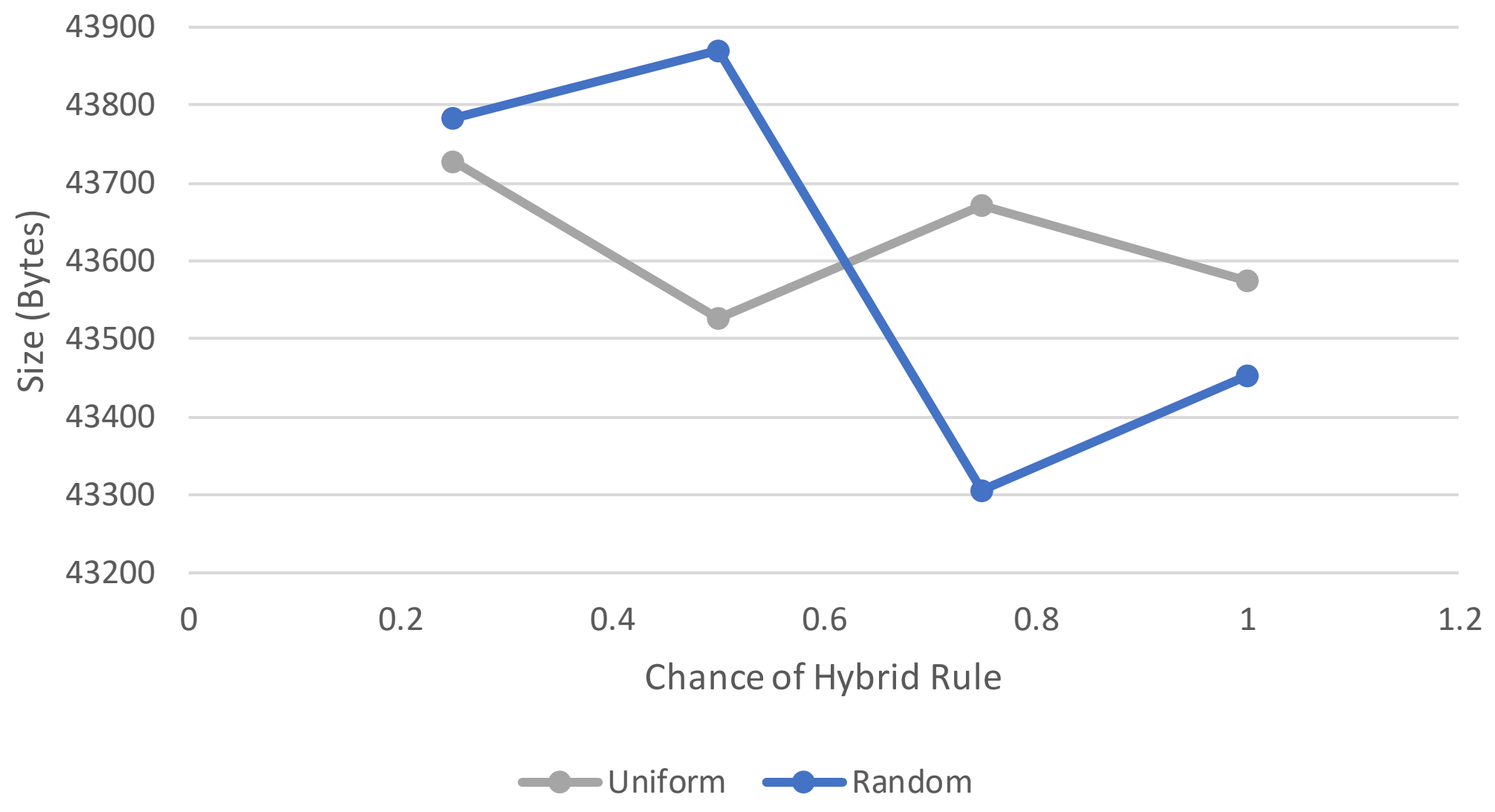

Figure 48. Hybrid Rule Chance by Total Storage Size.

Figure 49 shows the average path length of networks, as compared with the hybrid rule chance setting. Uniform and random networks performed roughly the same for this test and there is very little change as the hybrid rule chance setting changes.

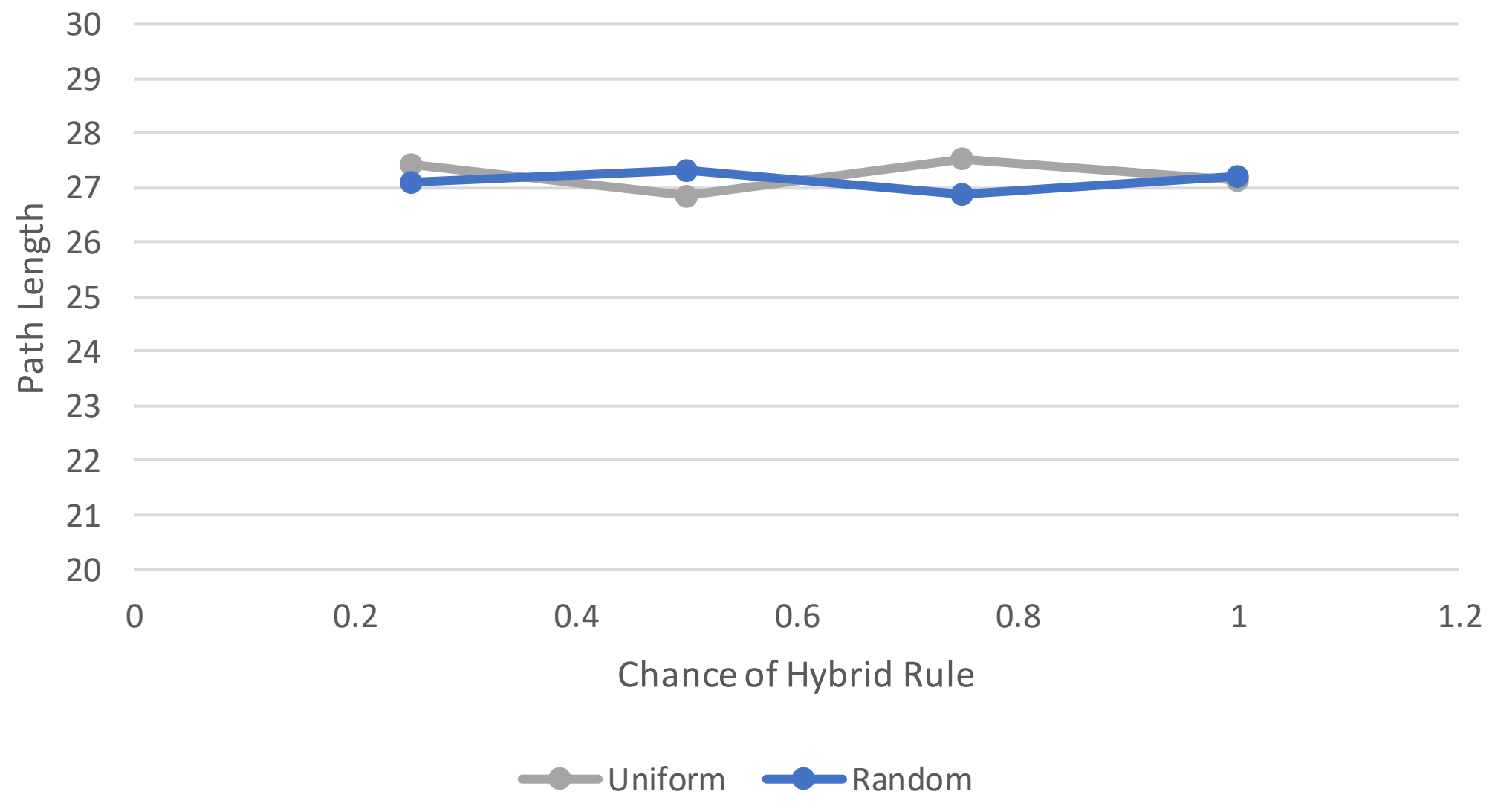

Figure 49. Hybrid Rule Chance by Average Path Length.

### *6.8. Rule Ignore Property Chance*

The eighth area of analysis is how the rule ignore property chance in the network affects its performance. Figure 50 shows that the time required to find the shortest path (in ticks), as compared with rule ignore property chance setting values. Random networks stay approximately constant, around 175 ticks. Uniform networks start just above random networks and then increase to above 200 ticks at 50% and then decrease again. While it is interesting that the rule ignore property chance has no discernable effect on random networks, more analysis is required to learn why a change in uniform networks' performance is observed.

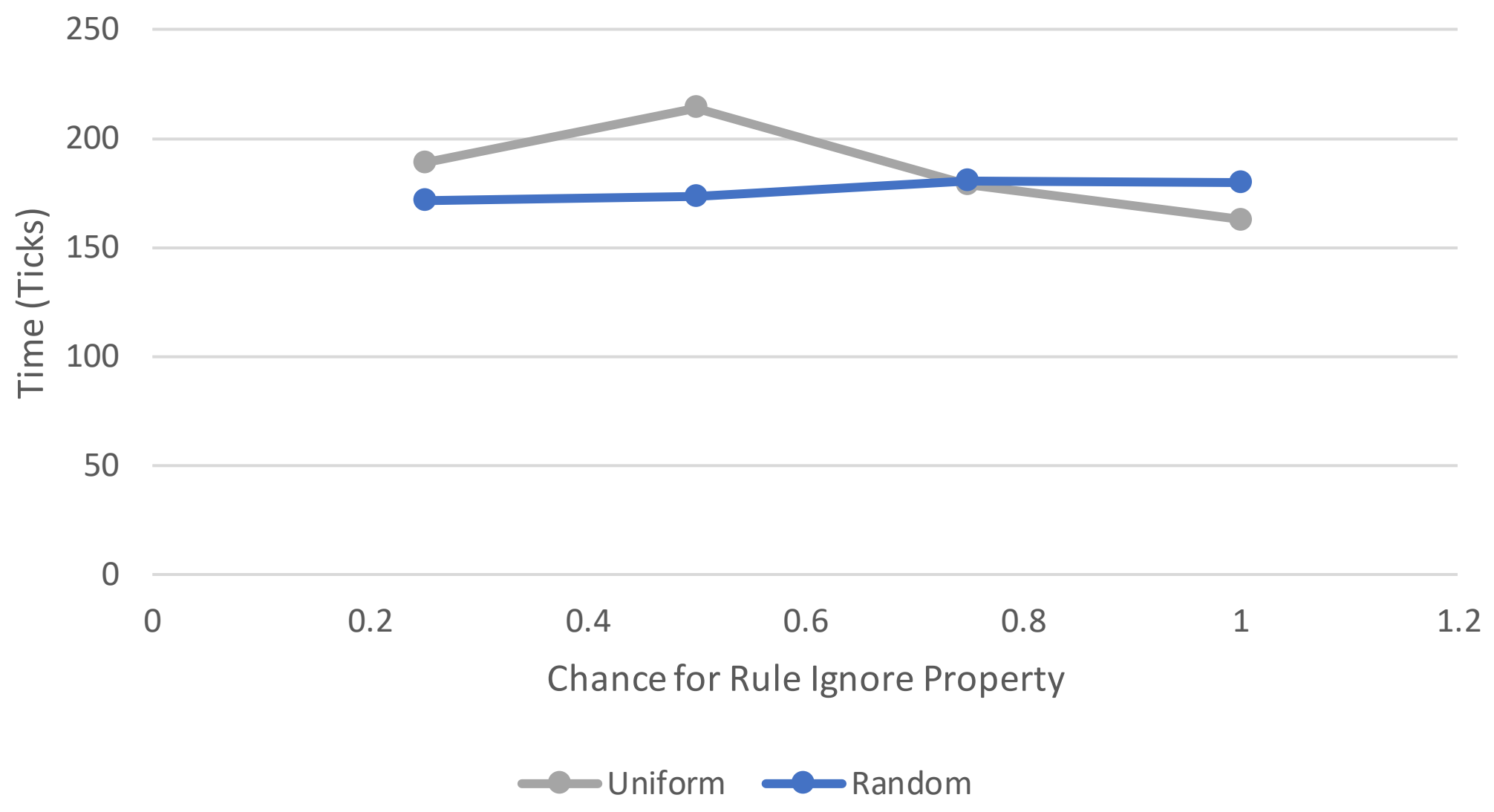

Figure 50. Rule Ignore Property Chance by Time to Find Shortest Path.

Figure 51 shows the average time required to reach the next network state (in ticks), as compared with rule ignore property chance setting values. Uniform and random networks performed roughly the same for this test. There is a slight positive correlation between network state time and rule ignore property chance values. This makes sense, as the rule ignore property leads to more rules being able to be run. More rules running results in longer run times.

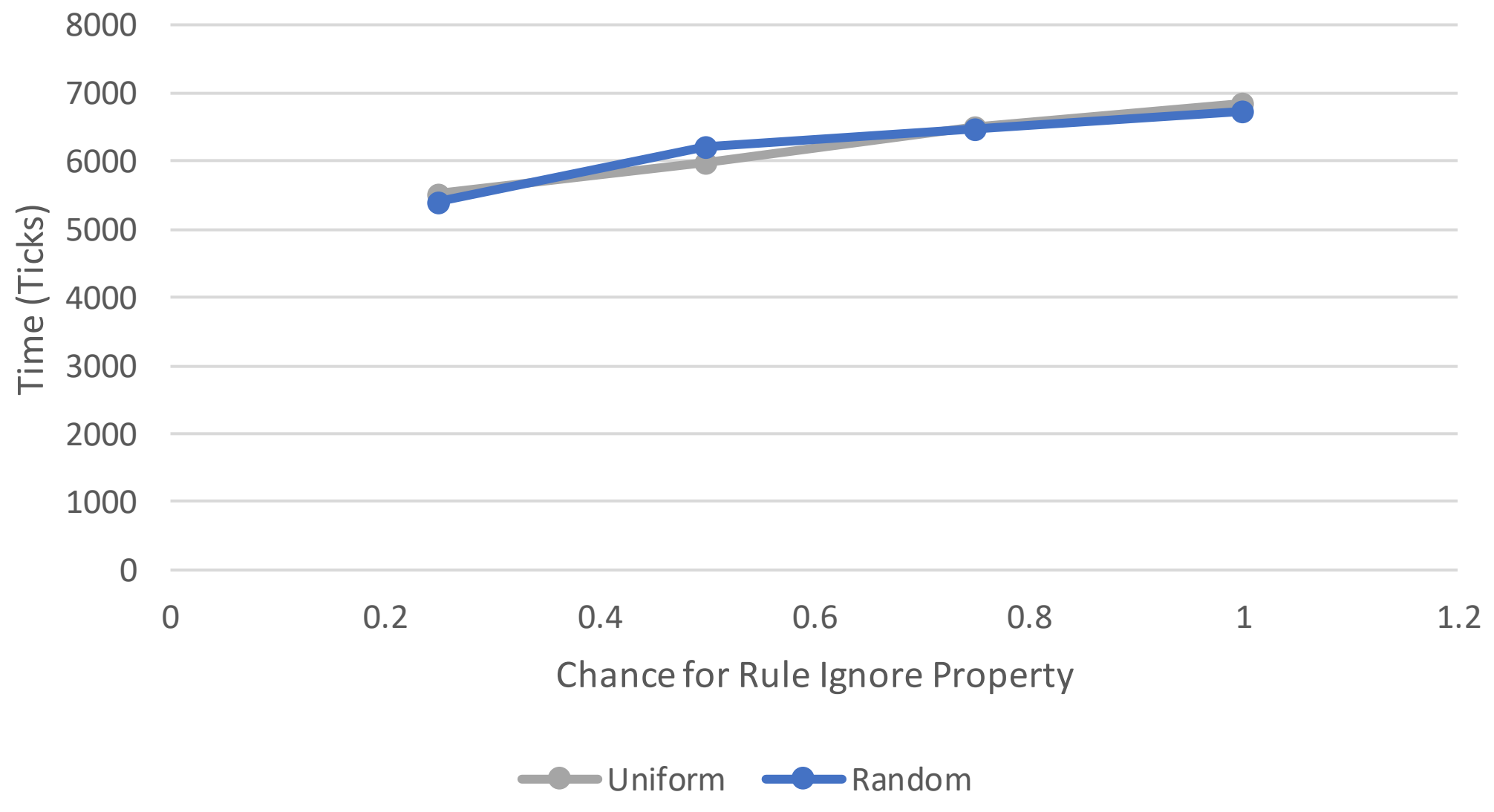

Figure 51. Rule Ignore Property Chance by Average Time to Next State.

Figure 52 shows the total network traversal time, as compared with the rule ignore property chance setting. Uniform and random networks preformed roughly the same for this test. There is a slight positive correlation between total network traversal time and rule ignore property chance. This makes sense as the total traversal time is mostly a summation of state changes' time costs. As seen in Figure 51, there is a slight positive correlation, which explains the slight positive correlation shown here.

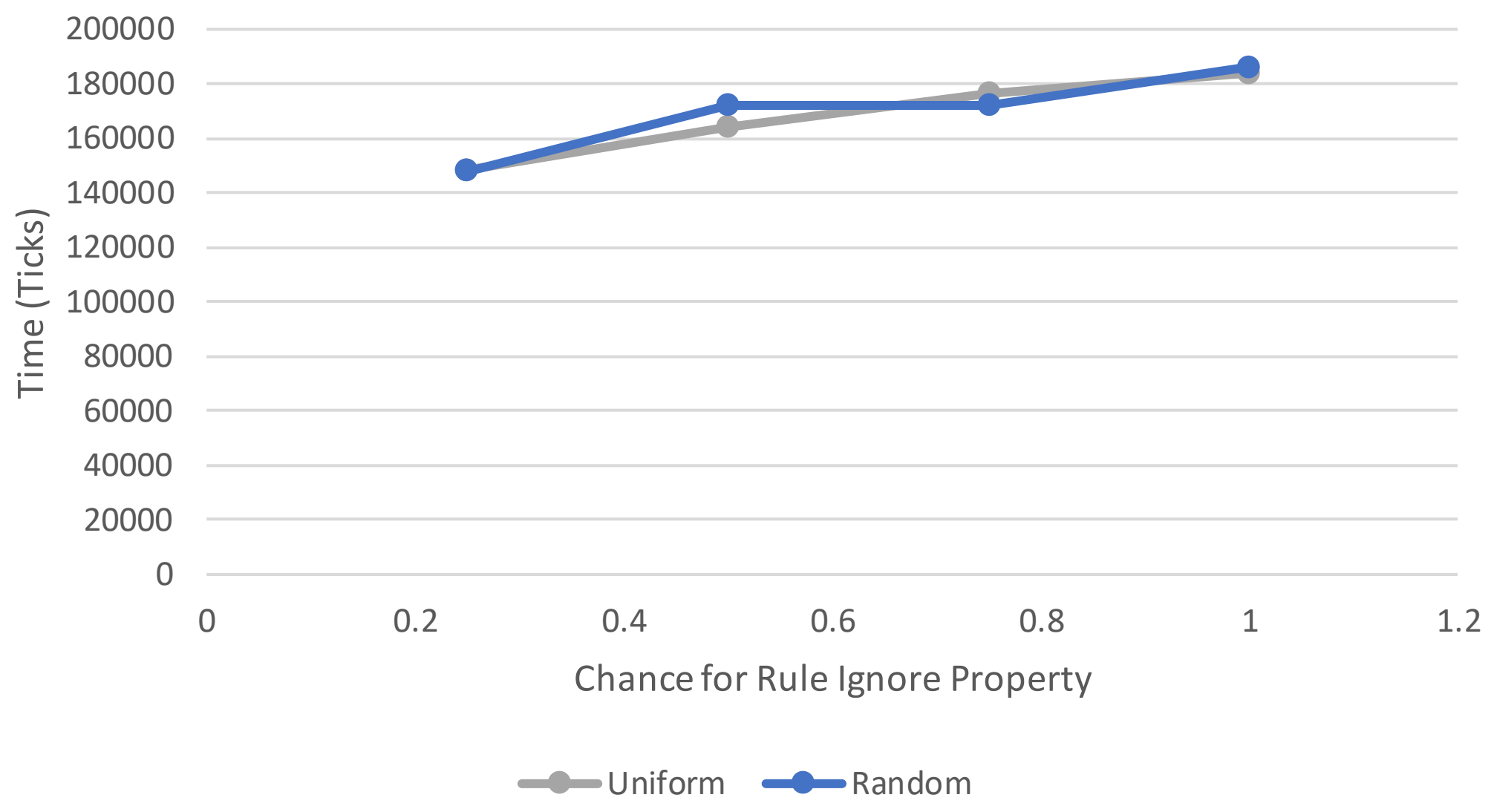

Figure 52. Rule Ignore Property Chance by Total Traversal Time.

Figure 53 shows the initial network storage size required (in bytes), as compared with the rule ignore property chance setting. There is a clear negative correlation shown in Figure 53. This is likely the result of plaintext style of storage. If a rule has the rule ignore property, then it will contain the string: "True." If the rule does not have the rule ignore property chance, then the string will instead be "False." Because there is a 1-character difference between these two words, rules with the ignore property will take up slightly less space. Because the data is stored as UTF-8, this is a 1-byte difference. At 100 rules by default, the difference between 25% and 100% should be about 75 bytes. This is consistent with the data presented in Figure 53.

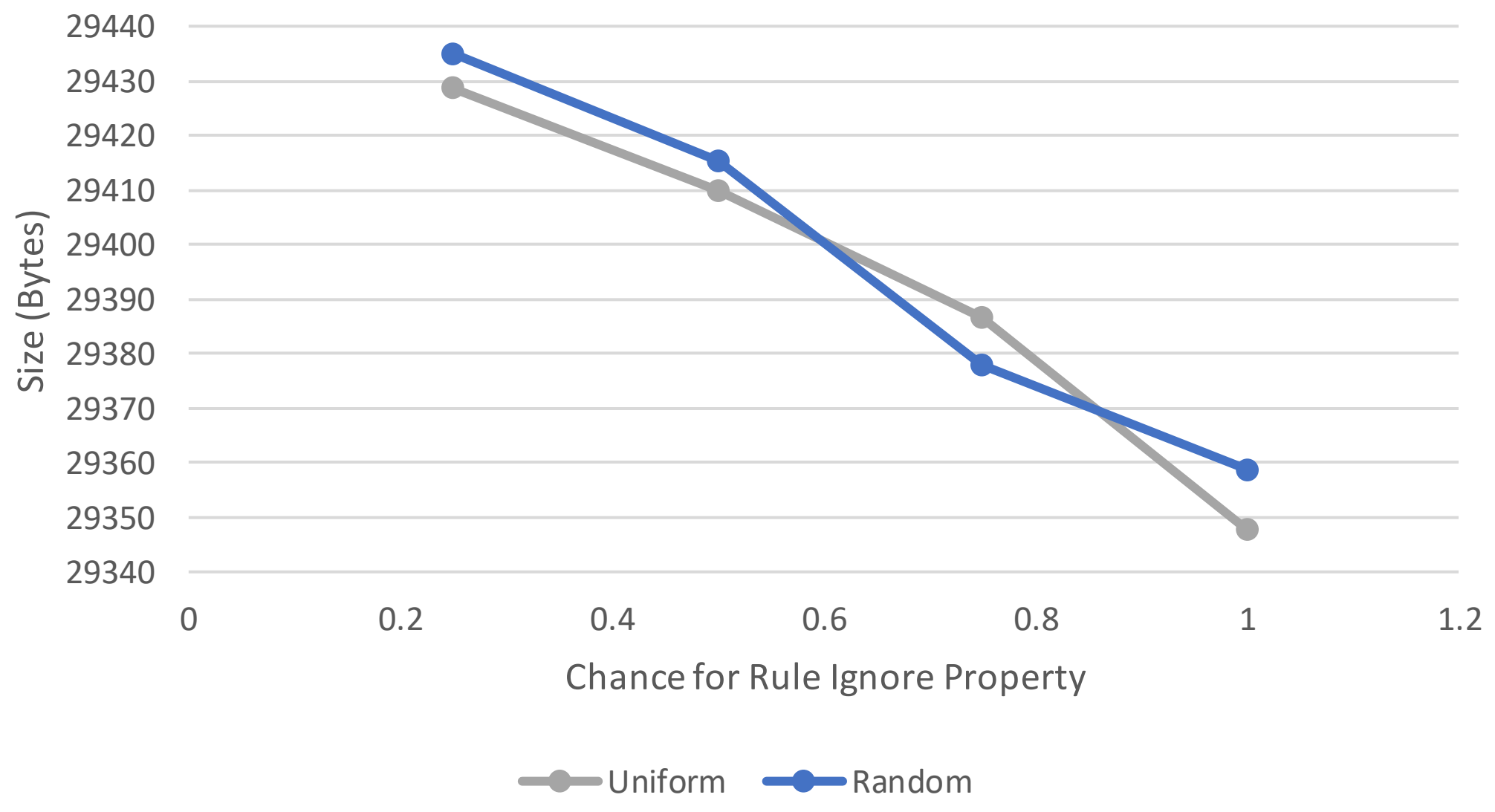

Figure 53. Rule Ignore Property Chance by Initial Network Storage Size.

Figure 54 shows the average storage size of one network state (in bytes), as compared with the rule ignore property chance setting. Uniform and random networks performed roughly the same for this test. There is a slight positive correlation between the state storage size and the rule ignore property chance setting value. This is potentially explained because the rule ignore property being set to true makes rules easier to run. More rules running results in more network changes, which are then recorded to the state file. Therefore, more rules with the ignore property causes an increase the state storage size.

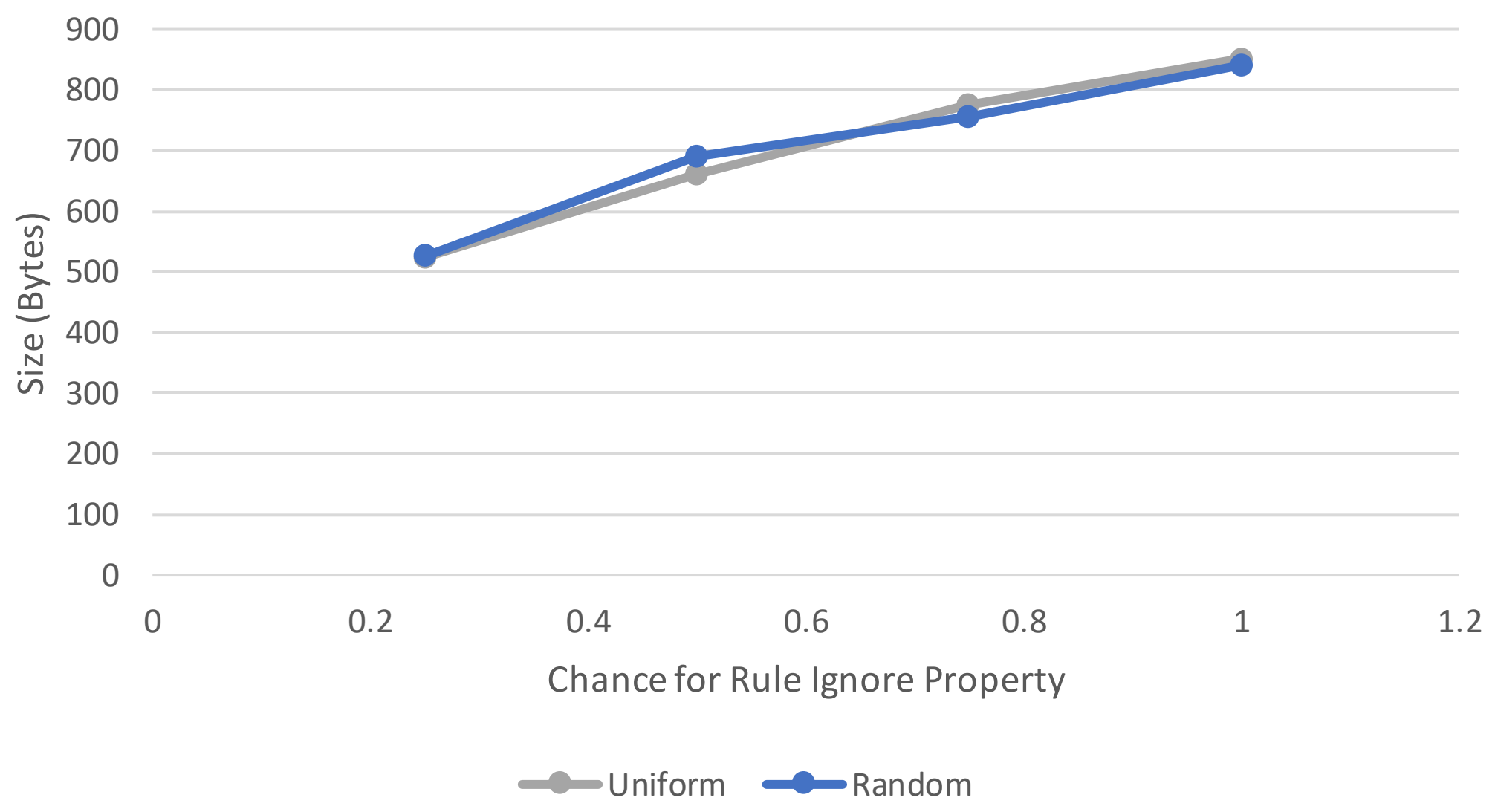

Figure 54. Rule Ignore Property Chance by Average Storage Size of One State.

Figure 55 shows the total storage size of the network (in bytes), as compared with the rule ignore property chance setting value. Uniform and random networks performed roughly the same for this test. There is a slight positive correlation between the network total storage size and the rule ignore property chance value. This is explained by the total storage size being a summation of all state sizes and the initial network size. While the initial network size has a negative correlation, it only had a difference of around 75 bytes. The average storage size of one state, on the other hand, was shown to have a difference of up to 300 bytes, which easily trumps the reduced initial size. Therefore, the positive correlation being a result of the data depicted in Figure 54 makes sense.

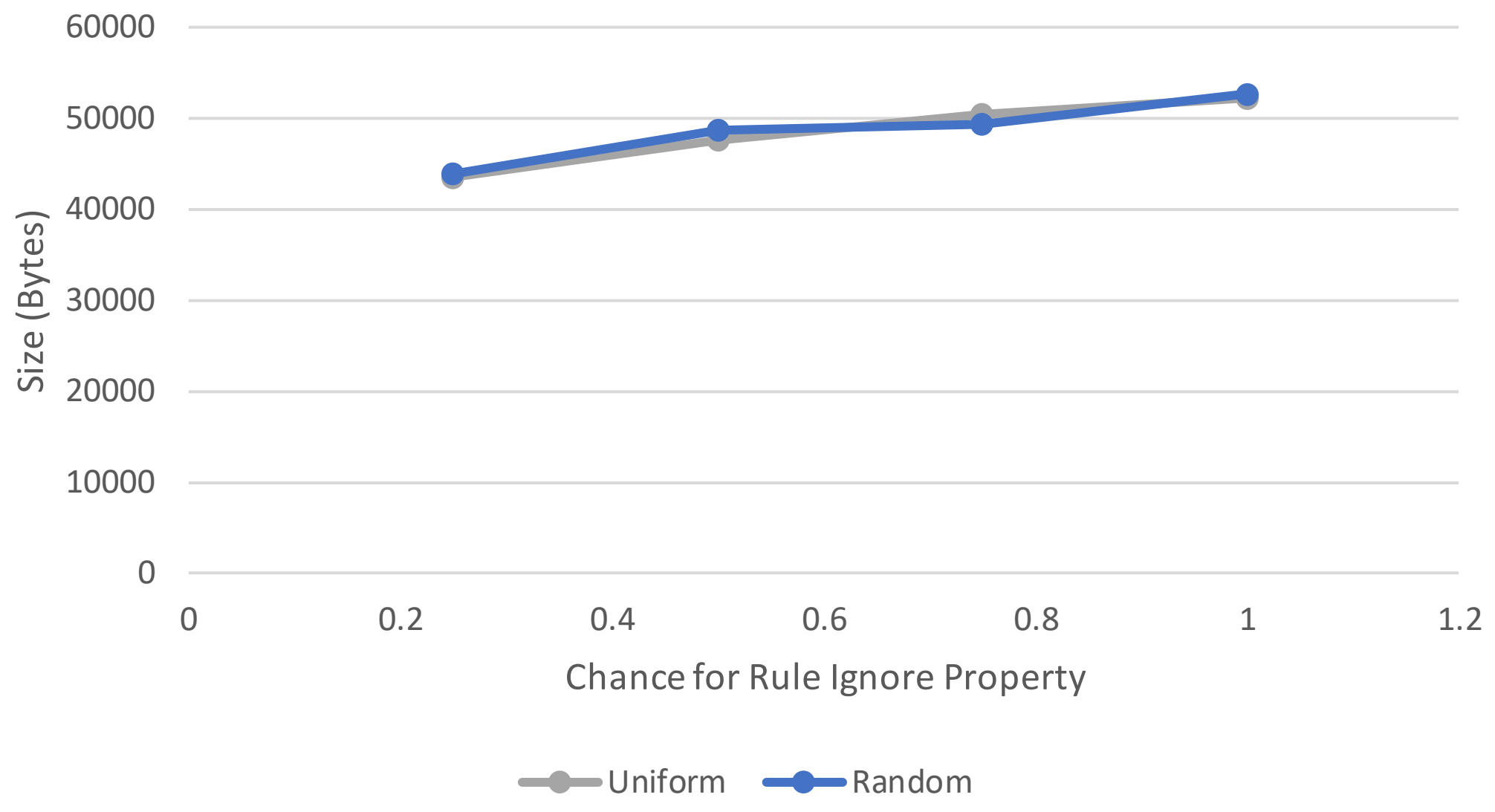

Figure 55. Rule Ignore Property Chance by Total Storage Size.

Figure 56 shows the average path length of a network, as compared to the rule ignore property chance setting value. Random networks have a slightly higher average path length, at all rule ignore property chances, except for 75%. No clear correlation is shown between the rule ignore property chance value and path length.

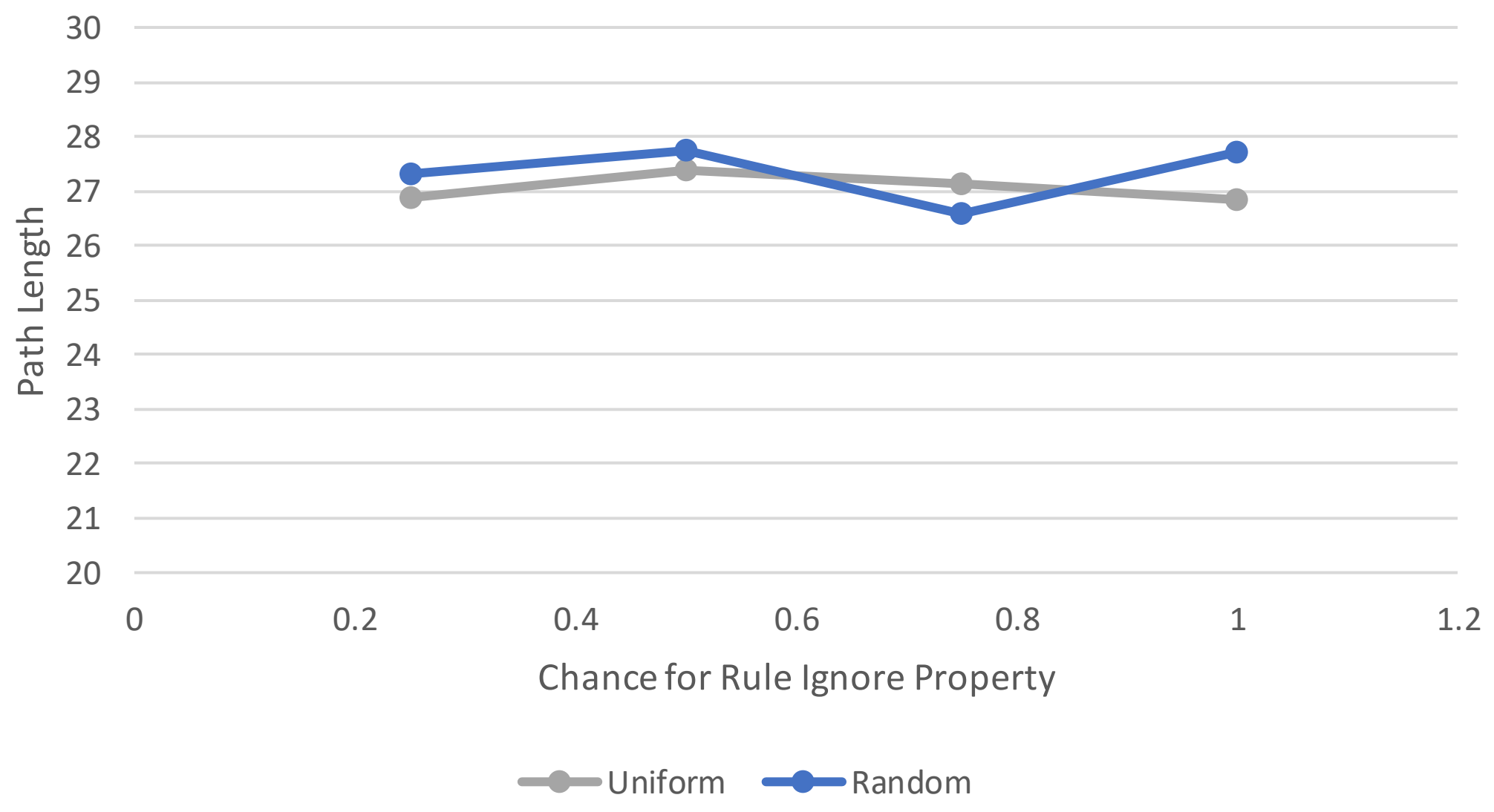

Figure 56. Rule Ignore Property Chance by Average Path Length.

***6.9. Rule Create Property Chance.***

The eighth area of analysis is how the rule ignore property chance value of the network effects its performance. Figure 57 shows the time to find the shortest path of a network (in ticks), as compared with the rule create property chance setting. Uniform networks start much higher than the other datapoints, at around 187 ticks. They then dip to 167 ticks. Uniform networks repeat this pattern, by increasing and decreasing again. Random networks, on the other hand, start at around 171 and then increase, until dipping down to around 166 ticks at the 100% rule create property chance level. The largest difference is only 20 ticks, suggesting that this variation may not be significant.

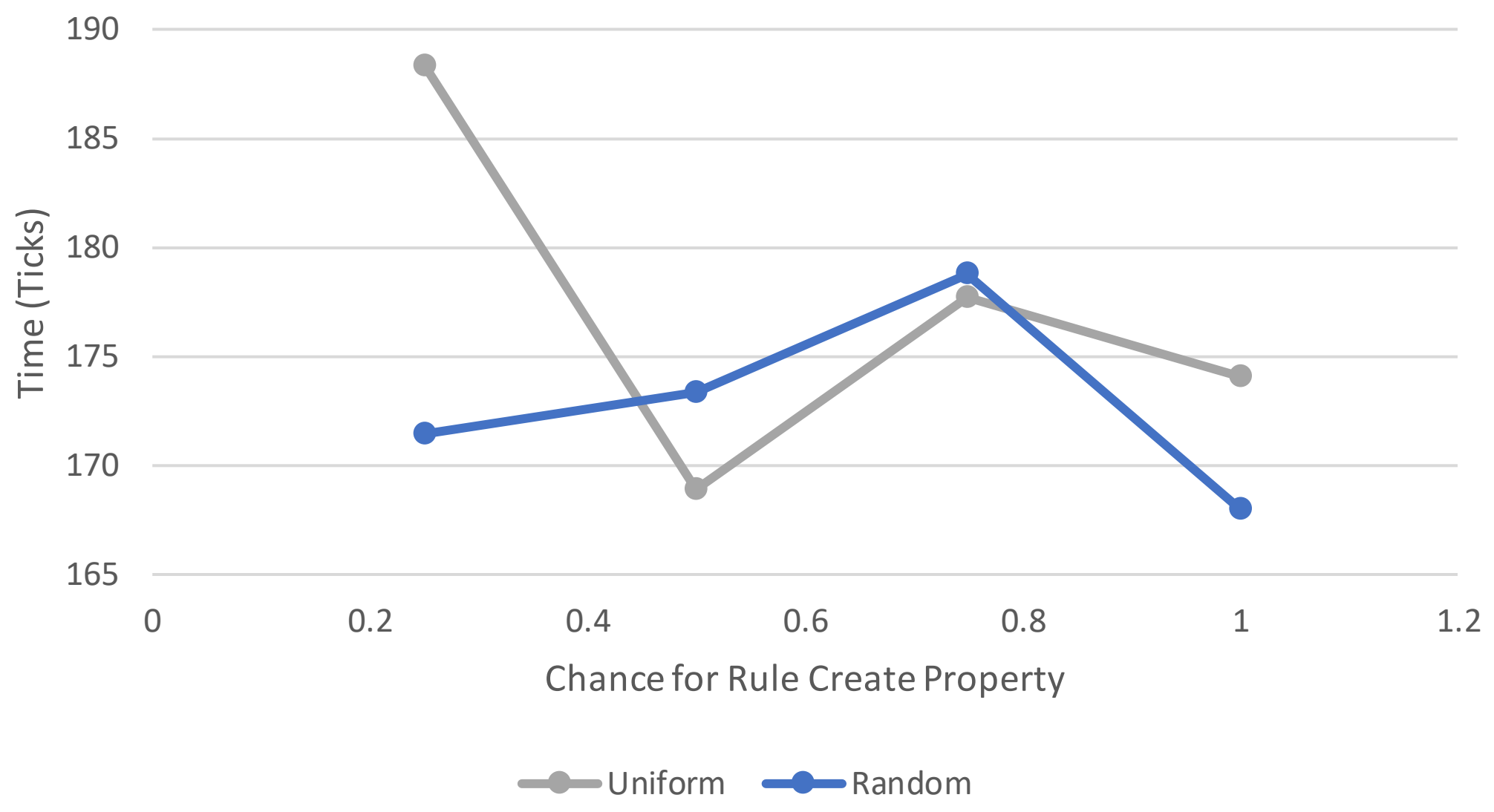

Figure 57. Rule Create Property Chance by Time to Find Shortest Path.

Figure 58 shows the average time to reach the next network state (in ticks), as compared with rule create property chance values. Uniform networks negatively correlate with this overall; however, random networks do not show a linear pattern. Random networks increase the time required greatly, from 25% to 50%; however, they then dip down at 75% and then increase again. Because the rule create property can lead to adding facts to rules during runtime, it is expected that the higher the rule create property chance value, the higher the average time to next state would be. It is unclear as to why the observed behavior contradicts this; however, the change is limited, as compared to the total value, suggesting that it could be caused by random variation.

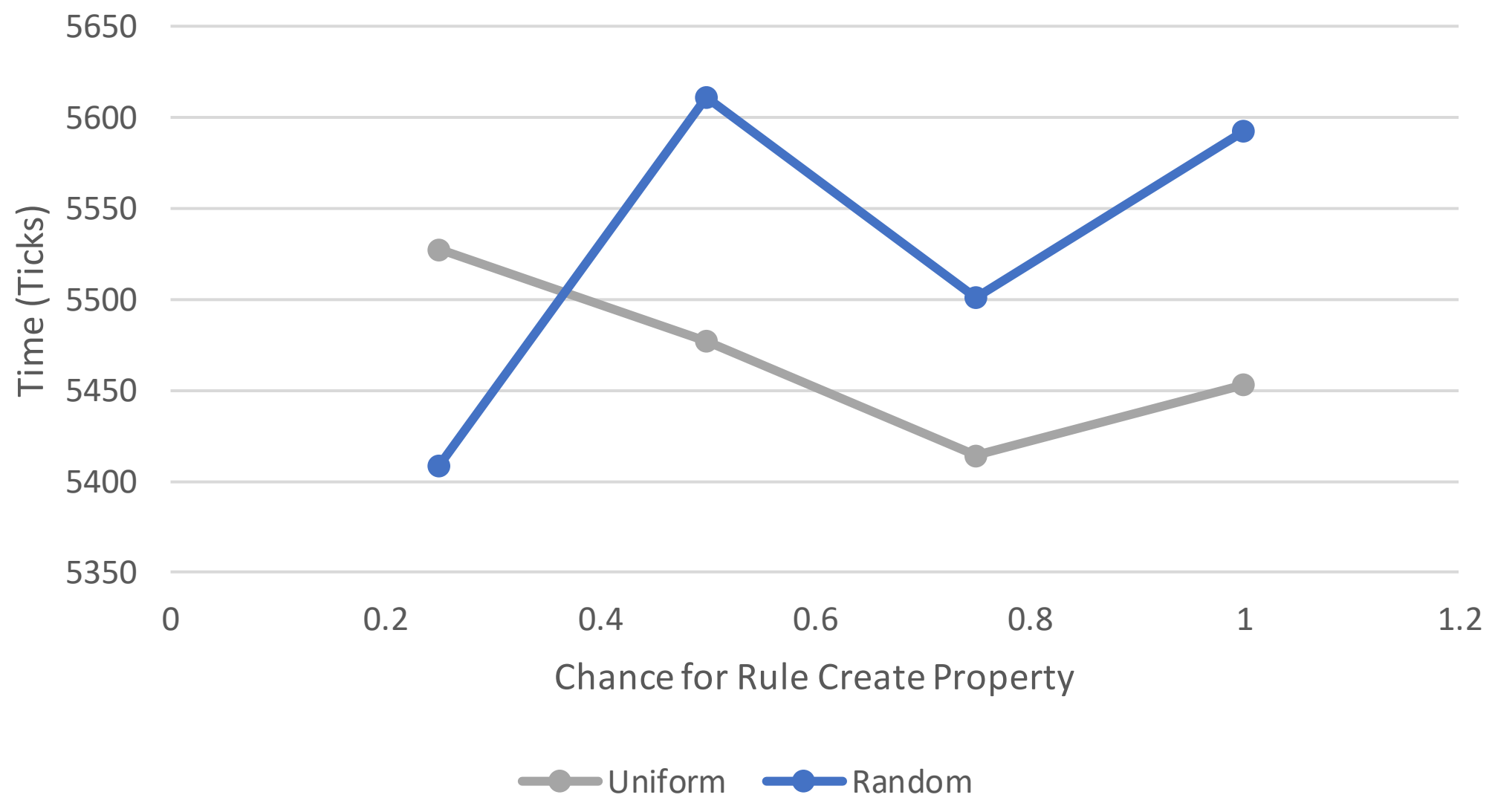

Figure 58. Rule Create Property Chance by Average Time to Next State.

Figure 59 shows the total network traversal time (in ticks), as compared to the rule create property chance setting values. Uniform and random networks exhibit inverse trends, with random networks starting with a large increase, then dipping back down at 75%, and then slightly increasing. Much like with the data shown in Figure 58, this contradicts the expected behavior. Because the rule create property can lead to adding facts to rules during runtime, it is expected that the higher the rule create property chance value, the higher the average time required to reach the next state would be. Again, though, the level of change is comparatively small, relative to the overall values.

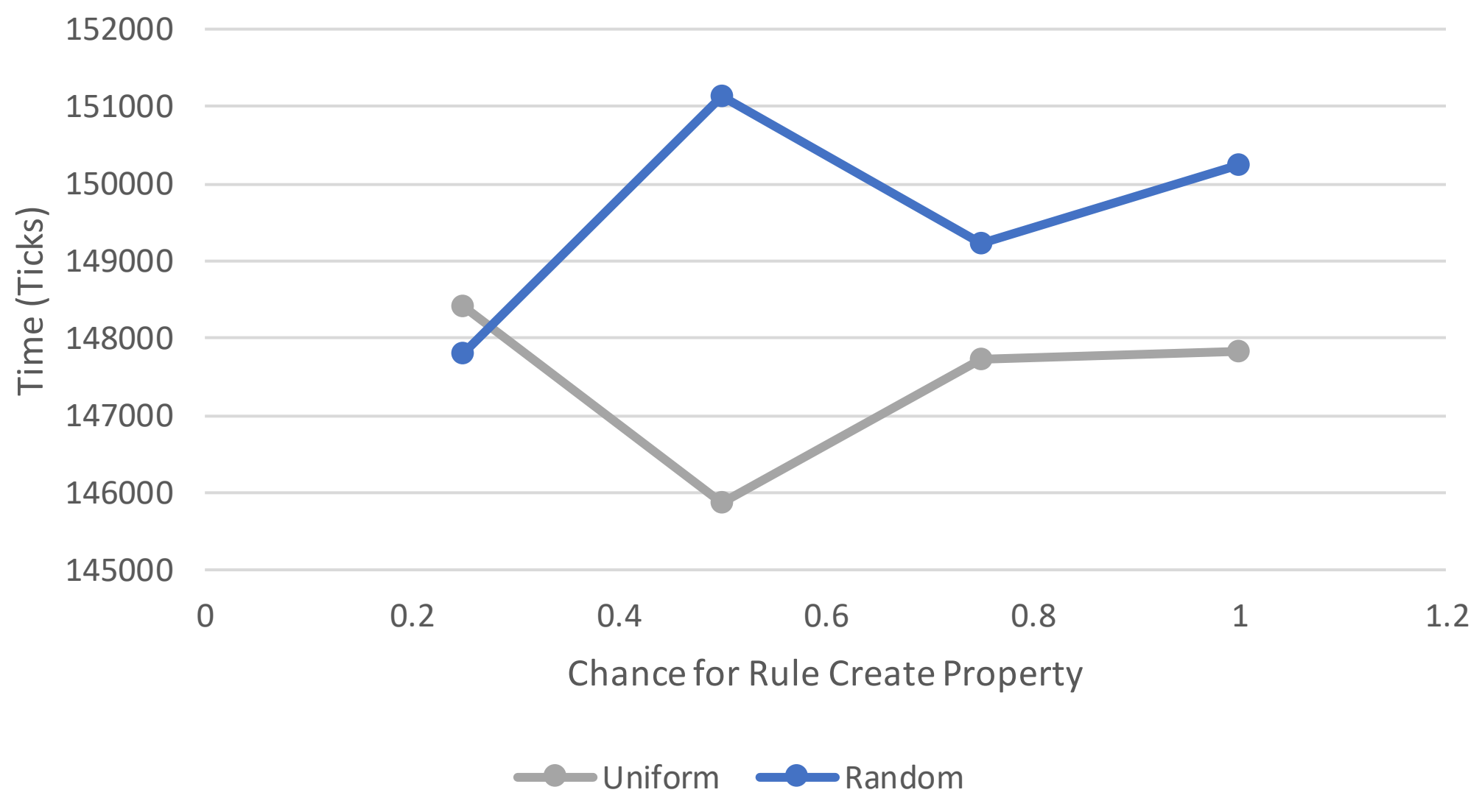

Figure 59. Rule Create Property Chance by Total Traversal Time.

Figure 60 shows the initial network storage size (in bytes), as compared to the rule create property chance setting values. The pattern shown is likely the result of the plaintext style of storage used. If the rule has the rule create property, then it will contain the string “True.” If the rule does not have the rule create property chance, then the string will instead be “False.” Because there is a 1-character difference between these two words, rules with the ignore property will take up slightly less space. Because the data is stored as UTF-8, this is a 1-byte difference. At 100 rules, by default, the difference between 25% and 100% should be about 75 bytes. This is consistent with the data in Figure 60.

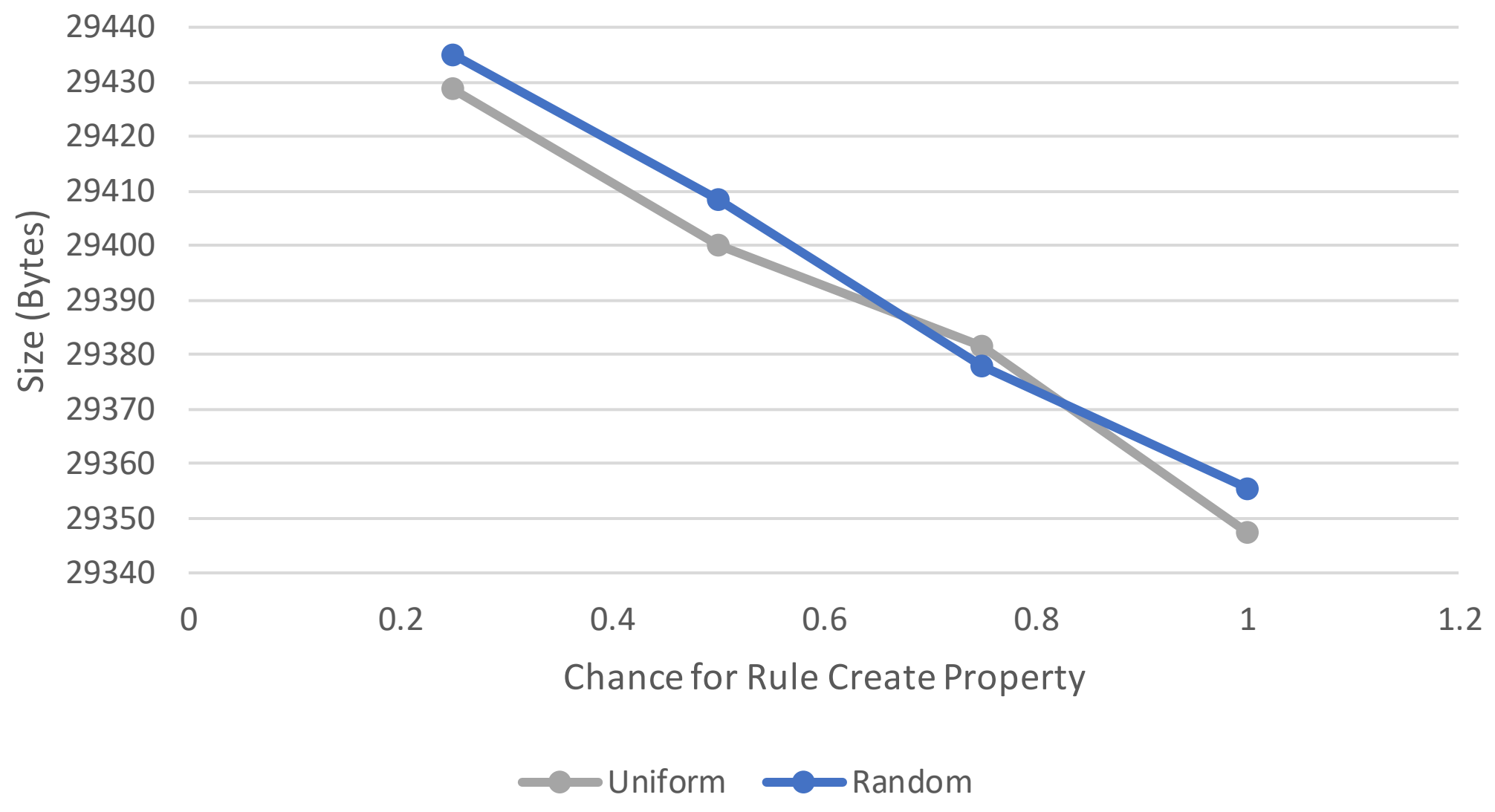

Figure 60. Rule Create Property Chance by Initial Network Storage Size.

Figure 61 shows the average storage size of one network state (in bytes), as compared with the rule create property chance setting values. The pattern shown here mirrors the pattern shown in Figure 59, namely that the uniform and random networks seem to exhibit the inverse patterns of each other. While random networks decrease between setting values of 50% and 75%, and then climb, uniform networks increase and then dip. The reason for the differences is because the rule create property chance affects if rules can create facts during runtime. These created facts are recorded in the state files, increasing their size. Given this, it is unknown why the observed data exhibits this pattern, as the expected pattern would be positive correlation.

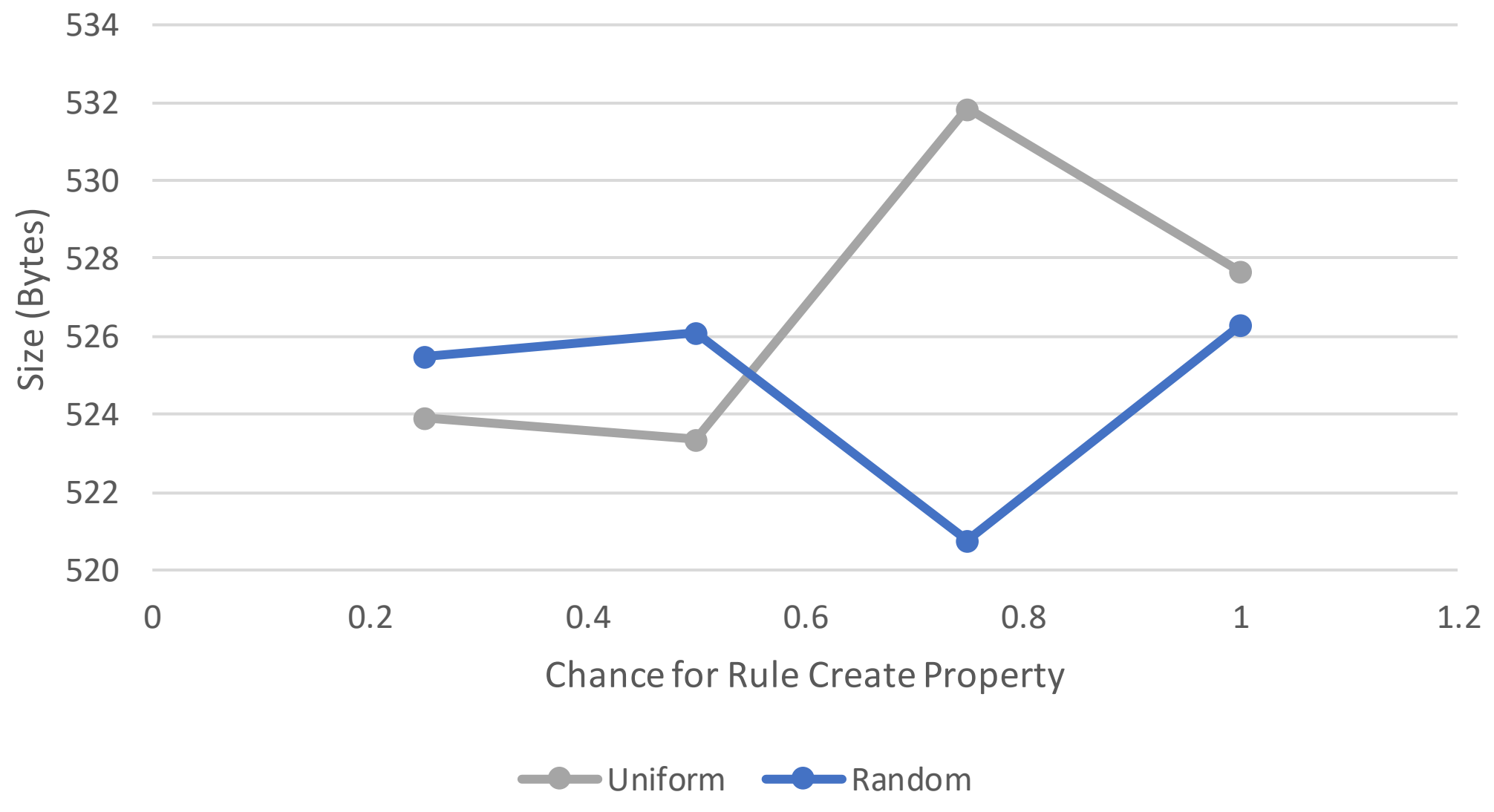

Figure 61. Rule Create Property Chance by Average Storage Size of One State.

Figure 62 shows the total size of the average network (in bytes), as compared with the rule create property chance setting values. Random networks have a clear negative correlation while uniform networks exhibit a peak at 75% and a trough at 50%. The rule create property allows rules to create facts during run time. These created facts are recorded in the state files, increasing their size. This is directly contradicted by the observed data for random networks. More testing is required to learn why this contradiction is observed.

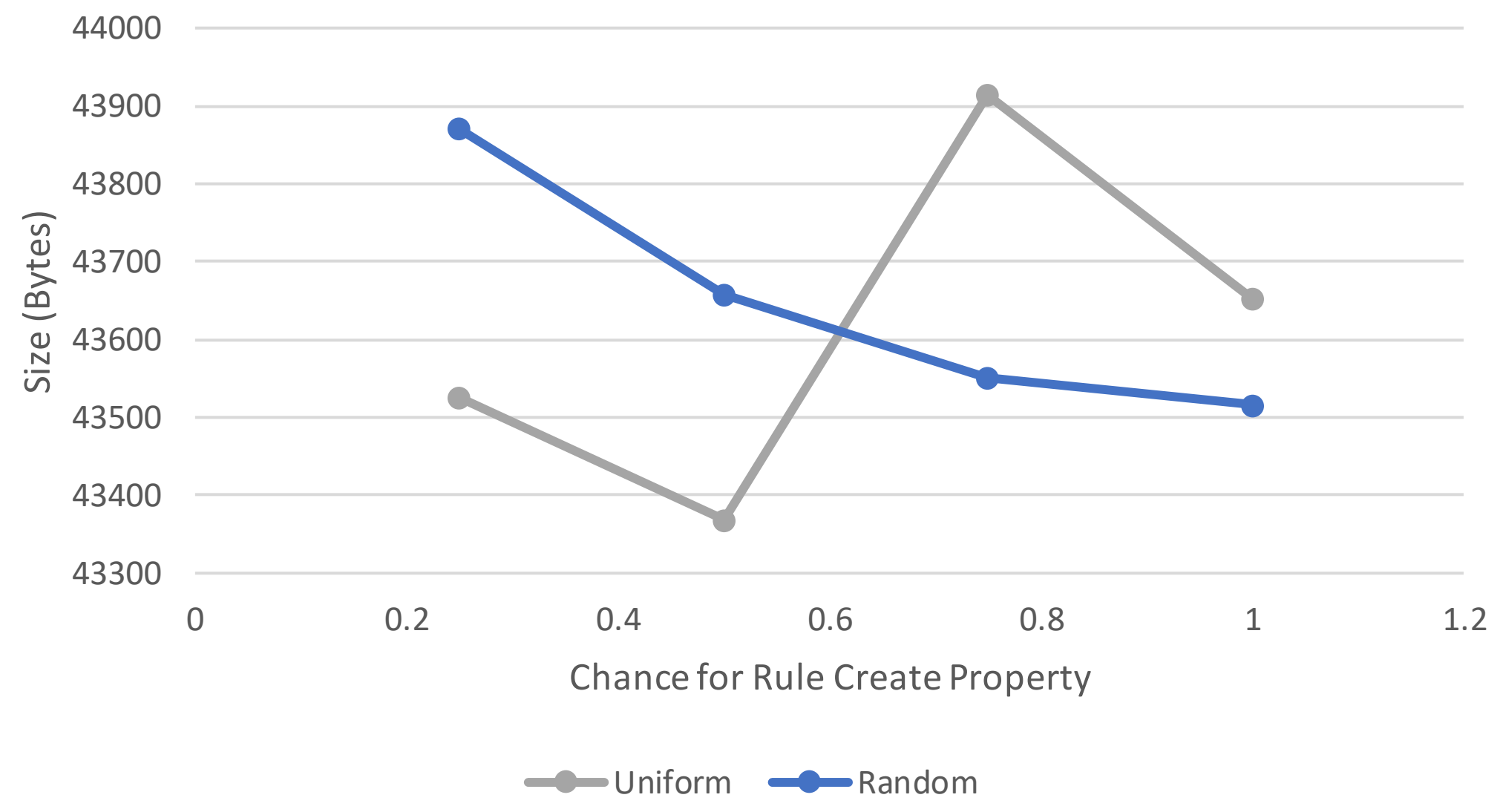

Figure 62. Rule Create Property Chance by Total Storage Size.

Figure 63 shows the average path length of a network, as compared with the rule create property chance setting values. Uniform and random networks preform roughly the same for this test. The path length stays roughly constant over the rule create property chance.

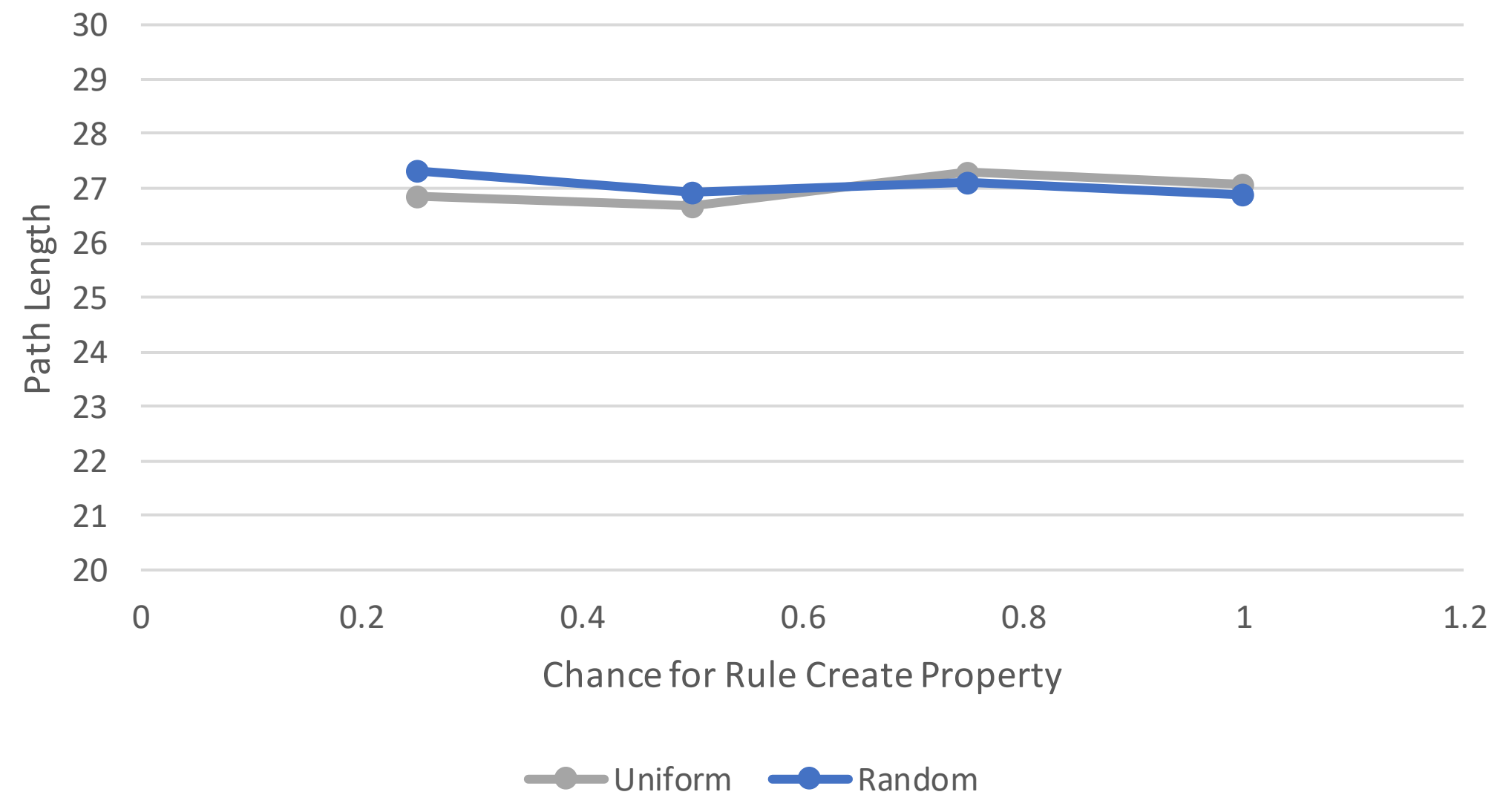

Figure 63. Rule Create Property Chance by Average Path Length.

**7. Data Analysis**

A large amount of the data collected has shown expected results. More objects will generally result in a longer run time and more storage being space needed. Where the results differ from expectations is where some key outcomes are found. However, in some cases, the differences are comparatively small and likely due to random variation in performance.

The first key outcome is that the link and container count had the largest effect on the time needed to find the shortest path. As shown in Figures 21 and 28, the search time is positively correlated with the link and container counts. In Figure 21, there is an average difference of 132.68 ticks between 300 and 100 links. In Figure 28, there is an average difference of 203.17 ticks between 200 and 50 containers.

The second key outcome is that the container count had the largest effect on the total traversal time. This is expected, as traversal time largely depends on the average path length. As shown in Figure 28, container count has a large positive correlation with average path length. This carries over into Figure 24 where the average time difference between 200 and 50 containers is 245,381.58 ticks.

The observed effects of link and container counts on network properties are expected and have been tested extensively before this paper. It is shown that the inclusion of generic rules and common properties does not notably effect the link and container count properties. Taking a look at generic rules specifically, Figures 8 through 14 show nearly the same properties as with rules from previous versions of the architecture. More rules results in a longer run time and more storage space being needed. Common properties have many of the same effects as links, containers, and rules. Having more common properties results in a longer run time and more storage space being needed. This expectation is generally supported by Figures 30 though 35. Where common properties differ is shown in Figure 29. This figure shows the time to find the shortest path, as compared with the number of common properties. It is expected that the figure would remain roughly flat or be positively correlated. Instead, the graph remains roughly flat except for at a common property count of 100. There is a distinct outlier above the trend of the graph for both network types. It is unknown why the common property count of 100, in particular, leads the search function to be roughly 50 ticks less efficient; however, this isn't a notable time cost.

The rule hybrid chance setting also had a notable effect on the run time of the network. While the property had no notable effect on the storage size, Figures 44 and 45 show a notable outlier for random networks at the 50% chance level. At a hybrid rule chance setting of 50%, random networks showed about a 100-tick increase in efficiency, as compared to other datapoints. At the same chance of 50%, uniform networks showed an outlier of about an 80-tick decrease in efficiency. The hybrid rule chance setting is the individual chance for each common property in a rule's 'post' column to be randomized, instead of matching the 'pre' column. It is unknown why a 50% chance, in specific, leads to outliers in both network types.

The rule ignore property had an interesting effect of having a slight positive correlation with run time and storage size, as shown in Figures 50 through 56. This is likely because the rule ignore property increases the number of rules that will be able to run which leads to a larger storage size being needed to store changes and a longer run time.

The rule create property had the most data that contradicted the expected results. Many of its figures had the random network type being the inverse of the uniform network type. Because the rule create property can lead to adding facts to rules during runtime, it was expected that the higher the rule create property chance, the longer traversal would take and the more space would be required to store the new facts. Figures 58, 59, 61, and 62 all contradict this expected result and require more testing to fully understand.

**8. Conclusions and Future Work**

While the Blackboard Architecture has been shown to be previously useful for numerous application areas, it historic lack of ability to deal with non-propositional logic relationships has limited its utility for certain applications. In these areas, the benefits of using the Blackboard Architecture can be greatly enhanced by using common properties and generic rules.

Using these additions, systems can be created which have well defined organizational and other structures with collections of data – and rules which act on this data, potentially triggering actions or modifying other data elements. The common properties and generic rules capabilities can be used in numerous application areas, such as automated diagnoses, penetration testing and system control decision making where other types of relationships would need to be implemented as a complex collection of rules and facts, when they can be more easily (and easily understandably) implemented through containers and links which are reasoned with using generic rules and supporting common properties.

While the addition of containers, links, generic rules and common properties presents benefits, it does incur a computational cost. While, in some cases, this cost is offset by savings on propositional logic processing, this paper has shown that more resources are needed – as would be expected – with larger and more complex networks. Thus, the main limitation to this architecture is the number of containers, links and generic rules used. The number of containers used, in particular, can greatly affect the system's run time and storage size.

A variety of enhancements are planned as potential areas of future work. One key area of exploration is the interaction between using generic rules and actions. As generic rules are already executing and rules trigger actions, there isn't a notable difference between this functionality for generic rules versus basic ones. However, actions provide an actuation capability within the environment and must be considered when making network operations decisions.

An additional consideration could also be applied to the generic rule matching process. At present, during generic rules' checking phase, only the 'before run' pairs are checked. Additional logic could be added to allow post-condition common properties to be part of this matching process with an option to ignore missing post-condition properties, if desired for a particular rule.

**Acknowledgements**

This work has been funded by the U.S. Missile Defense Agency (contract # HQ0860-22-C-6003). Thanks are given to Cameron Kolodjski, Jordan Milbrath and Matthew Tassava for their work on the software system that this technology operates within.